\documentclass[10pt, a4paper]{article}
\usepackage[utf8]{inputenc}
\usepackage[english]{babel}
\usepackage{amsmath, amssymb, amsfonts, amsthm}
\usepackage{graphicx}
\usepackage{hyperref}
\usepackage{geometry}
\usepackage{booktabs}
\usepackage{xcolor}
\usepackage{tikz}
\usetikzlibrary{shapes, arrows.meta, positioning, fit, calc, backgrounds}
\usepackage{cite}
\usepackage{algorithmic}
\usepackage[ruled,vlined,linesnumbered]{algorithm2e}
\usepackage{textcomp}
\usepackage{subcaption}

\SetKwInput{KwInput}{Input}
\SetKwInput{KwOutput}{Output}

\newcommand{\bs}[1]{\boldsymbol{#1}}
\newcommand{\vect}[1]{\boldsymbol{#1}}

\title{\textbf{Variational Graph Neural Networks for Uncertainty Quantification in Inverse Problems}}

\author{
    David González$^{1}$, Alba Muixí$^{2,3}$, Beatriz Moya$^{4}$, Elías Cueto$^{1}$ \vspace{0.2cm}\\
    \small $^1$ Keysight-UZ Chair of the Spanish National Strategy on AI. \\ \small Aragon Institute of Engineering Research (I3A), Universidad de Zaragoza, Spain \\
    \small $^2$ Laboratori de Càlcul Numèric (LaCàN), Universitat Politècnica de Catalunya - BarcelonaTech (UPC), Spain \\
    \small $^3$ Centre Internacional de Mètodes Numèrics en Enginyeria (CIMNE), Barcelona, Spain \\
    \small $^4$ PIMM Lab. Arts et Métiers Institute of Technology, Paris, France
}

\begin{document}
\maketitle

\begin{abstract}
\noindent The increasingly wide use of deep machine learning techniques in computational mechanics has significantly accelerated simulations of problems that were considered unapproachable just a few years ago. However, in critical applications such as Digital Twins for engineering or medicine, fast  responses are not enough; reliable results must also be provided. In certain cases, traditional deterministic methods may not be optimal as they do not provide a measure of confidence in their predictions or results, especially in inverse problems where the solution may not be unique or the initial data may not be entirely reliable due to the presence of noise, for instance. Classic deep neural networks also lack a clear measure to quantify the uncertainty of their predictions. In this work, we present a variational graph neural network (VGNN) architecture that integrates variational layers into its architecture to model the probability distribution of weights. Unlike computationally expensive full Bayesian networks, our approach strategically introduces variational layers exclusively in the decoder, allowing us to estimate cognitive uncertainty and statistical uncertainty at a relatively lower cost.

In this work, we validate the proposed methodology in two cases of solid mechanics: the identification of the value of the elastic modulus with nonlinear distribution in a 2D elastic problem and the location and quantification of the loads applied to a 3D hyperelastic beam, in both cases using only the displacement field of each test as input data. The results show that the model not only recovers the physical parameters with high precision, but also provides confidence intervals consistent with the physics of the problem, as well as being able to locate the position of the applied load and estimate its value, giving a confidence interval for that experiment.

\vspace{0.5em}
\end{abstract}

\section{Introduction}
The most transformative aspects that today's society is experiencing are undoubtedly the development and practical implementation of technologies applied to problems that just a few years ago were considered unfeasible due to cost or complexity (fast genomic sequencing; algorithms that predict the response to certain drugs; smart, personalized, and adaptive production). Without these technologies, precision medicine and industry, among others, would remain limited and applied to specific and undoubtedly costly cases. These phenomena have a common denominator, which is the ability to manage and exploit enormous volumes of data in real time for decision-making.

In particular, industry and precision medicine also share a need: to merge the physical world with the virtual world through digital twins \cite{chinesta2023hybrid}. To be effective, numerical simulation must be capable of delivering results in real time, a requirement that classical methods generally cannot meet due to their high computational cost.

Scientific Machine Learning has emerged as a promising solution. The development of artificial neural network architectures informed by physics (PINNs) \cite{RAISSI2019686}, thermodynamics (TINNs) \cite{hernandez2022thermodynamics}, or neural operators (DeepONet) \cite{wang2021learning,mousavi2026rigno} has demonstrated their ability to emulate complex physics at response speeds superior to those of classical solvers. However, the vast majority of these models are deterministic, meaning that given input data, they produce output data without informing the user about the reliability of that prediction.

This aspect can be considered critical in some inverse problems, where the aim is to obtain the causes that have generated certain effects, for example, to obtain the properties of the material that have produced certain displacements under a given load. It is well known that the nature of the input data (noised data) or the lack of uniqueness of the solution to the problem can lead to significant errors if one relies entirely on the prediction obtained with this type of architecture.

This paper proposes a hybrid architecture---combining the capabilities provided by the geometric biases of graph neural networks (GNNs) \cite{corso2024graph, pichi2024graph, tierz2025graph} with variational inference \cite{graves2011practical,pinheiro2021variational}---which, in our point of view, overcomes the limitations of conventional networks by providing efficient quantification of uncertainty. 

To do this, we use a combination of graph neural networks with variational layers that let us get a non-deterministic solution to the inverse problems posed, since it treats the weights of the network of these variational layers not as fixed values, but as random variables with a probability distribution. This allows our architecture to quantify two types of uncertainty: cognitive (or epistemic) uncertainty, which is the uncertainty associated with a lack of data or knowledge of the model---this type of uncertainty can be reduced with more data---; and statistical (or random) uncertainty, which is the uncertainty associated with the noise inherent in the input data.


There are two types of architectures that allow the probability distribution of weights to be processed. On the one hand, there are Bayesian neural networks (BNNs) \cite{kononenko1989bayesian}, which, as mentioned above, go a step further than conventional neural networks by treating the network weights as random variables that follow a probability distribution, rather than obtaining fixed values for the weights. 

The main objective of Bayesian inference in this context is to obtain the posterior distribution of the weights for some input data (training data). This is why BNNs can quantify the uncertainty in their predictions, making them particularly valuable in applications such as reliable uncertainty estimation or outlier detection.

However, training a BNN is computationally expensive. For this reason, approximate methods such as variational inference, among others, are used. This approach is widely used when working with very complex models or large data sets. Instead of calculating the probability distribution of the weights, variational inference seeks a simpler distribution that approximates it as closely as possible \cite{zhang2018advances}. Variational neural networks incorporate variational inference directly into the architecture to model uncertainty, improve robustness, and encourage better generalization.

In summary, variational inference is a Bayesian technique that replaces complex and intractable posterior distributions with simpler distributions that are more efficient to optimize, which is the alternative used in this work.

As already mentioned, there is a neural PDE solver architecture that has demonstrated a great capacity to adapt to changes in geometry and boundary conditions: graph networks \cite{battaglia2018relational,tierz2025feasibility}. Combined with a Bayesian variational approach to solving inverse problems, the resulting architecture promises to provide some very important features, particularly for the design of digital twins in changing environments \cite{gorpinich2025bridging, moya2022digital}.

The purpose of this work is precisely to explore the possibilities offered by both methodologies: on the one hand, the quantification of uncertainty provided by the Bayesian approach to the problem. On the other hand, the generalisability of graph networks, already demonstrated in the literature. We thus hope that the result will be a technique with a high degree of generalisation in the face of changing scenarios and even those not seen in training.

The outline of the paper is as follows. In Section \ref{s2} we introduce the proposed architecture for our networks. Section \ref{s3} briefly introduces the strategy followed for training. Section \ref{s4} presents the chosen numerical examples that demonstrate the sought advantages of our approach. Finally, in Section \ref{s5} we draw some conclusions.

\section{Neural Network Architecture}\label{s2}

In computational mechanics, continuum models are discretized using usually unstructured meshes (triangles, tetrahedrons) to adapt to the complex geometries of the different problems under study. However, classical deep learning methods, such as convolutional neural networks (CNNs) that dominate image processing, are constrained to structured data like regular pixel grids \cite{sermanet2012convolutional}. Projecting these meshes onto regular grids to use CNNs can introduce interpolation errors or loss of important topological information.

Graph Neural Networks (GNNs) generalize convolution to domains that are not necessarily regular \cite{corso2024graph,pichi2024graph, AGonzalezGNS, Pfaff2020MeshGraphNets}. A graph is a mathematical structure used to represent relationships between objects. It consists of vertices that represent the objects and edges that represent the relationships between those objects. This relationship can be easily adapted to a finite element mesh, where the nodes correspond to the vertices of the graph and the edges reflect the connectivity of the finite elements. So, GNNs treat the computational mesh of the model directly as if it were a graph $\mathcal{G} = (\mathcal{V}, \mathcal{E})$, composed by a set of vertices $\mathcal V$ and edges $\mathcal E$, preserving the original connectivities and guaranteeing an indispensable physical property: it is invariant to the ordering of the nodes, i.e., the result does not depend on the order in which the nodes have been enumerated. 

\begin{figure*}[ht]
\centering
\includegraphics[width=0.4\linewidth]{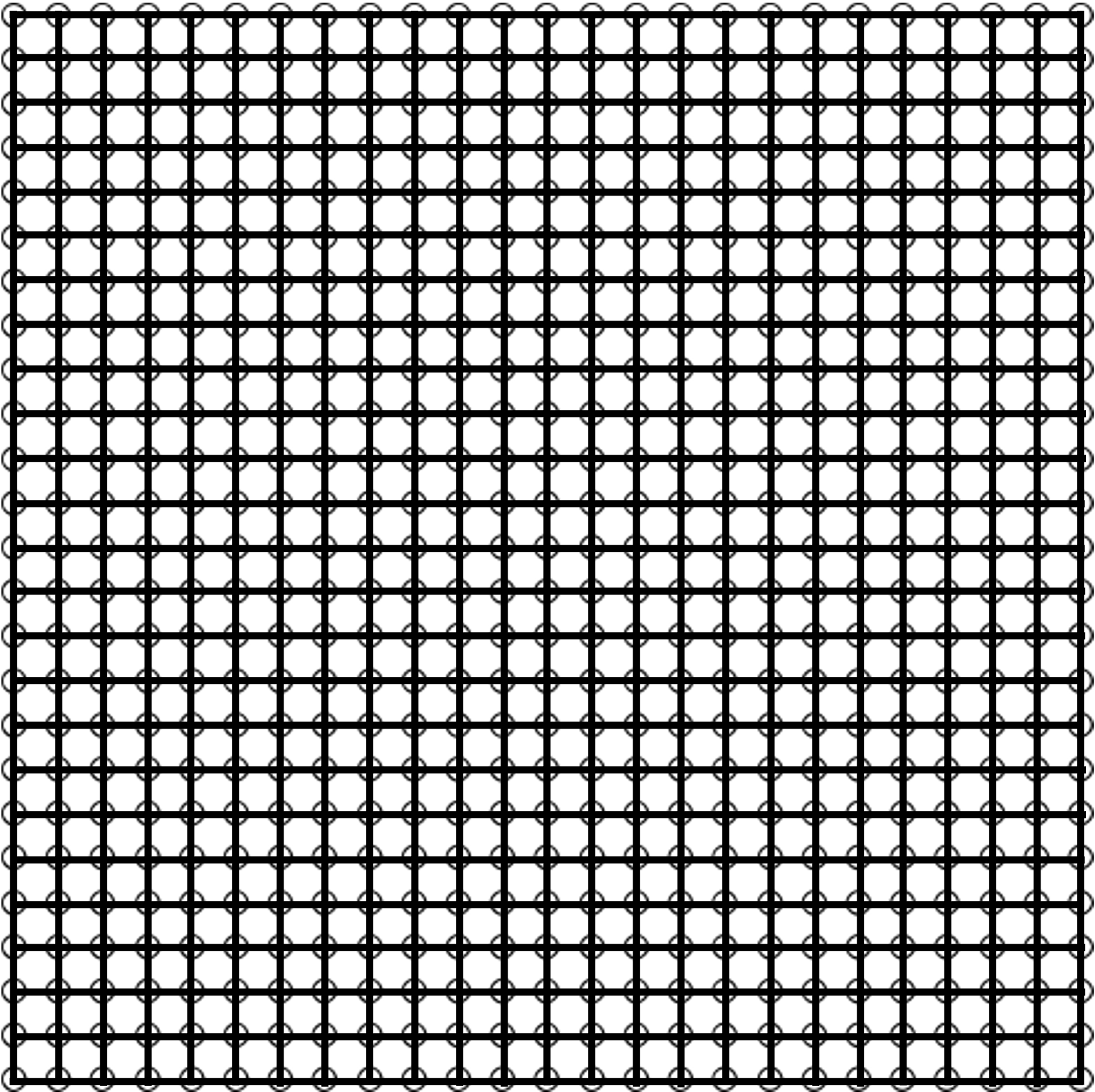}
\hspace{0.5cm}
\raisebox{-0.25cm}{\includegraphics[width=0.44\linewidth]{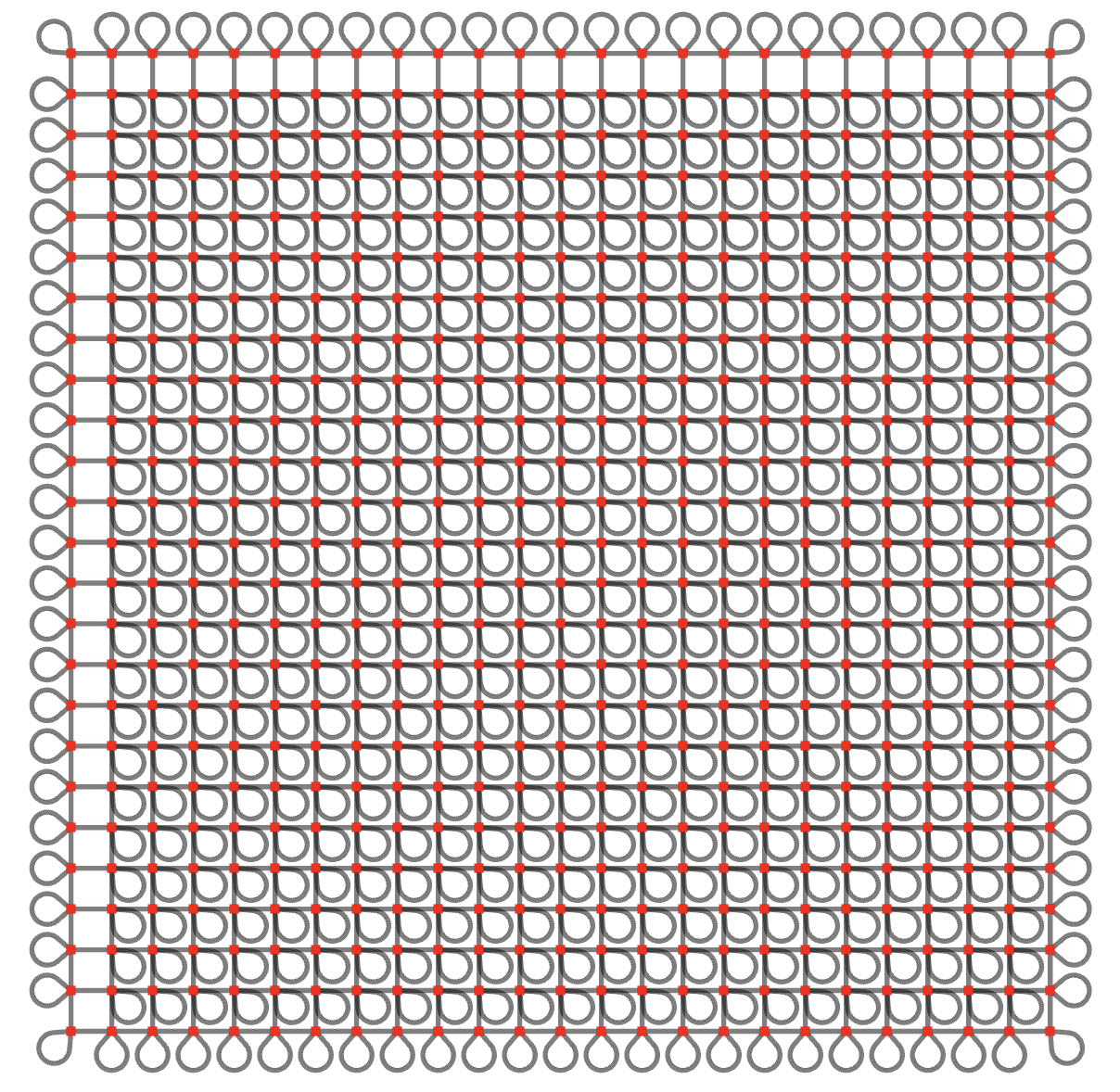}}
\caption{Example of a 2D finite elements mesh (left) and the equivalent self-informed graph (right). This graph is the one used in the example in Section \ref{ex:NASA}.}
\label{Grafo01}
\end{figure*}

In Fig. \ref{Grafo01}, we can see the equivalence between a 2D finite element mesh (left) and an example of a self-informed graph on the right. In graphs, it is possible to increase the information received by each vertex by adding information specific to that vertex.

Graph networks employed as a neural Partial Differential Equation (PDE) solvers classically  follow the procedure proposed by Battaglia et al. \cite{battaglia2018relational} and proceed through an encoder, a processor and a decoder, whose structure we describe next.

\subsection{Encoder}

The encoder processes the input data (dimension of a mesh node, processing all at once in the dimension reserved for the batch) and projects it onto the latent space of a higher dimension than the input one. Let $\bs{x}_i$ be the feature vector of node $i$. In this work, the coordinates and displacements (observable variables) of each node are considered. In order to characterize and differentiate each experiment or simulation to which this displacement field is attributed, the mean of this displacement field of the experiment is also taken into account. The characteristics of the edge are denoted by $\bs{a}_{ij}$, which in this work considers the relative distance between neighboring nodes (defined by the connectivity of the graph). The encoder consists of two independent multi-layer perceptron (MLP) networks for nodes and edges:

\begin{eqnarray}
    \bs{v}_i^{(1)} = f(\bs{W}_{v,1} \bs{x}_i + \bs{b}_{v,1}), \quad \bs{e}_{ij}^{(1)} = f(\bs{W}_{e,1} \bs{a}_{ij} + \bs{b}_{e,1}),\\
        \bs{v}_i^{(l)} = f(\bs{W}_{v,l} \bs{v}_i^{l-1} + \bs{b}_{v,l}), \quad \bs{e}_{ij}^{(l)} = f(\bs{W}_{e,l} \bs{e}_{ij}^{l-1} + \bs{b}_{e,l}), \quad l=2,3
\end{eqnarray}
where $f$ is a nonlinear activation function. Specifically, in this work, the SwishLayer activation function has been used \cite{ramachandran2017searching}, and three fully connected layers have been considered for each MLP. The subscripts $i,j$ run through the number of nodes in the model.


\subsection{Processor}
The processor is the interaction core that performs $m$ message passing steps to propagate information across the domain using the connectivity defined in the graph. This is essential for simulating the physics of the problem, since local interactions between neighboring nodes give rise to global behaviors of the entire model. For each message step $k=1,\ldots,m$:
\begin{enumerate}
\item The information for each edge of the graph is updated by calculating the ``message''---interaction value from the information stored in the connected nodes,
\begin{equation}
    \bs{e}_{ij}^{(k+1)} = \bs{e}_{ij}^{(k)} + \phi_e \left( \bs{v}_i^{(k)}, \bs{v}_j^{(k)}, \bs{e}_{ij}^{(k)} \right),
\end{equation}
where $\phi_e$ are basic neural networks (3-layer fully connected MLPs with SwishLayer activation functions).

\item Message passing step, where the aggregate message per node $\bs m_i$ is calculated as
\[\bs m_i = \sum_{j \in N(i)} \bs{e}_{ij}^{(k+1)}\]
where $N(i)$ represents the neighbourhood of node $i$, i.e., the set of nodes connected to it by the graph. 
The number of message passes is a key parameter in the performance of graph networks. In \cite{under-reach-mpi}, the minimum number of message passes required to ensure a given level of prediction accuracy is studied in depth, depending on the type of partial differential equation governing the physical process.

\item Node aggregation and updating step. For each node, it aggregates the messages from its neighborhood $N(i)$ and updates its state.
\begin{equation}
    \bs{v}_i^{(k+1)} = \bs{v}_i^{(k)} + \phi_v \left( \bs{v}_i^{(k)}, \bs m_i \right) = \bs{v}_i^{(k)} + \phi_v \left( \bs{v}_i^{(k)}, \sum_{j \in N(i)} \bs{e}_{ij}^{(k+1)} \right),
\end{equation}
where $\phi_v$ are again basic neural networks (3-layer fully connected MLPs with swishLayer activation functions). It is important to note that the number of steps $m$ determines how much information each node receives. As discussed in \cite{under-reach-mpi}, an insufficient $m$ causes the phenomenon known as \textit{under-reaching}, where physical information does not have time to propagate throughout the domain (analogous to the CFL condition in explicit numerical methods), preventing or making convergence of the inverse problem more difficult.
\end{enumerate}

\subsection{Variational Decoder}
Unlike Graph Networks, where all the weights calculated by the network are constant, and unlike Bayesian Networks, where all the weights in the network are distributions (which considerably increases the computational cost and significantly hinders the convergence of the network), in this work we propose a hybrid approach. The Encoder and Processor are deterministic, extracting deep features and therefore their weights are considered constant, and uncertainty is modeled only in the decoder, defining a Variational Decoder whose weights are probability distributions.

To solve the inverse problem, we seek the posterior distribution of the parameters to be estimated $\bs{y}$ (e.g., Young's modulus, or the position and value of the load) given the input data or observations $\bs{x}$ (e.g., the displacement field from the experiment). In this work, we assume that the output follows a Normal distribution, so the network seeks to obtain the best weights to obtain the mean and variance that best approximate the solution of the training data, i.e.,
\begin{equation}
    p(\bs{y}|\bs{x}) = \mathcal{N}(\bs{y} | \vect{\mu}_\theta(\bs{x}), \text{diag}(\vect{\sigma}_\theta^2(\bs{x}))),
\end{equation}
where $\theta$ represents the network weights. The proposed network architecture therefore has two outputs: one predicts the mean vector $\vect{\mu}$ and the other predicts its variance $\vect{\sigma}^2$, which represents the noise in the network predictions, ensuring a certain degree of numerical stability.

\begin{figure*}[h!]
\centering
\resizebox{\textwidth}{!}{
\begin{tikzpicture}[
    node distance=1.5cm,
    >=stealth,
    font=\sffamily\small,
    block/.style={rectangle, draw=black!70, very thick, rounded corners, minimum height=5.35cm, minimum width=2.2cm, align=center, fill=green},
    block1/.style={trapezium, trapezium angle=60, draw=black!70, very thick, rounded corners, minimum height=2.5cm, minimum width=1.5cm, align=center, fill=white, shape border uses incircle},
    arrow/.style={->, very thick, color=black!70},
    label/.style={text=black, font=\bfseries} 
]
    \node(a) at (0.5,0.25) {};
    \node(a1) at (1,0.3) {};
    \node(b) at (3.5,0.35) {};
    \node(b1) at (4.1,0.3) {};
    \node(c) at (9.75,2.3) {};
    \node(c1) at (9.0,0.3) {};
    \node(d) at (12.5,0.3) {};
    
    \node(input) [align=center] {\includegraphics[width=1.5cm]{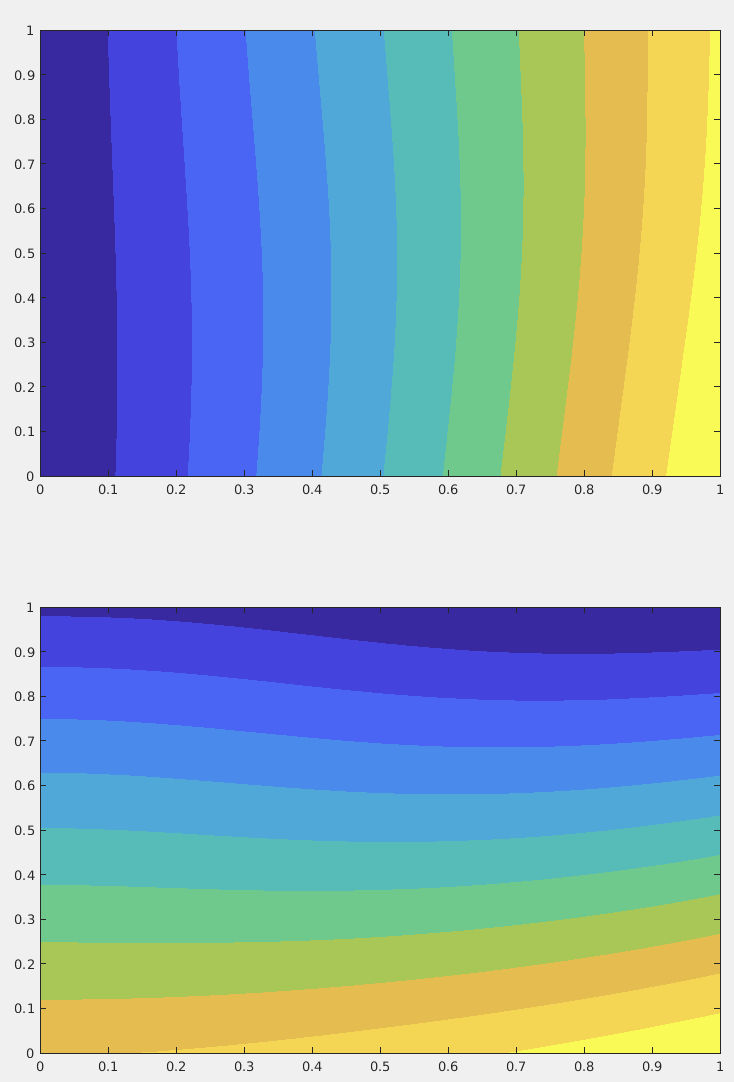} \\ \textbf{Input Node} \\ $(\bs{u}, \overline{\bs{u}},\Gamma_u)$ \\ \includegraphics[width=1.5cm]{InputB.png} \\ \textbf{Input Edge} \\ $(\bs{x},  \overline{\bs{u}})$};
    
    \node(encoder) [block1, below right= of a, fill=blue!10, rotate=90] {};
    
    \node(processor) [block, right= of b, fill=green!10, minimum width=4cm] {\textbf{Processor} \\ \scriptsize Message Passing \\ $\times m$ Passes \\ \textit{Interaction Network}};

    \node(decoder) [block1, right= of c, fill=red!10, rotate=270] {};

    \node(enc_text) [above=0.8cm of encoder.center] {\textbf{Encoder}};
    \node(enc_text) [above=0.0cm of encoder.center] {\scriptsize MLP per node/edge};
    \node(enc_text) [above=-0.8cm of encoder.center] {$\mathbb{R}^d \to \mathbb{R}^{h>d}$};

    \node(dec_text) [above=0.8cm of decoder.center] {\textbf{Variational}};
    \node(dec_text) [above=0.4cm of decoder.center] {\textbf{Decoder}};
    \node(mu_box) [rectangle, draw=red!80, fill=white, below right=-0.1cm and -2cm of decoder.north, minimum width=1.75cm] {$\vect{\mu}$ (mean)};
    \node (sigma_box) [rectangle, draw=red!80, fill=white, below=0.1cm of mu_box, minimum width=1.75cm] {$\vect{\sigma}$ (std dev)};

   \node (sampling)[circle, draw=black, thick, right=0.8cm of d, inner sep=2pt] {$\mathcal{N}$};
    
    \node (output)[right=0.8cm of sampling, align=center] {\includegraphics[width=1.5cm]{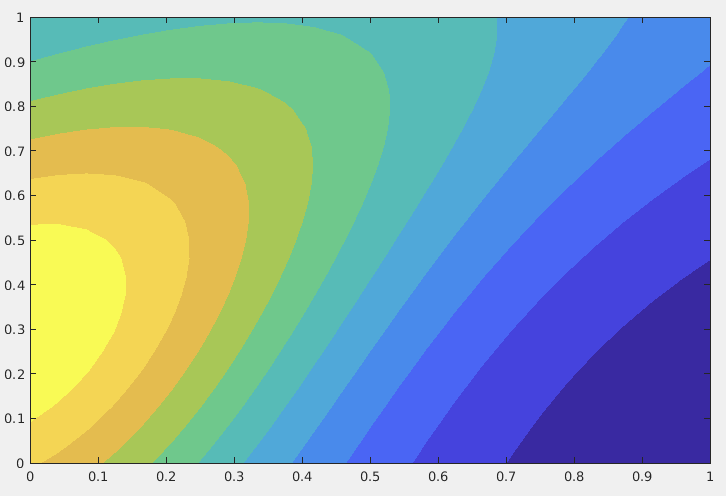} \\ \textbf{Output} \\ $C(\bs{x}) \sim \mathcal{N}(\mu, \sigma)$};

  \draw[arrow] (a1) -- (encoder);
  \draw[arrow] (b1) -- node[above]{\scriptsize Latent} node[below]{\scriptsize Graph} (processor);
  \draw[arrow] (c1) -- (decoder);
    
   \coordinate(dec_split) at ($(decoder.north)!0.3!(sampling.center)$);
   \draw[thick, red!80] (sampling.west) -- (dec_split);
   \draw[->, thick, red!80] (dec_split) |- (mu_box.east);
   \draw[->, thick, red!80] (dec_split) |- (sigma_box.east);

   \draw[arrow] (sampling) -- (output);

   \node(nota1) [below=0.2cm of processor, text=blue] {Deterministic Logic};
   \node [below=-0.1 of nota1, text=blue] {(Physics)};
   \node(nota2) [below right=1.6cm and -0.05cm of decoder, text=red] {Probabilistic Logic};
   \node [below=-0.1cm of nota2, text=red] {(Uncertainty)};

\end{tikzpicture}
}
\caption{A sketch of the Variational Graph Neural Network Architecture.}
\label{fig:arquitectura}
\end{figure*}

Fig. \ref{fig:arquitectura} shows a diagram of the network architecture proposed in this work. With this proposed architecture, non-deterministic behavior is considered only in the Decoder, allowing the Processor to learn complex physical characteristics in a deterministic and efficient manner.

The dimensions of the input vector required by the network are those of the displacement field of a single node---2D or 3D, depending on the problem---concatenated with the average displacement of the experiment---used as a global feature to identify and differentiate each experiment and which adds, again, 2 or 3 dimensions to the input vector---and the information regarding whether or not the node belongs to the contour using a binary variable.

Additionally, graph information is provided, describing the node's neighborhood (its connections with neighboring nodes). Specifically, it requires the Euclidean distance between neighboring nodes in each spatial direction and the modulus of that distance. This structure also includes the global characteristic of the experiment --in this work, it has been considered sufficient to use only the mean of the displacements of the entire mesh. In summary, our input dataset has $[5 \times {\tt nen} \times {\tt nBatch}]$ dimensions for each node encoder in a 2D problem with ${\tt nen}$ number of nodes and ${\tt nBatch}$ number of training experiments, where $5$ is the dimension of $u_x$, $u_y$, mean$(u_x)$, mean$(u_y)$, and $\Gamma_u$ flag (belongs or does not belong to the Dirichlet boundary), for each node. Therefore, following this procedure for a 3D problem we will consider an input of $[7 \times {\tt nen} \times {\tt nBatch}]$ dimensions for each node encoder. For each edge encoder the input has and $[5 \times {\tt nEdges} \times {\tt nBatch}]$ dimensions, where we consider the edges' size $\bs{q}$ and  $q_x$, $q_y$,  norm$(q)$ mean$(u_x)$ and mean$(u_y)$, being for 3D problem $[7 \times {\tt nEdges} \times {\tt nBatch}]$, that are $q_x$, $q_y$, $q_z$,  norm$(q)$ mean$(u_x)$, mean$(u_y)$ and mean$(u_z)$.

The output of the neural network returns the parameters of a normal distribution of the parameter studied for the input node: the mean and the variance. As an example, Fig. \ref{fig:arquitectura} shows the Young's modulus obtained for the displacement field given at the input. It is important to emphasize that only the input data from a single node is required for the network input, and that the network returns the Young's modulus of that node (its distribution). During the inference phase, when all the nodes of the mesh are processed simultaneously (taking advantage of the batch size), the complete distribution of the parameter throughout the model is obtained.

Precisely because it is a node-to-node approach that explicitly incorporates graph connectivity, the model allows inference to be performed on meshes other than those used during training. This property is inherent to Graph Neural Networks (GNN) and is one of the main reasons for their versatility in computational mechanics---and in other disciplines. However, it is important to note that if only meshes with a limited range of element sizes (edges) have been processed during training, performance is likely to decline significantly when inferring on meshes with elements of a size considerably different from those processed during the training phase, as the network will have learned physical patterns strongly associated with a specific spatial scale.

This dual-output structure, combined with the variational nature of the decoder, allows the VGNN to explicitly distinguish between two fundamental types of uncertainty. On one hand, epistemic (cognitive) uncertainty is modeled by the probability distribution of the weights in the variational decoder. By treating these weights as random variables rather than fixed values, the model captures the uncertainty that comes from a lack of data or incomplete model knowledge, which can be reduced as more training samples are provided \cite{blundell2015weight}. On the other hand, statistical (aleatory) uncertainty is captured through the network's predicted variance $\sigma^2$ and the learned global sampling noise $\sigma_{\text{noise}}$. This term enables the model to characterize the inherent noise in the input displacement fields and observations, allowing the network to ``admit'' its inability to produce a deterministic answer when the input data is unreliable or the inverse problem is ill-conditioned.

\subsection{Loss Function}
Based on Bayes' theorem, 
\begin{equation}
    P(\theta|\mathcal{D}) = \frac{P(\mathcal{D}|\theta)\cdot P(\theta)}{P(\mathcal{D})} \propto P(\mathcal{D}|\theta)\cdot P(\theta),
    \label{BAYES}
\end{equation}
the posterior probability $P(\theta|\mathcal{D})$ could be calculated from the likelihood probability distribution $P(\mathcal{D}|\theta)$ and the prior probability $P(\theta)$. Therefore, $P(\theta|\mathcal{D})$ is obtained by normalizing $P(\mathcal{D}|\theta)\cdot P(\theta)$ so that it sums to 1.

$P(\mathcal{D}|\theta)$ reflects how the parameters of the neural network are updated based on the data. Therefore, it can be used to make new predictions for new data (inputs), which is known as Bayesian inference.

The denominator of Bayes' Theorem, Eq. \ref{BAYES}, is called Evidence and is a marginal distribution that measures the probability of the data according to the model, that is, it is a normalizing constant that guarantees that the Posterior distribution adds up to $1$.
Therefore, it is possible to update the certainty about the weights ($\theta$) based on the data $\mathcal{D}$,
\begin{equation}
    P(\mathcal{D}) = \int P(\mathcal{D}|\theta)\cdot P(\theta) d\theta.
    \label{BAYES02}
\end{equation}

As mentioned above, due to the difficulty of calculating the posterior distribution $P(\theta|\mathcal{D})$, we want to approximate it with a variational distribution $Q(\theta)$ that is closest to the posterior and is represented by a small set of parameters, such as means and variances of a multivariate Gaussian distribution.

Now, how do we know that we are in the variational distribution closest to the true posterior distribution if we do not know what this distribution is like? The answer is by using the Kullback-Leibler divergence between the two distributions, $D_{KL}[Q(\theta) || P(\theta|\mathcal{D})]$. The divergence $D_{KL}$ is also known as information divergence, information gain, or relative entropy, and is a non-symmetric measure of the similarity or difference between two distribution functions.

To do this, we simply need to arrive at the expression for the Evidence Lower Bound (ELBO), which is a useful lower bound of $\log(p(\mathcal{D}|\theta))$ (Log-likelihood) of the observed data,
\begin{equation}
       \underbrace{\log P(\mathcal{D})}_{\text{Evidence}} - \underbrace{D_{KL}[Q(\theta) || P(\theta|\mathcal{D})]}_{\text{Our Goal}} = \overbrace{\underbrace{\mathbb{E}_{Q}[\log P(\mathcal{D}|\theta)]}_{\text{Classification Loss}} - \underbrace{D_{KL}[Q(\theta|\mathcal{D}) || P(\theta)]}_{\text{Regularization Loss}}}^{ELBO}.
    \label{ELBO}
\end{equation}
Thus, we have converted an inference problem into an optimization problem. By optimizing ELBO, we maximize the evidence, that is, we minimize the divergence between $Q(\theta)$ and $P(\theta|\mathcal{D})$.

Training is performed by maximizing the Evidence Lower Bound (ELBO) \cite{kingma2013auto}. For a dataset $\mathcal{D}$, the loss function to be minimized consists of two parts:
\begin{equation}
    \mathcal{L} = \mathcal{L}_{\text{NLL}} + \beta \cdot \mathcal{L}_{\text{KL}},
    \label{eq:loss}
\end{equation}
where $\mathcal{L}_{\text{NLL}}$ and  $\mathcal{L}_{\text{KL}}$ are known as Negative Log-Likelihood and Kullback-Leibler loss functions, respectively, and where $\beta \in [0,1]$ minimizes the cost of each minibatch in the training process, since,
\begin{equation}
    \beta_i = \frac{2^{M-i}}{2^M - 1},
    \label{eq:beta}
\end{equation}
$\beta$ being a scaling factor for estimating the posterior distribution of weights with $\sum_i^M\beta_i = 1,$ where $M$ is the total number of minibatches.

\section{Training Methodology}\label{s3}

The basis of this proposal is stochastic training of the neural network using Variational Inference, \cite{blundell2015weight}. The aim is to approximate the true posterior distribution $P(w|\mathcal{D})$ of the weights ($w$) that best fit the data ($\mathcal{D}$). Finding this true distribution $P(w|\mathcal{D})$ can be highly difficult and costly, so another variational distribution $Q(\theta)$ parameterized by $\theta = (\mu, \sigma)$ is sought that best approximates the true $P(w|\mathcal{D})$.

To enable backpropagation through random sampling, we use the Local Reparameterization Trick proposed for variational inference in \cite{kingma2013auto}. The weights $w$ are expressed as:
\begin{equation}
    w = \mu + \log(1 + e^\rho) \odot \epsilon, \quad \epsilon \sim \mathcal{N}(0, I),
    \label{eq:pesos}
\end{equation}
that is, the standard deviation is $\sigma = \log(1 + e^\rho)$.

This reparameterization allows the use of stochastic gradient descent for variational inference, since it decouples the randomness represented by $\epsilon$ from the parameters to be learned $\theta = (\mu, \rho)$, allowing the calculation of stable gradients. 

In Section \ref{propAlgo} a set of variational parameters $\theta$ as the union of the parameters for both weights and biases across all layers of the decoder is defined. For any given layer, $\theta$ is divided into $\theta_{\text{weights}} = \{\mu_w, \rho_w\}$ and $\theta_{\text{bias}} = \{\mu_b, \rho_b\}$. Consequently, the reparameterization trick introduced in Eq. (\ref{eq:pesos}) is applied independently to obtain the sampled weights ($w_{\text{weights}}$) and biases ($w_{\text{bias}}$) used during the forward pass.

\subsection{Variational Inference and ELBO}
Since the calculation of the exact posterior distribution $P(w|\mathcal{D})$ is intractable, the Kullback-Leibler (KL) divergence between $Q(\theta)$ and $P(w|\mathcal{D})$ is minimized. As developed above and reflected in Eq. (\ref{ELBO}), this is equivalent to maximizing the Evidence Lower Bound (ELBO), or minimizing the Negative ELBO cost function, since the optimizers implemented in neural networks perform gradient descent optimally,
\begin{equation}
    ELBO \equiv\mathcal{L}(\theta) = \mathbb{E}_{Q(\theta)}[\log P(\mathcal{D}|\theta)] - D_{KL}[Q(\theta) || P(\theta)],
    \label{eq:elbo}
\end{equation}
where the first term measures the fit to the data given the model weights (likelihood) and the second term acts as a complexity-dependent regularizer (prior), measuring how much the model deviates from the initial assumptions.

Rewriting Eq. (\ref{eq:elbo}) and taking into account Eq. (\ref{eq:loss}) above, where the loss function has been defined, it is determined that,
\begin{equation}
    -ELBO = -l + \beta\cdot D_{KL}[Q(\theta) || P(\theta)]. 
    \label{eq:ELBO2}
\end{equation}

On the one hand, we want to minimize the loss function known as Negative Log-Likelihood ($\mathcal{L}_{\text{NLL}} = -l$). This loss function measures the fitting error in the data, weighted by the predicted uncertainty. In other words, it estimates the probability of the network's prediction given the actual values and sampling noise.

In this work, we have chosen to use the logarithm of a normal distribution as the probability density function, thus,
\begin{equation}
     \log\mathcal{N}(\bs{x},\mu,\sigma) = -\frac{(\bs{x} - \mu)^2}{2\sigma_i^2} - \frac{1}{2}\log(2\pi) - \log(\sigma).
\end{equation}
Therefore, we calculate Log-Likelihood as the logarithm of the probability of the data given in the model, which for a Gaussian output such as the one proposed would be equivalent to:
\begin{equation}
    l = \mathbb{E}_{Q(\theta)}[\log P(\mathcal{D}|\theta)]  \approx \sum_{i} \log P(\mathcal{D}|\theta_i) = \sum_{i}\left( -\frac{||\bs{y}_i - \vect{\mu}_i||^2}{2\vect{\sigma}_i^2} - \frac{1}{2}\log(2\pi) - \log(\vect{\sigma}_i) \right),
    \label{eq:loglikelihood}
\end{equation}
with $\bs{y}_i\in \mathcal{D}$. This term allows the neural network to reduce loss by reducing error ($\bs{y} \approx \vect{\mu}$) or increasing variance ($\vect{\sigma}$) if error is unavoidable. In other words, it allows the model to ``admit'' that it does not know the answer, which is known as statistical or random uncertainty.

The second term of Eq. (\ref{eq:ELBO2}) is the Kullback-Leibler divergence ($D_{KL}$) and is a regularizer of the weights---or latent distributions---, which helps to avoid overfitting.

In particular, the Kullback-Leibler divergence for discrete probability distributions is defined according to
\begin{alignat}{1}
    D_{KL}[Q(\theta) || P(\theta)] &= \sum_i Q(\theta_i)\log\left( \frac{Q(\theta_i)}{P(\theta_i)}\right)\nonumber =\sum_i Q(\theta_i)\left[\log\left(Q(\theta_i)\right) - \log\left(P(\theta_i)\right) \right]\\ &= \sum_i Q(\theta_i)\log\left( \frac{1}{P(\theta_i)}\right) - \sum_i Q(\theta_i)\log\left( \frac{1}{Q(\theta_i)}\right)  \\
    &= \mathcal{H}(Q,P) - \mathcal{H}(Q,Q) \cong \text{logPost} - \text{logPrior}, \nonumber
    \label{eq:DKL}
\end{alignat}
where $\mathcal{H}(Q,P)\cong\log(P(\theta)) \equiv \text{logPost}$ is known as Cross-Entropy and $\mathcal{H}(Q,Q)\cong\log(Q(\theta)) \equiv \text{logPrior}$ as entropy.

Therefore, Eq. (\ref{eq:ELBO2}) can be rewritten as
\begin{equation}
    -ELBO = -l + \beta\cdot \left( \text{logPost} - \text{logPrior} \right).
    \label{eq:ELBO3}
\end{equation}

\begin{algorithm}[ht]
\SetAlgoLined
\DontPrintSemicolon
\caption{Variational Graph NN Training}
\label{alg:vgnn_training}

\KwInput{Dataset $\mathcal{D} = \{(x_n, y_n)\}$, number of epochs $nEpoch$}
\KwInput{Prior Hyperparameters: $\pi_w, \sigma_1, \sigma_2$}
\KwOutput{Variational Parameters: $\theta^* = \{\mu, \rho\}$, $\sigma_{\text{noise}}$}
\BlankLine
\tcp{Initialization}
Initialize trainable parameters;
\BlankLine
\For{epoch $e \leftarrow 1$ \KwTo $nEpoch$}{
    Shuffle $\mathcal{D}$ and divide in mini-batches $\mathcal{B}$\;
    \For{each batch $b \in \mathcal{B}$}{
        \tcp{Forward step}
        Forward noise $\epsilon \sim \mathcal{N}(0, I)$\;
        $\sigma \leftarrow \log(1 + \exp(\rho))$\;
        $w \leftarrow \mu + \sigma \odot \epsilon$\;
        Prediction: $\hat{y} \leftarrow f_{\text{VGNN}}(x_b; w)$\;
        \BlankLine
        \tcp{ELBO Loss}
        $\mathcal{L}_{\text{data}} = \sum \log \mathcal{N}(y_b | \hat{y}, \sigma_{\text{noise}}^2)$\;
        $D_{KL} = \log Q(w|\theta) - \log P(w)$\;
        $\mathcal{L}_{\text{total}} \leftarrow \beta \cdot D_{KL} - \mathcal{L}_{\text{data}}$\;
        \BlankLine
        \tcp{Optimization (Backward)}
        Gradients computation: $\nabla_\theta \mathcal{L}, \nabla_{\sigma_{\text{noise}}} \mathcal{L}$\;
        $\theta \leftarrow \text{Adam}(\theta, \nabla_\theta)$\;
        $\sigma_{\text{noise}} \leftarrow \text{Adam}(\sigma_{\text{noise}}, \nabla_{\sigma_{\text{noise}}})$\;
    }
}
\Return $\mu, \rho, \sigma_{\text{noise}}$\;
\end{algorithm}

As reflected above in the implemented Algorithm \ref{alg:vgnn_training}, which is analogous to the one developed in \cite{blundell2015weight}, to calculate $\text{logPrior}$ we use \textit{Scale Mixture Prior} to calculate the probability $p(\theta)$. This distribution is a weighted sum (we have considered $\pi_w=\frac{1}{2}$) of two Gaussians centered at zero with different variances ($\sigma_1^2, \sigma_2^2$),
\begin{equation}
    p(\theta) = \prod_j \pi_w \mathcal{N}(\theta_j, 0, \sigma_1^2) + (1-\pi_w) \mathcal{N}(\theta_j,0, \sigma_2^2),
\end{equation}
that are considered hyperparameters of the network and that in this work have been considered learnable during the training process.

This approach encourages the calculation of sparse weights. It assigns a peaked/narrow distribution to weights that are expected to be close to zero or irrelevant, and a much wider distribution to those weights that are significant.

Therefore, the Cross-Entropy $\mathcal{H}(Q,P)\cong\log(P(\theta))$ is calculated using,
\begin{equation}
\begin{split}
    \text{logPrior} =& \sum_{all} \pi_w \log\mathcal{N}(\theta_{\text{weights}}, 0, \sigma_1^2) + (1-\pi_w) \log\mathcal{N}(\theta_{\text{weights}},0, \sigma_2^2)  \\ 
    & +\pi_w \log\mathcal{N}(\theta_{\text{bias}}, 0, \sigma_1^2) + (1-\pi_w) \log\mathcal{N}(\theta_{\text{bias}},0, \sigma_2^2).
\end{split}       
    \label{eq:logPrior}
\end{equation}
Finally, to calculate $\text{logPost}$, Gaussian Variational Posterior is implemented as a diagonal Gaussian distribution. The weights $\theta$ are obtained from the standard Normal as discussed in section \ref{s3}. Therefore, the Entropy $\mathcal{H}(Q)=\mathcal{H}(Q,Q)\cong\log(Q(\theta))$ is calculated using,
\begin{equation}
\begin{split}
    \text{logPost} =& \sum_{\text{all}} \log\mathcal{N}(\theta_{\text{weights}},\mu_{\text{weights}} + \log(1 + e^{\rho_{\text{weights}}}) \cdot\epsilon_{\text{weights}},\log(1 + e^{\rho_{\text{weights}}})) \\ +
    & \log\mathcal{N}(\theta_{\text{bias}},\mu_{\text{bias}} + \log(1 + e^{\rho_{\text{bias}}}) \cdot\epsilon_{\text{bias}},\log(1 + e^{\rho_{\text{bias}}})).
\end{split}       
    \label{eq:logPost}
\end{equation}
When maximizing the Evidence Lower Bound, a balance is desired between adjusting the data and not overfitting it.

\subsection{Proposed Algorithm}\label{propAlgo}

The algorithm implemented in MATLAB for VGNN training is described in Algorithm \ref{alg:vgnn_training}. This procedure favors robust convergence of the variational parameters and the estimation of data noise, \cite{blundell2015weight}.

To provide a clear bridge between the theoretical ELBO and its numerical implementation, the total loss $\mathcal{L}_{\text{total}}$ in Step (11) of Algorithm \ref{alg:vgnn_training} is computed as $\mathcal{L}_{\text{total}} = \beta \cdot D_{KL} - \mathcal{L}_{\text{data}}$. Here, $\mathcal{L}_{\text{data}}$ (Step (9)) represents the log-likelihood term, specifically implemented as the sum of the log-probabilities of the observed data given a Gaussian distribution, Eq. \ref{eq:loglikelihood}. This term effectively seeks to minimize a noise-weighted mean squared error. 
The $D_{KL}$ term is computed in Step (10), and it is weighted by the factor $\beta$ that serves as a complexity regularizer. This $\beta$-weighting scheme is defined in Eq. \ref{eq:beta}, ensures that the model maintains a balance between accurately fitting the displacement fields and adhering to the simplicity of the prior distributions, thus preventing overfitting and promoting sparse weight representations \cite{blundell2015weight}.

During training, the algorithm updates the weights and biases and the (global) sampling noise.  This last parameter represents the noise in the network's predictions and is a parameter that the network learns. Specifically, it will calculate the weights for each layer $\mu_{\text{weights}}$, $\rho_{\text{weights}}$, $\mu_{\text{bias}}$, $\rho_{\text{bias}}$, $\sigma_1$ and $\sigma_2$. 


Once trained, the neural network will return a sample of the weights and biases $[w_{\text{weights}},w_{\text{bias}}]$ in the variational layers of the decoder, with values $\epsilon_{\text{weights}}, \epsilon_{\text{bias}} \sim \mathcal{N}(0, 1)$, according to,
\[w_{\text{weights}} = \mu_{\text{weights}} + \log(1 + e^{\rho_{\text{weights}}})\cdot\epsilon_{\text{weights}},\]
\[w_{\text{bias}} = \mu_{\text{bias}} + \log(1 + e^{\rho_{\text{bias}}})\cdot\epsilon_{\text{bias}}.\]
Therefore, the value predicted by the network will be
\[\vect{\mu}_{\text{NN}} = \vect{X}_{\text{newData}}\cdot \vect{w}_{\text{weights}} + \vect{w}_{\text{bias}} .\]
$\epsilon_{\text{weights}}, \epsilon_{\text{bias}}$ represent independent samples of parameter-free noise. This explicit notation ensures consistency with the complexity cost calculations presented in Eqs. (\ref{eq:logPrior}) and (\ref{eq:logPost}), where the prior and posterior distributions are evaluated separately for weights and biases to encourage sparsity and robust uncertainty estimation.

\section{Numerical Examples}\label{s4}

The VGNN architecture proposed for parameter estimation in inverse problems is applied to two numerical examples already described at the beginning of this work. The framework has been fully implemented in MATLAB using its Deep Learning Toolbox and executed on NVIDIA GeForce 4090RTX.

\subsection{Estimation of the Elastic Modulus in a 2D Plate under Plane Stress}\label{ex:NASA}
The first example in which the technique developed above is applied seeks to obtain Young's modulus $E(x,y)$ at the nodes of a two-dimensional plate under plane stress, as shown in Figure \ref{NASA01}.

This is a square domain with side length $L=1.0$, and the problem is considered dimensionless without loss of generality. The Dirichlet boundary conditions in $\Gamma_u$ are homogeneous in the horizontal direction, allowing vertical displacement at that boundary, except at the origin. The Neumann boundary $\Gamma_t$ is the right side of the plate, which is subjected to a tensile load $\sigma_x = 1.5$.

\begin{figure*}[ht]
\centering
\resizebox{0.4\textwidth}{!}{
\begin{tikzpicture}[
    >=Stealth,
    force/.style={->, very thick, red!80!black},
    support/.style={circle, fill=black, inner sep=1.2pt},
    roller/.style={circle, draw=black, fill=white, inner sep=1.8pt},
    boundary/.style={thick},
    axis/.style={->, thick, black!99}
]

\coordinate (O)  at (0,0);
\coordinate (A)  at (3.5,0);
\coordinate (B)  at (3.5,3.5);
\coordinate (B1)  at (1.75,1.5);
\coordinate (C)  at (0,3.5);
\coordinate (C1)  at (4.5,1.75);

\fill[blue!8] (O) rectangle (B);
\draw[boundary] (O) rectangle (B);

\node at (0.25, 1.75) {$\Gamma_u$};
\node at (3.25, 1.75)  {$\Gamma_t$};

\draw[-{Stealth[black]}, dashed] (-0.3,0) -- (4.5,0) node[below] {$x$};
\draw[-{Stealth[black]}, dashed] (0,-0.3) -- (0,4.5) node[right] {$y$};

\draw[thick, fill=gray!25] (-0.15,-0.15) -- (0,0) -- (0.15,-0.15) -- cycle;

\foreach \y in {0.4,1.0,1.6,2.2,2.8,3.4}{
    \node[roller, fill=gray!25] at (-0.25,\y+0.1) {};
    \node[roller, fill=gray!25] at (-0.25,\y-0.1) {};
    \draw[thick, fill=gray!25] (-0.15,\y-0.15) -- (0,\y) -- (-0.15,\y+0.15) -- cycle;
}

\foreach \y in {0.4,1.0,1.6,2.2,2.8,3.4}{
    \draw[red, -{Stealth[red]}] (3.5,\y) -- ++(0.7,0);
}

\node[above=0.75cm] at (B1) {$\Omega$};
\node at (B1) {$E(x,y)$};
\node at (C1) {$\sigma_x$};

\end{tikzpicture}
}
\caption{2D Plate with a nonlinear distribution of the elastic modulus under plane stress.}
\label{NASA01}
\end{figure*}

The behavior of the plate varies for each experiment since the expression of the Elastic Module varies, following a random LogNormal distribution, and whose expression is
\begin{equation}
    E(x,y) = \alpha + \gamma \exp\left( g(x,y)  \right ),
    \label{eq:NASA_E}
\end{equation}
\begin{equation}
    g(x,y) \sim  \mathcal{GP} \left( 0,\exp \left( -\frac{||\bs{x}-\bs{x'}||^2  }{2l^2}\right)\right),
    \label{eq:NASA_g}
\end{equation}
where $\alpha = \gamma = l = 1$ have been used. As reflected in \cite{warner2020inverse}, from where the dataset was obtained, the database is published in a public repository \url{https://github.com/NASA/pigans-material-ID}.

\begin{figure*}[ht]
\centering
\includegraphics[width=0.3\linewidth]{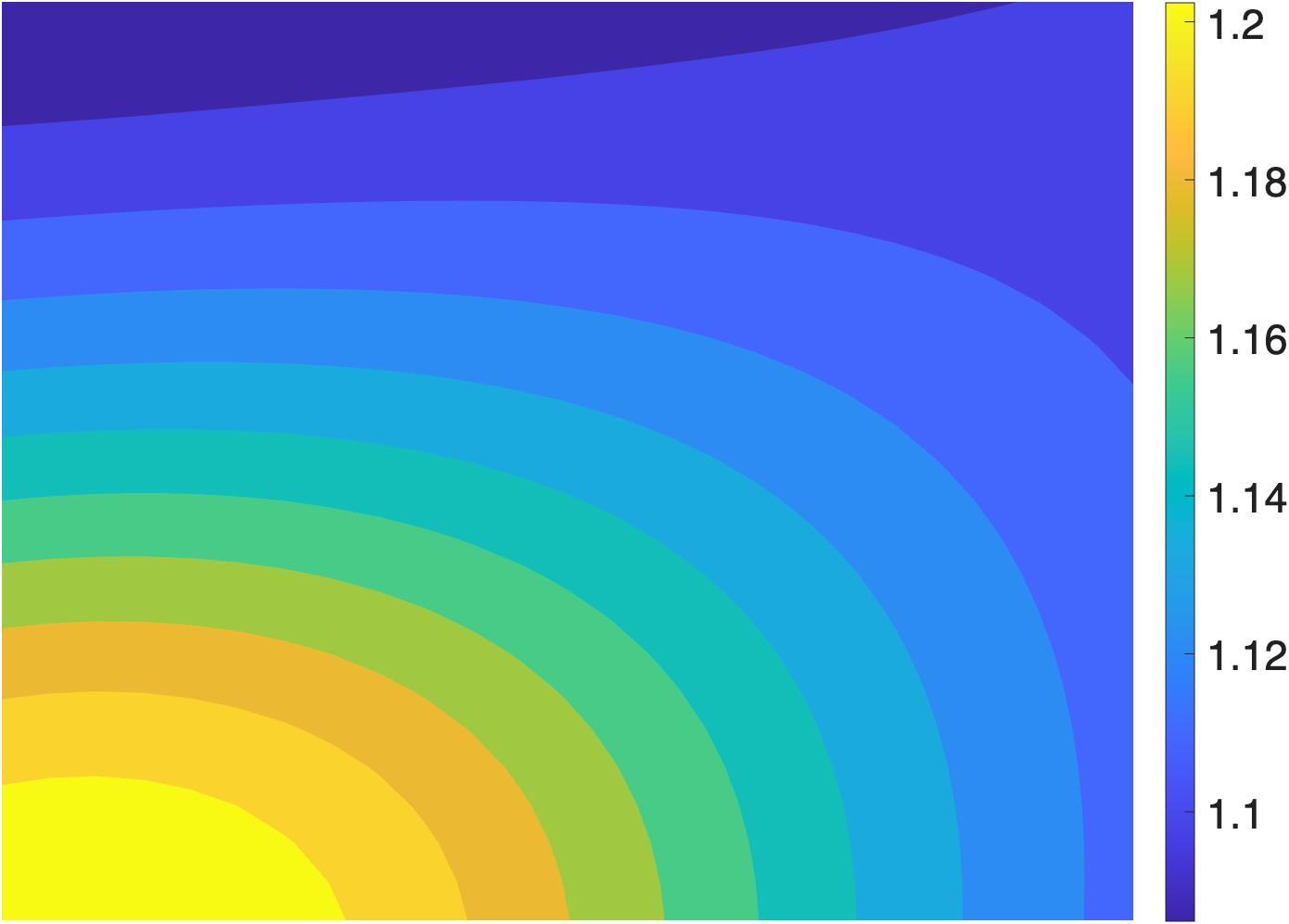}
\includegraphics[width=0.3\linewidth]{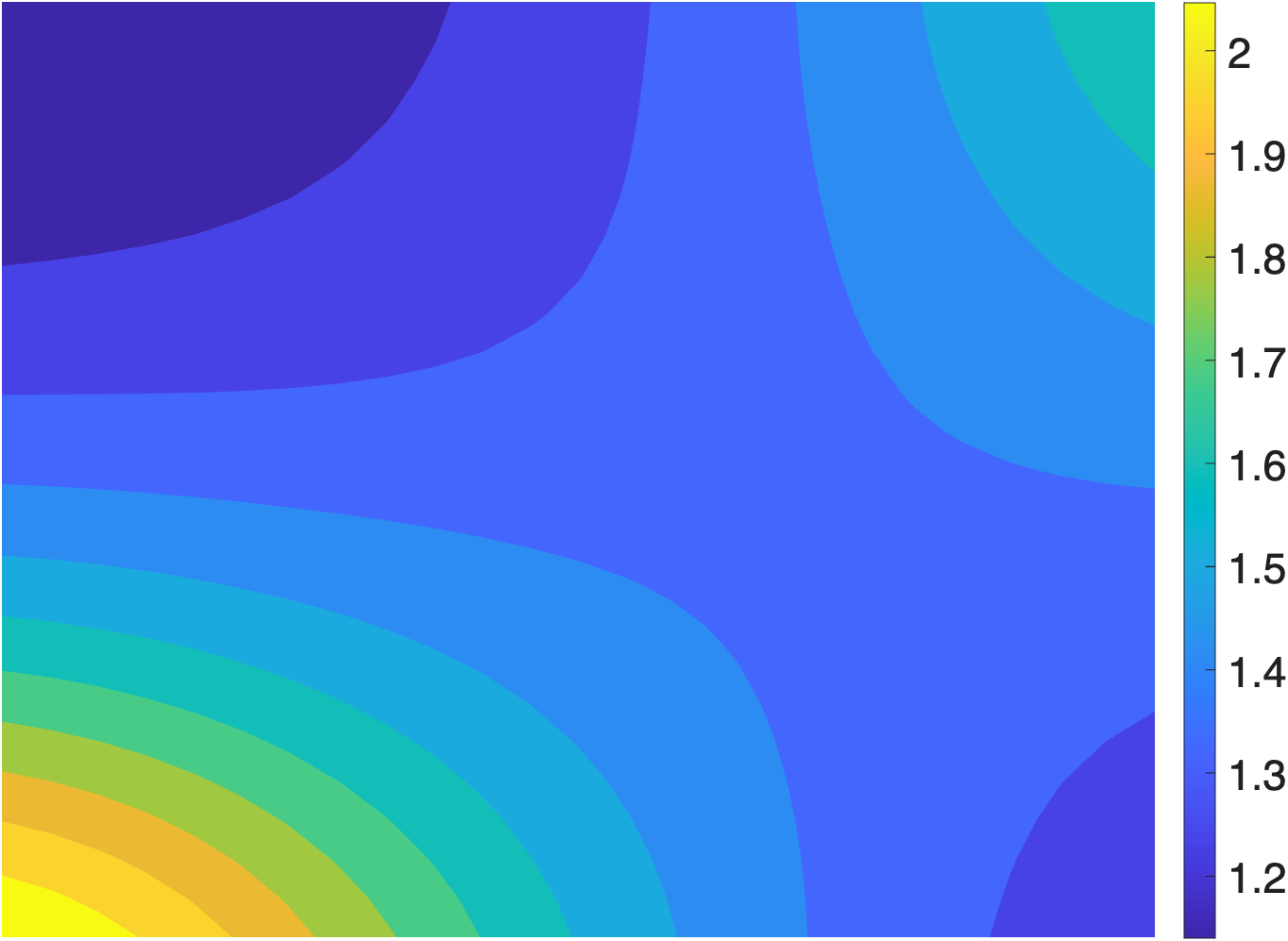}
\includegraphics[width=0.3\linewidth]{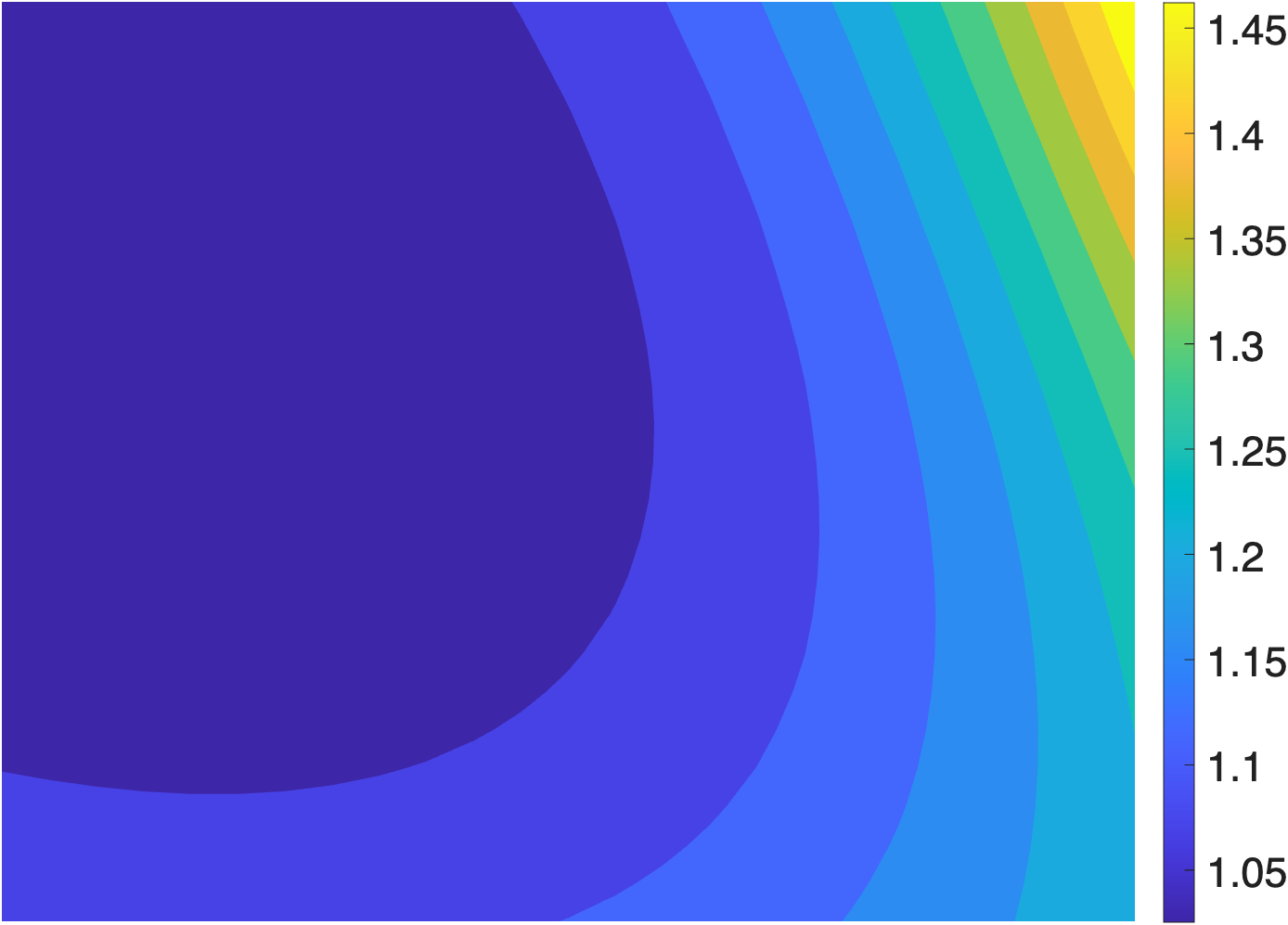}
\vspace{0.25cm}

\includegraphics[width=0.3\linewidth]{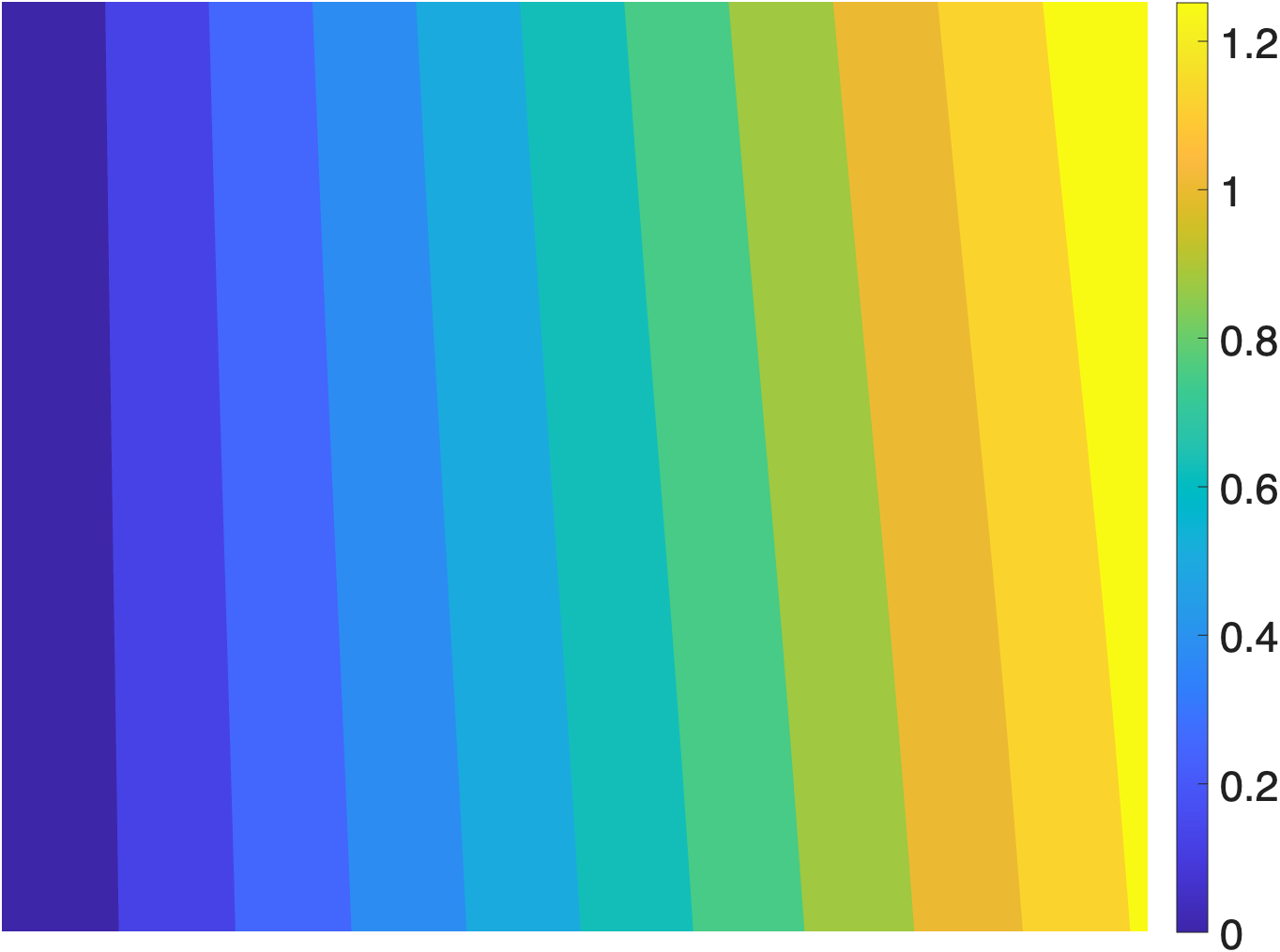}
\includegraphics[width=0.3\linewidth]{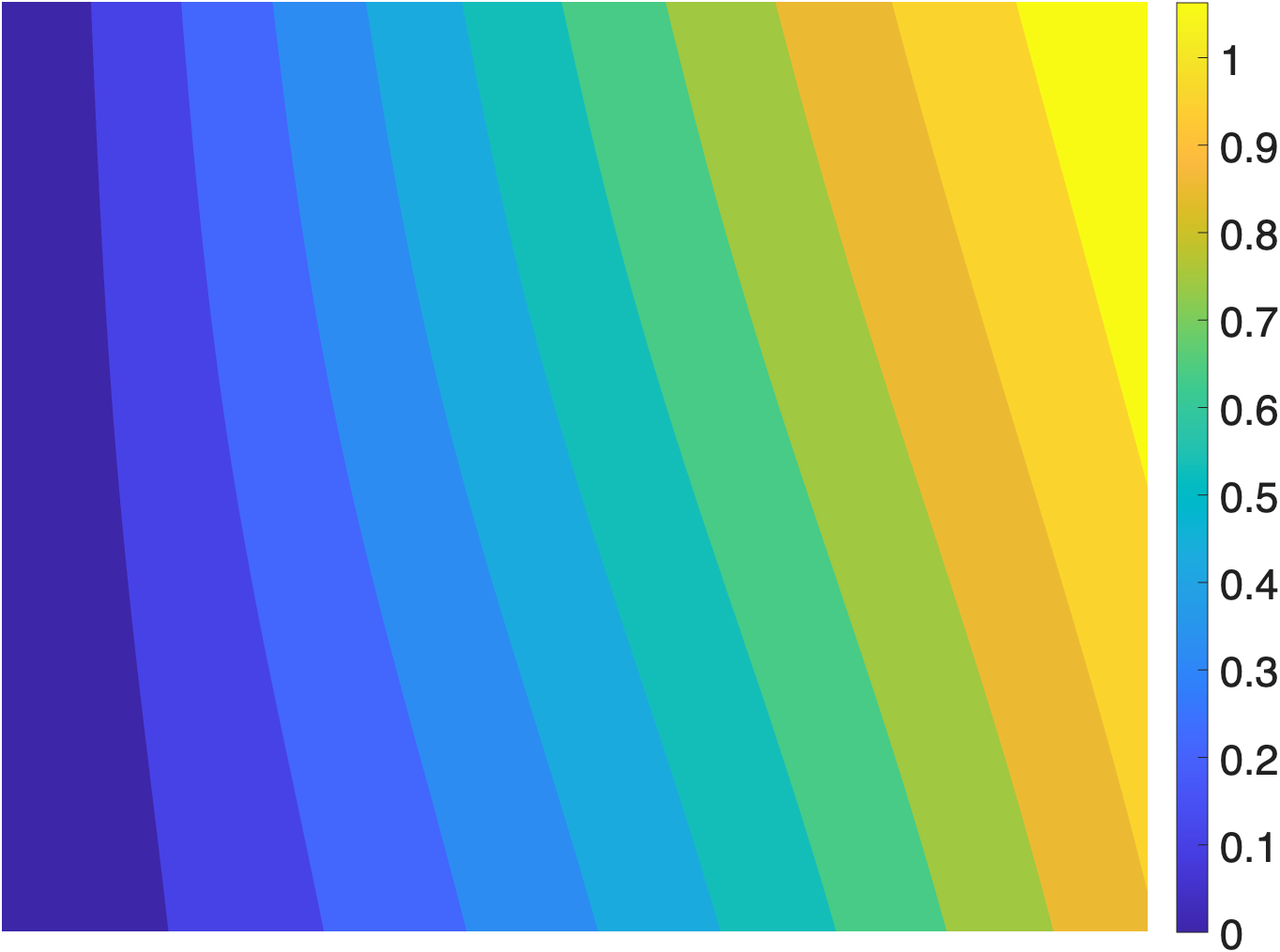}
\includegraphics[width=0.3\linewidth]{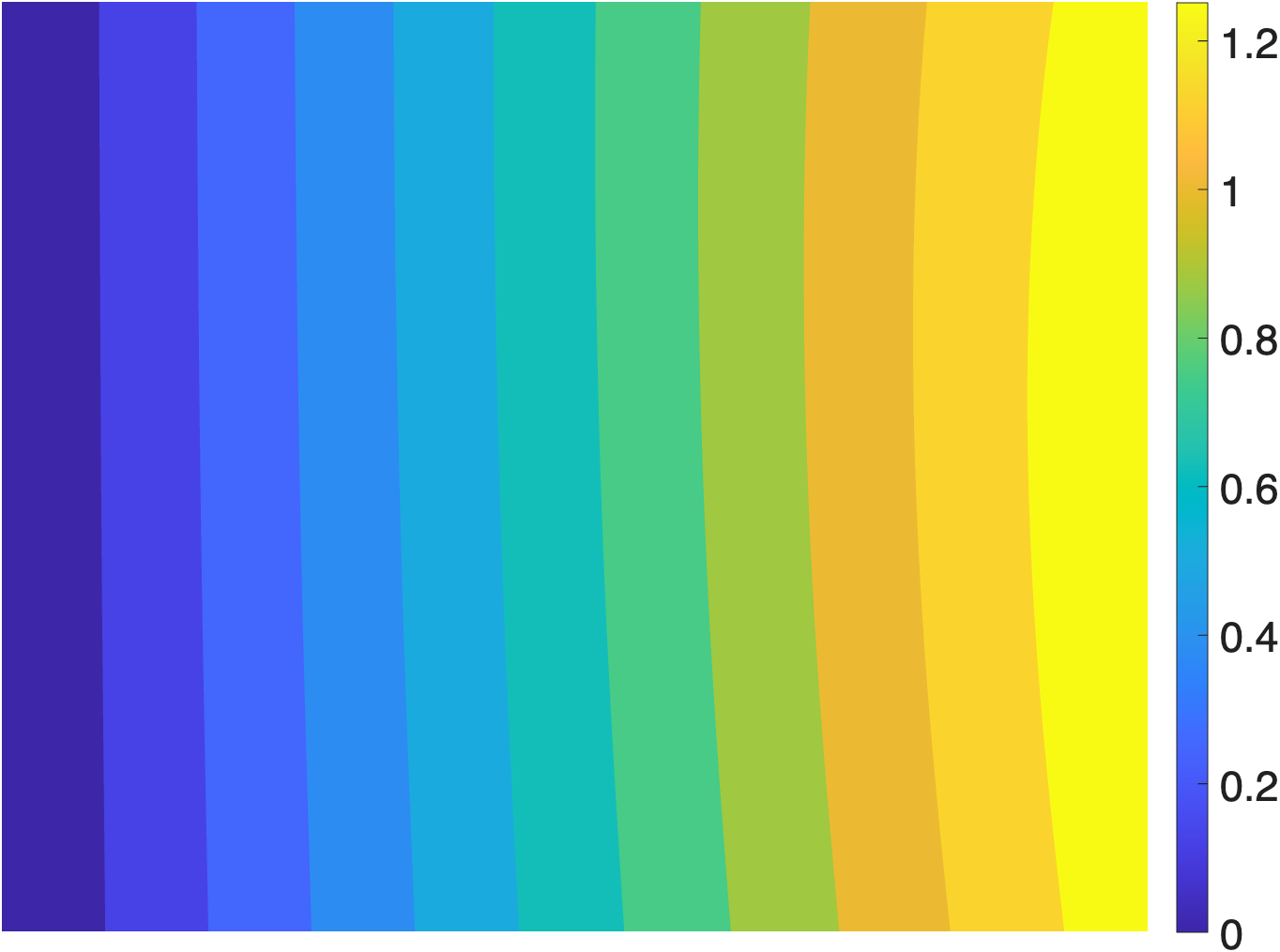}
\vspace{0.25cm}

\includegraphics[width=0.3\linewidth]{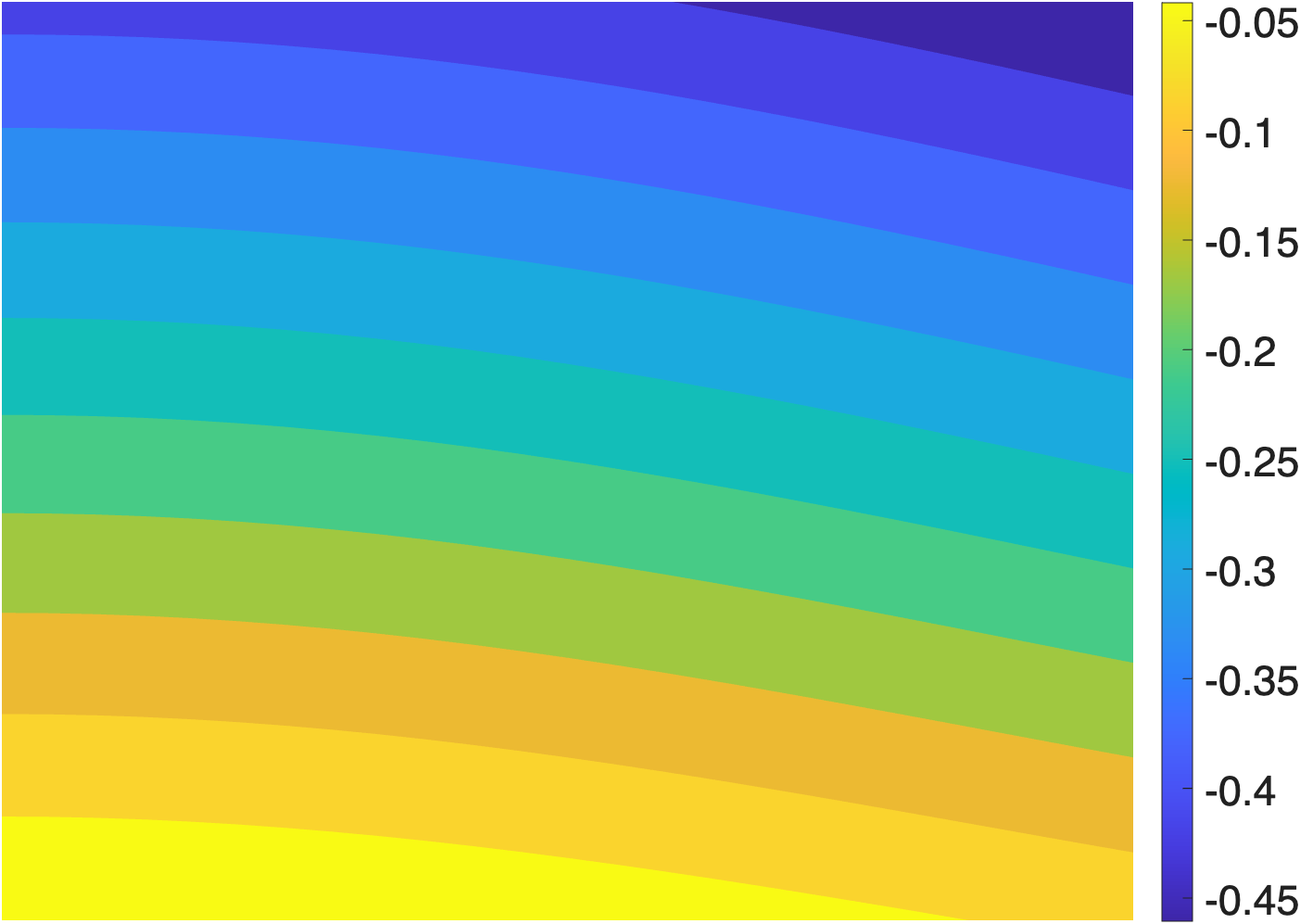}
\includegraphics[width=0.3\linewidth]{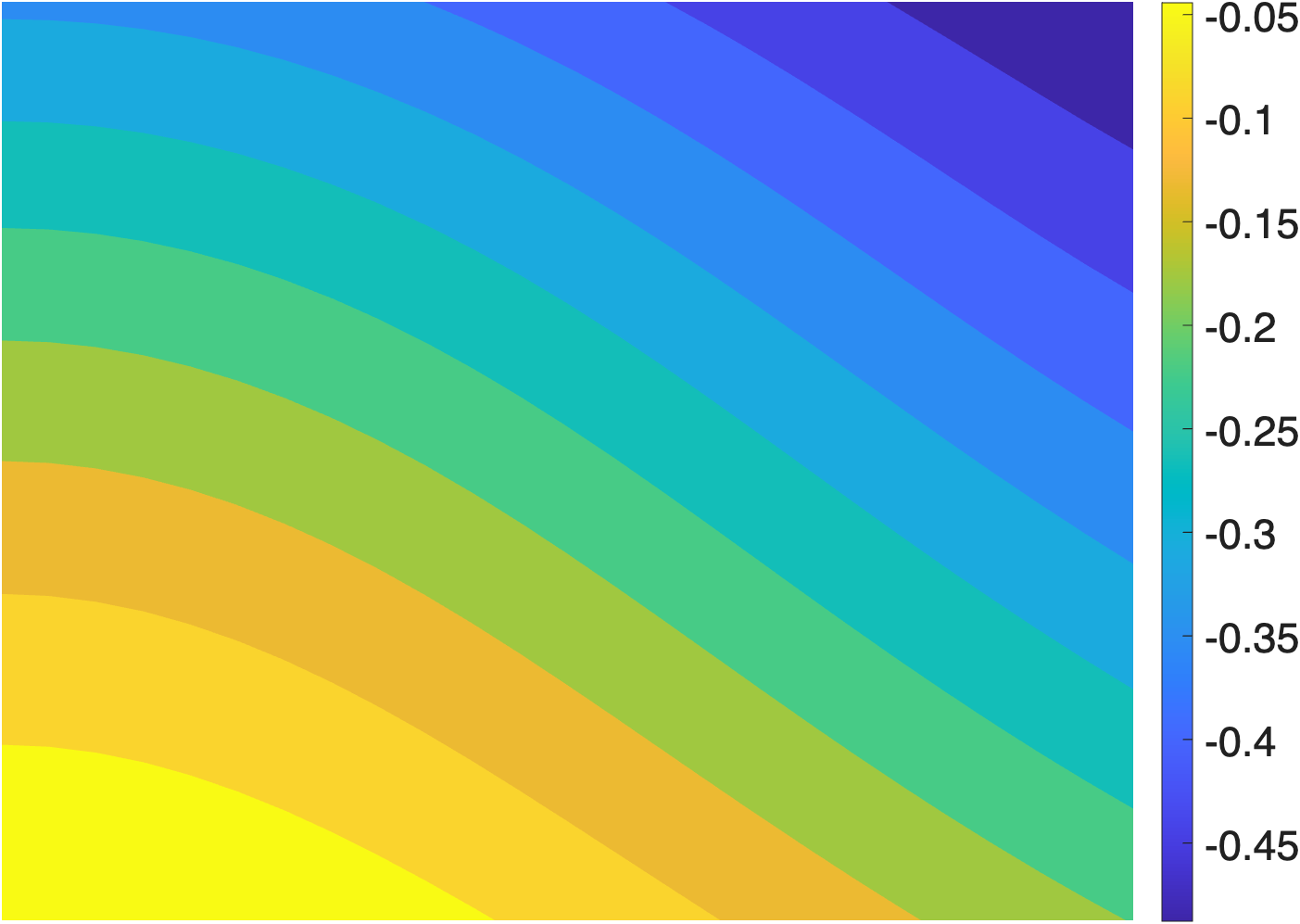}
\includegraphics[width=0.3\linewidth]{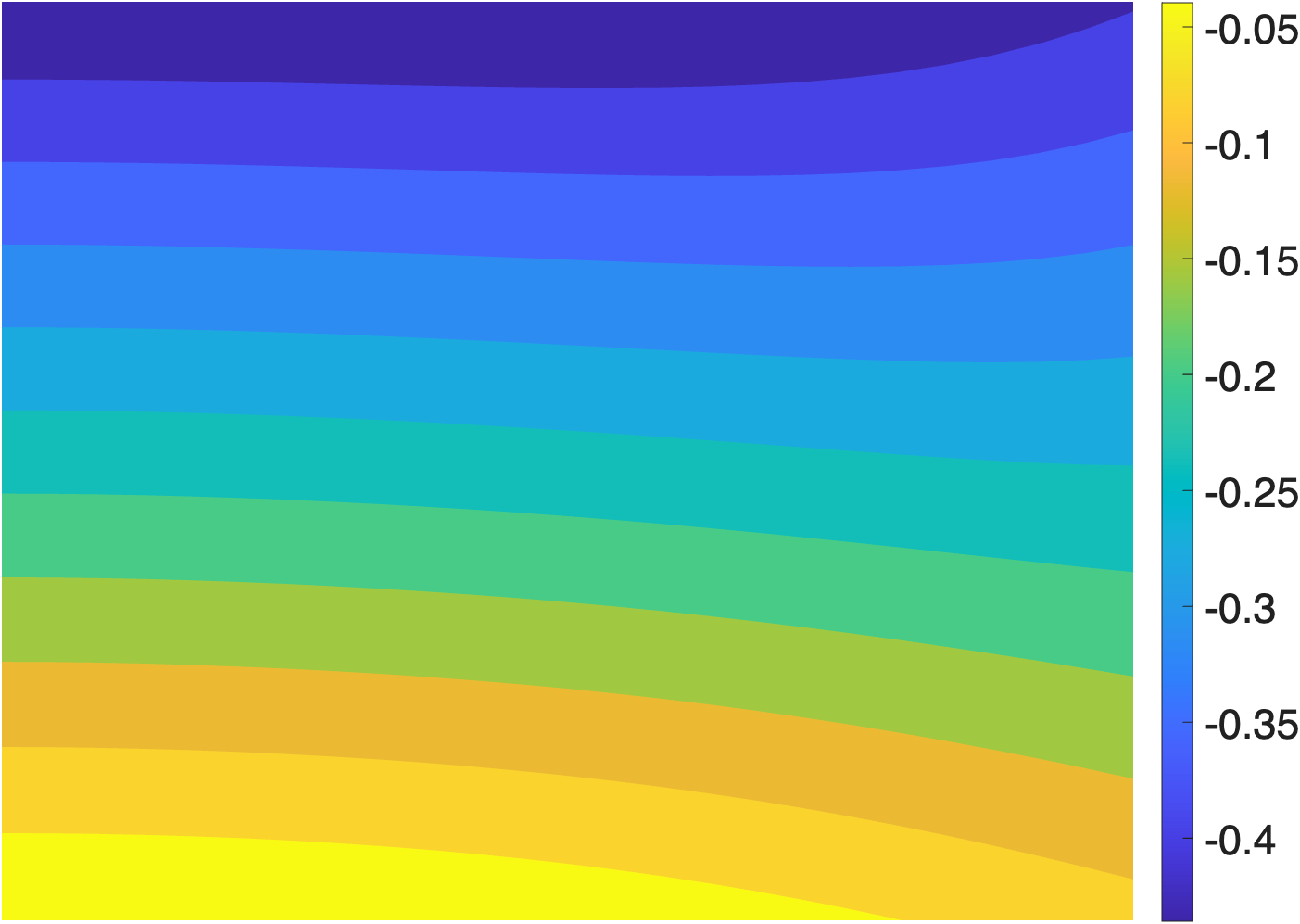}

\caption{Three examples of experiments used in neural network training. The first row represents the actual distribution of the Elastic Module, the next two rows represent the corresponding displacements --horizontal and vertical, respectively-- for the behavior of the material represented in the initial row.}
\label{NASA02}
\end{figure*}

For each randomly generated expression of the Elastic Modulus at each node in the domain ($25 \times 25$ nodes), a displacement field $\bs{u}$ has been obtained. The nodal distribution and the graph connectivity is depicted before in Fig. \ref{Grafo01}. For illustrative purposes, Fig. \ref{NASA02}  shows three simulations belonging to the dataset used in training the network.

\subsubsection{Specific training parameters}
To ensure the reproducibility of the results, the hyperparameters used and the generation process are detailed in Table \ref{tab:ex01}.

The initial dataset in the repository has a very high number of simulations. For the purposes of this work, $115$ simulations were used for training and $100$ for testing, which was considered sufficient to verify the validity of the proposed technique. All meshes are identical and the same load and boundary conditions were applied to them. Therefore, the associated graph does not vary and consists of $625$ nodes and $1825$ edges (it was considered appropriate to generate self-informed connectivity). The neural network used in this example has only 23 800 parameters.

\begin{table}[htbp]
    \centering
    \caption{Most important hyperparameters used in example \ref{ex:NASA}}
    \label{tab:ex01}    
    \begin{tabular}{l|c}
        \textbf{Hyperparameter} & \textbf{Value}\\
        \hline
        {\tt nBatch}  & $2$ \\
        Latent space dimension & $25$ neurons\\
        Message Passing steps ($m$) & $5$ \\
        $\sigma_1$ &  $\frac{1}{\exp(1)}$ \\
        $\sigma_2$ &  $\frac{1}{\exp(2)}$ \\
        Activation function & SwishLayer \\
        Number of Epochs & $4500$ \\ 
        Optimizer & ADAM \\
        Learning Rate & $10^{-3}$ \\
        Learning rate decay & $98\%$ every $700$ epochs\\
    \end{tabular}
\end{table}


It is worth noting that, although the neural network input consists of data from a single node, the entire simulation (at least) is entered in the batch direction. Therefore, a batch of $2$ means that the training is fed with $2$ complete simulations. Additionally, as this is a static problem where the neural network does not play a temporal integrator role and the input data is synthetic, it was not considered necessary to add noise to the data. In this case, considering that the input has only one field, that of displacements, and that this field is essential data for the characterization of each experiment, it was not deemed necessary to normalize the input data.

\begin{figure*}[ht]
\centering
\includegraphics[width=0.5\linewidth]{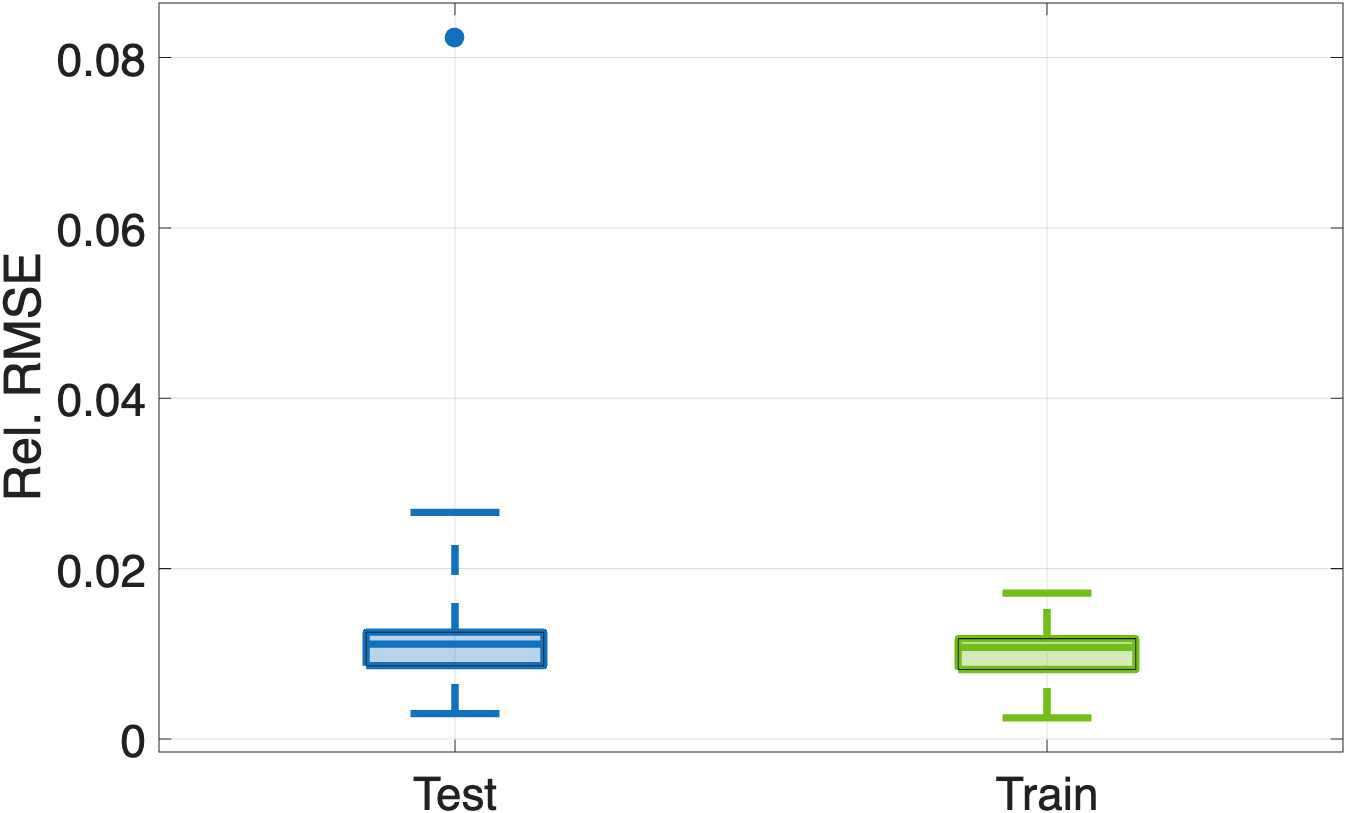} 
\caption{Relative RMSE error for training and test data for the problem of a 2D plate subjected to plane stress with nonlinear elastic modulus distributions.}
\label{NASA03}
\end{figure*}

\begin{figure*}[ht]
\centering
\includegraphics[trim = 50mm 0mm 47mm 0mm,width=0.49\linewidth]{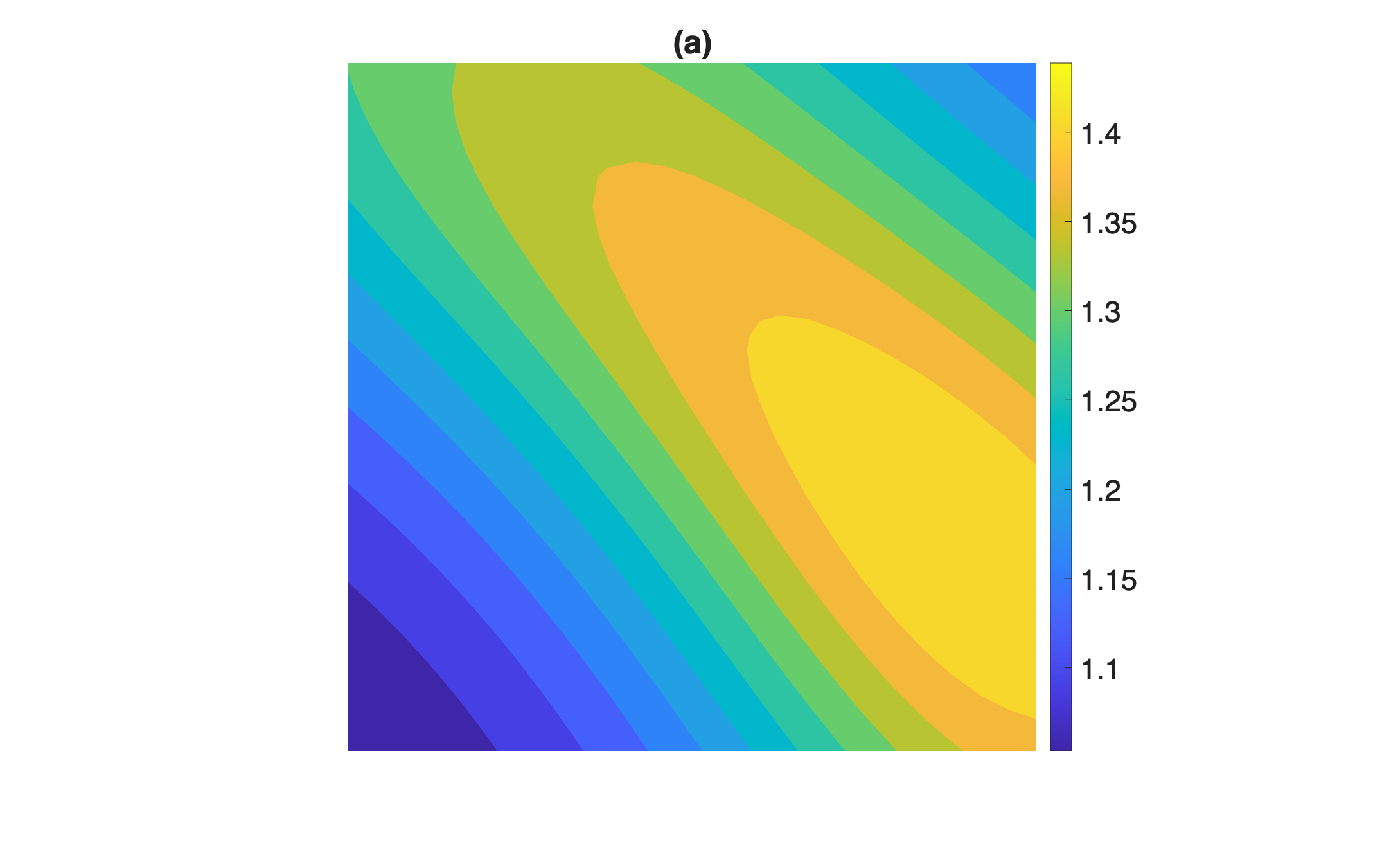} 
\includegraphics[trim = 50mm 0mm 50mm 0mm,width=0.49\linewidth]{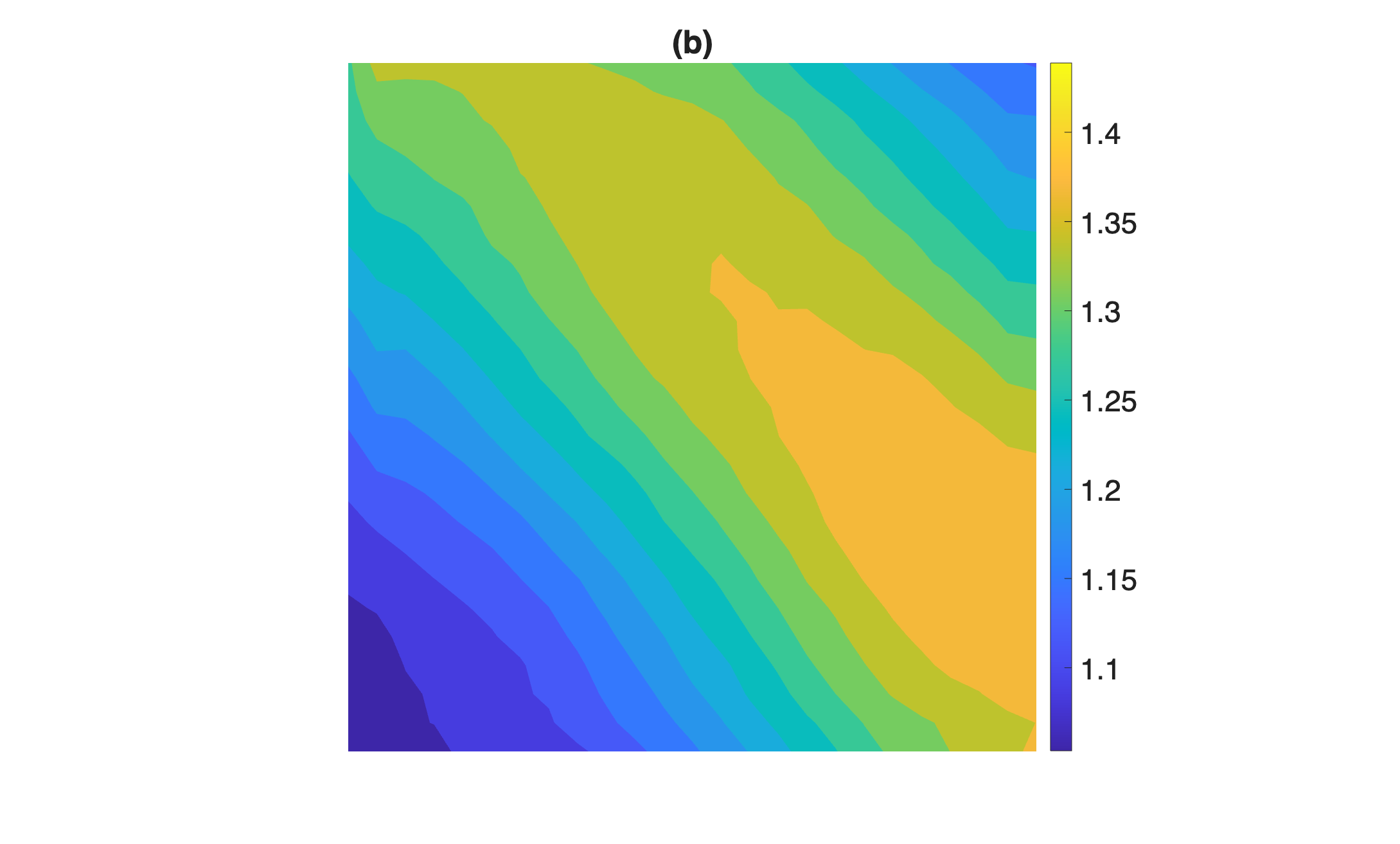} 
\includegraphics[trim = 50mm 0mm 47mm 0mm,width=0.49\linewidth]{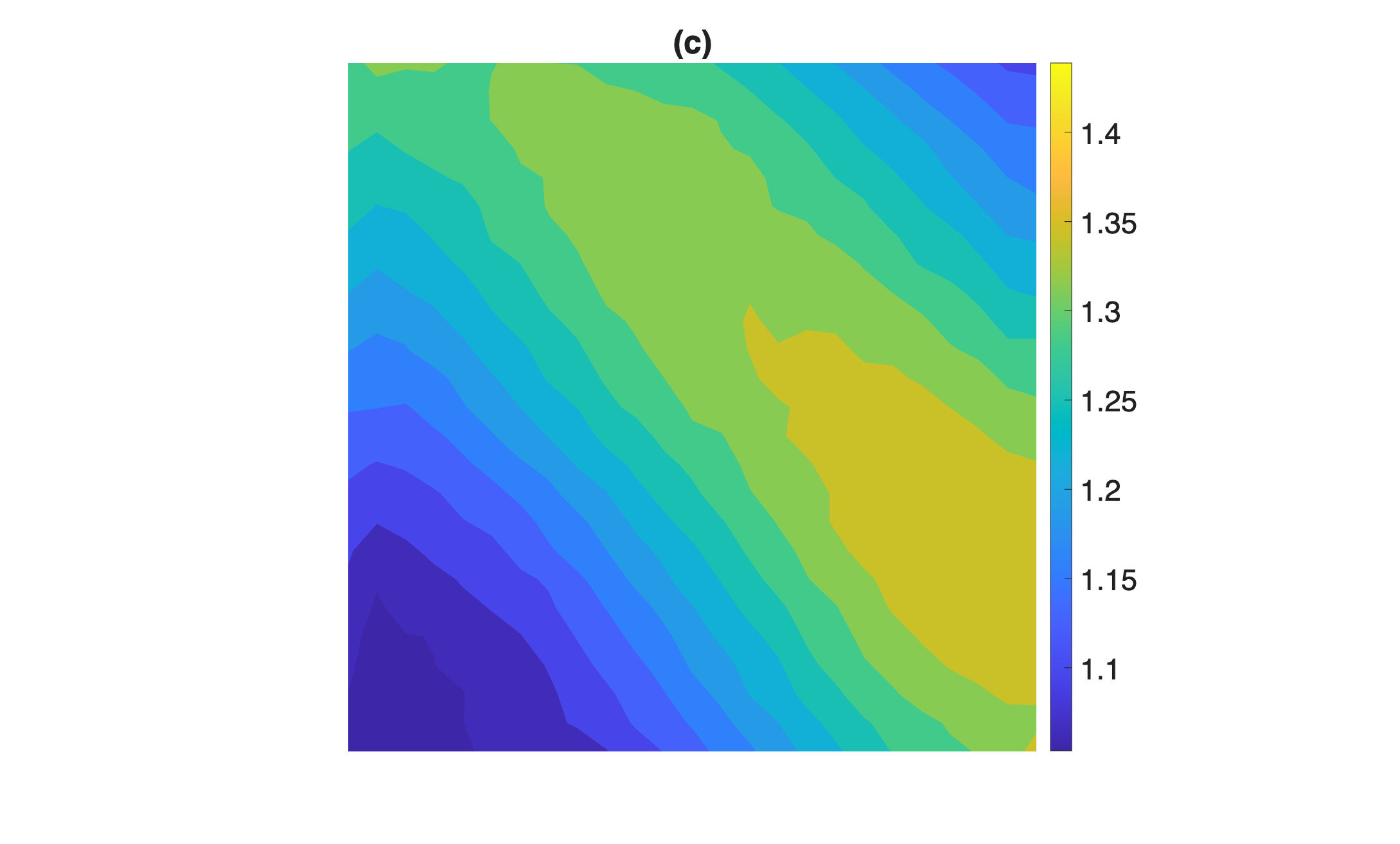} 
\includegraphics[trim = 50mm 0mm 50mm 0mm,width=0.49\linewidth]{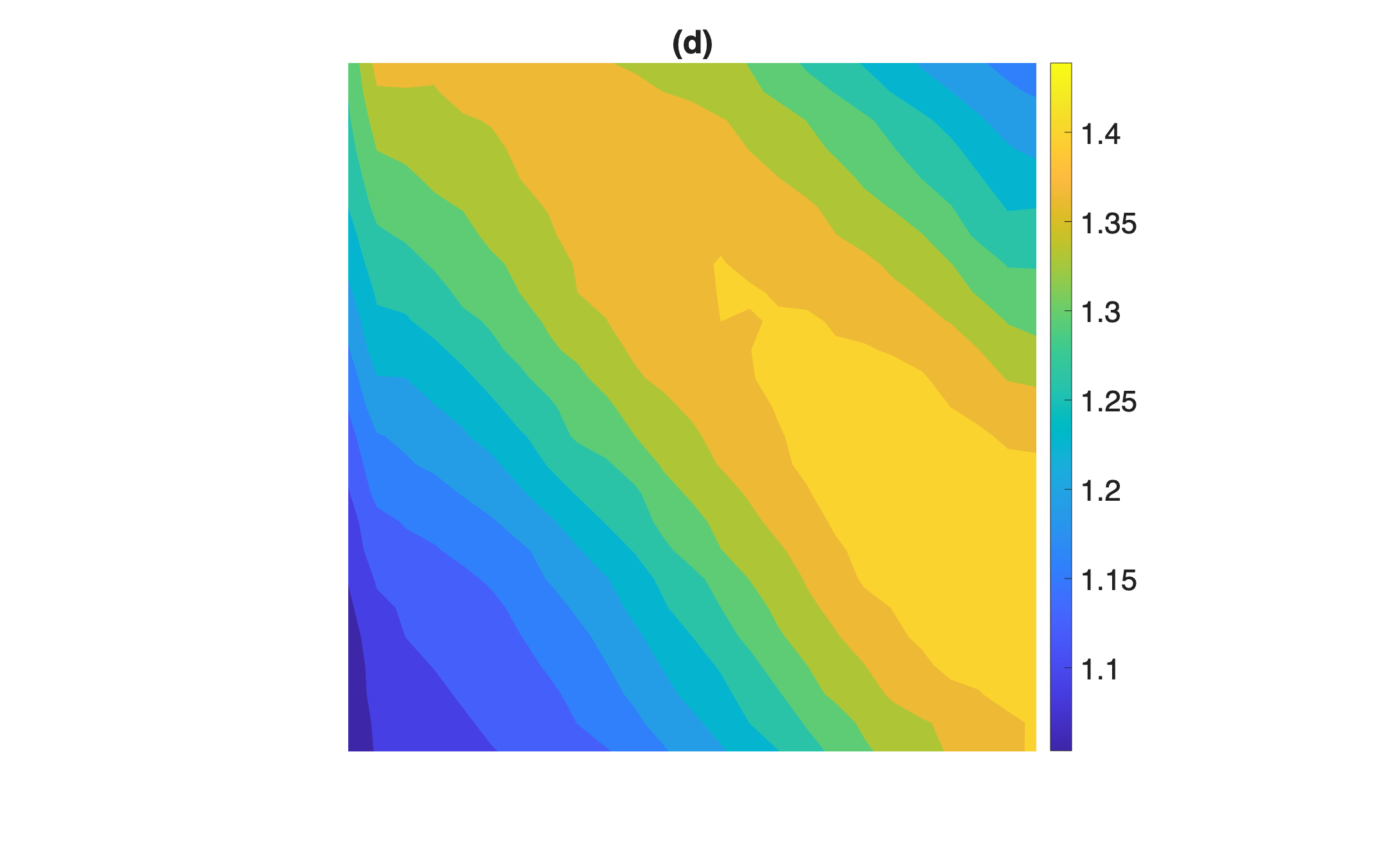} 
\caption{A first test example of the plate subjected to plane stress. (a) Distribution of the actual elastic modulus (Ground Truth), (b) Average prediction of that field, (c) Lower limit of that field, (d) Upper limit of that field.}
\label{NASA04}
\end{figure*}

\begin{figure*}[h!]
\centering
\includegraphics[trim = 50mm 0mm 47mm 0mm,width=0.49\linewidth]{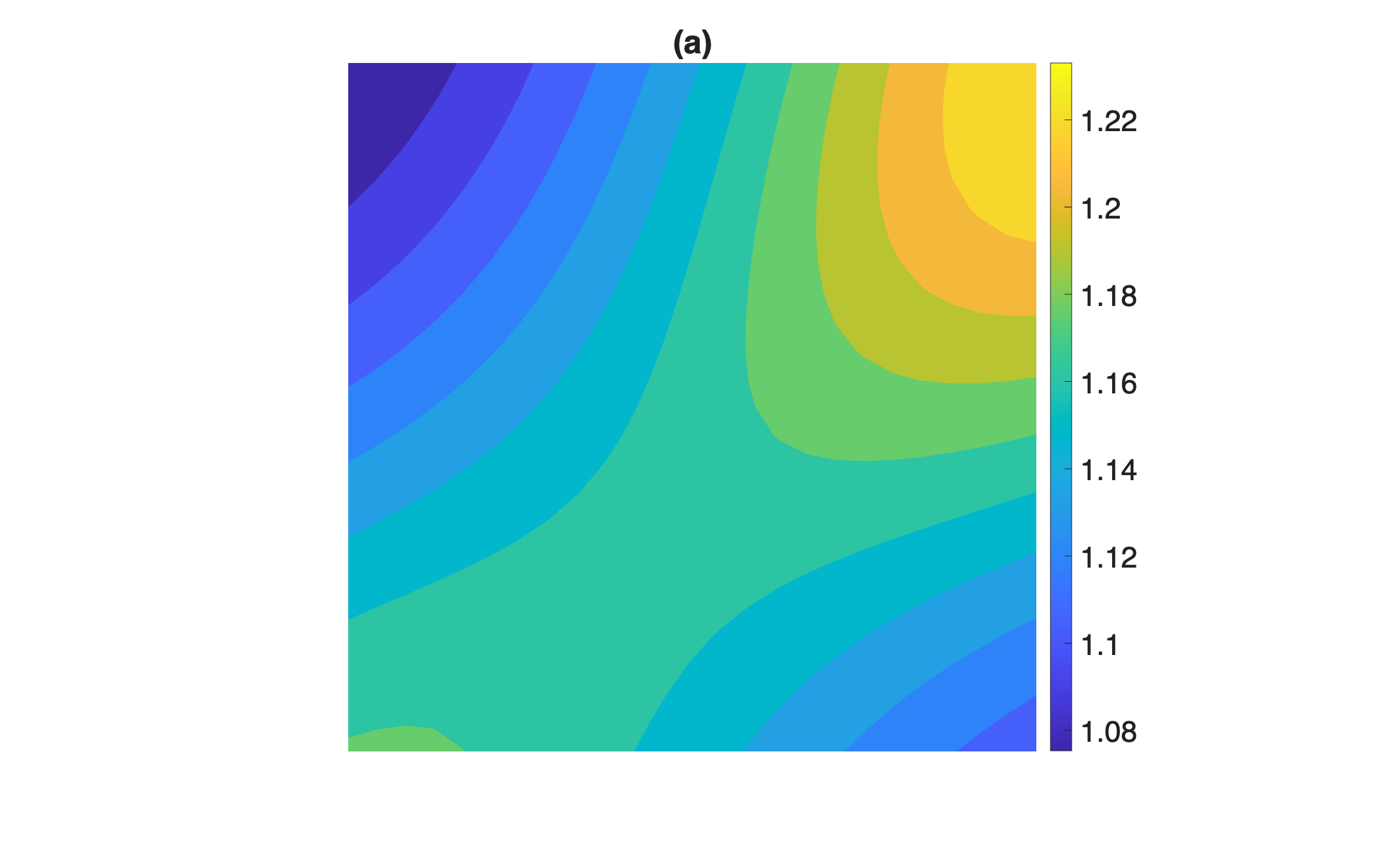} 
\includegraphics[trim = 50mm 0mm 50mm 0mm,width=0.49\linewidth]{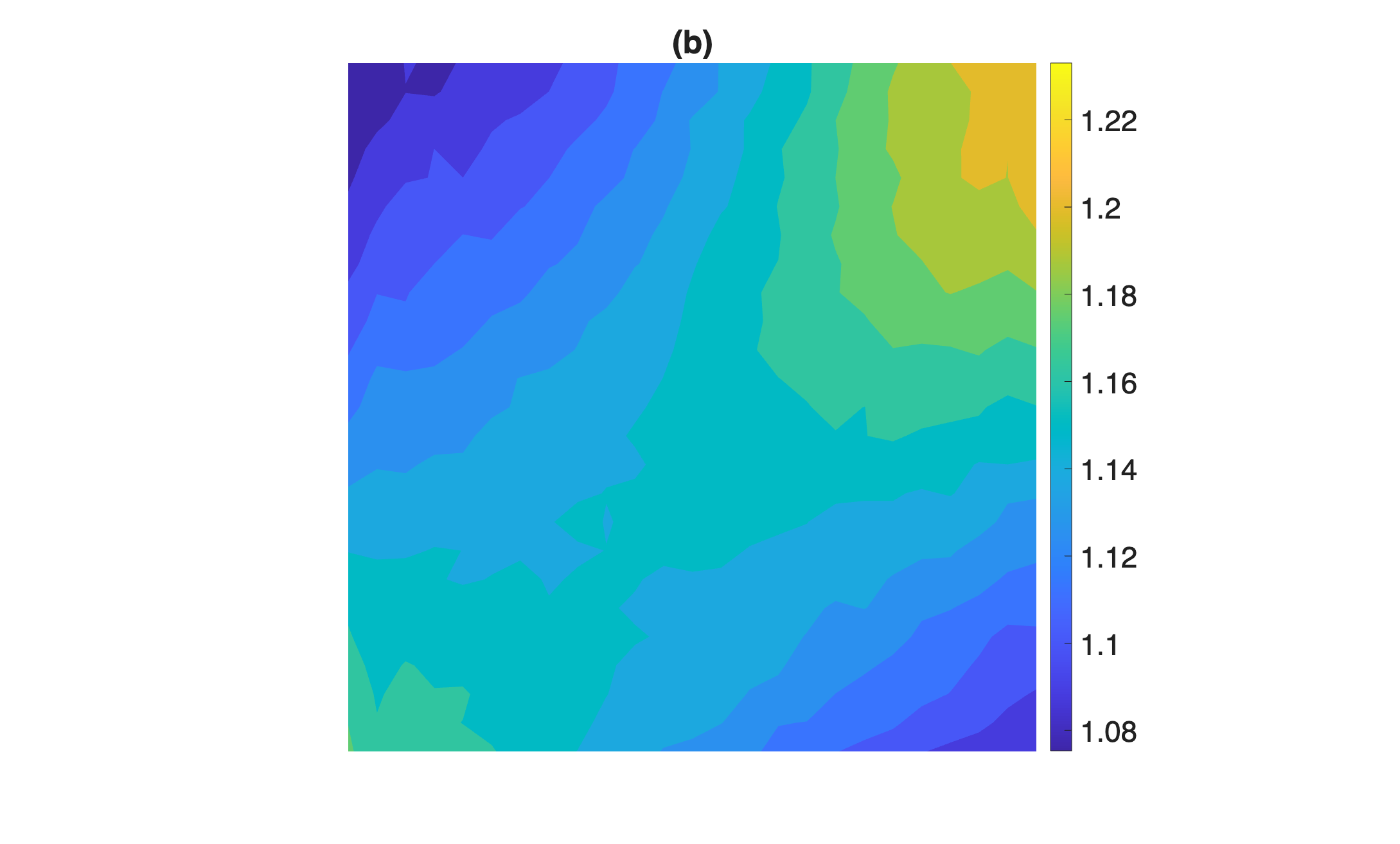} 
\includegraphics[trim = 50mm 0mm 47mm 0mm,width=0.49\linewidth]{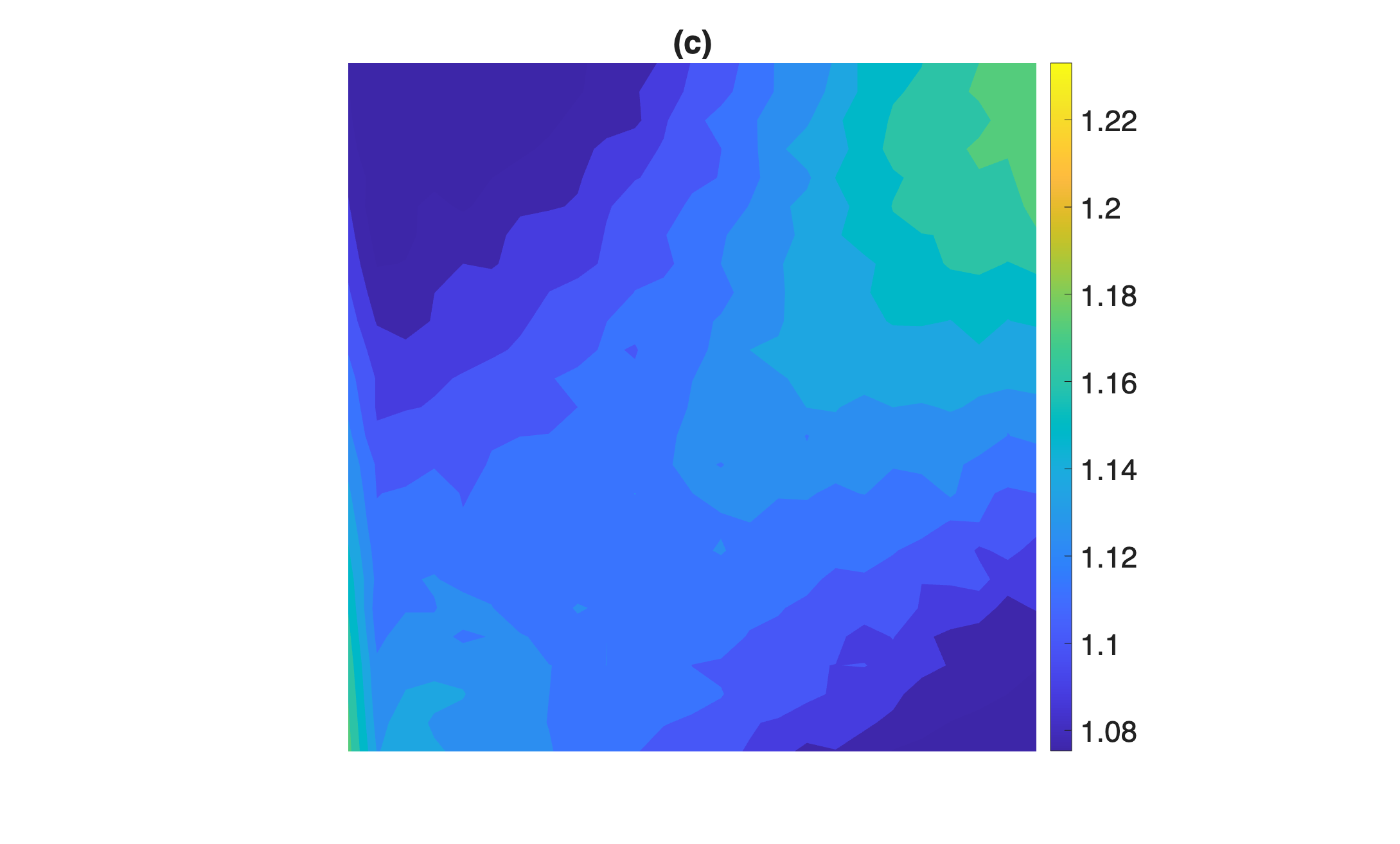} 
\includegraphics[trim = 50mm 0mm 50mm 0mm,width=0.49\linewidth]{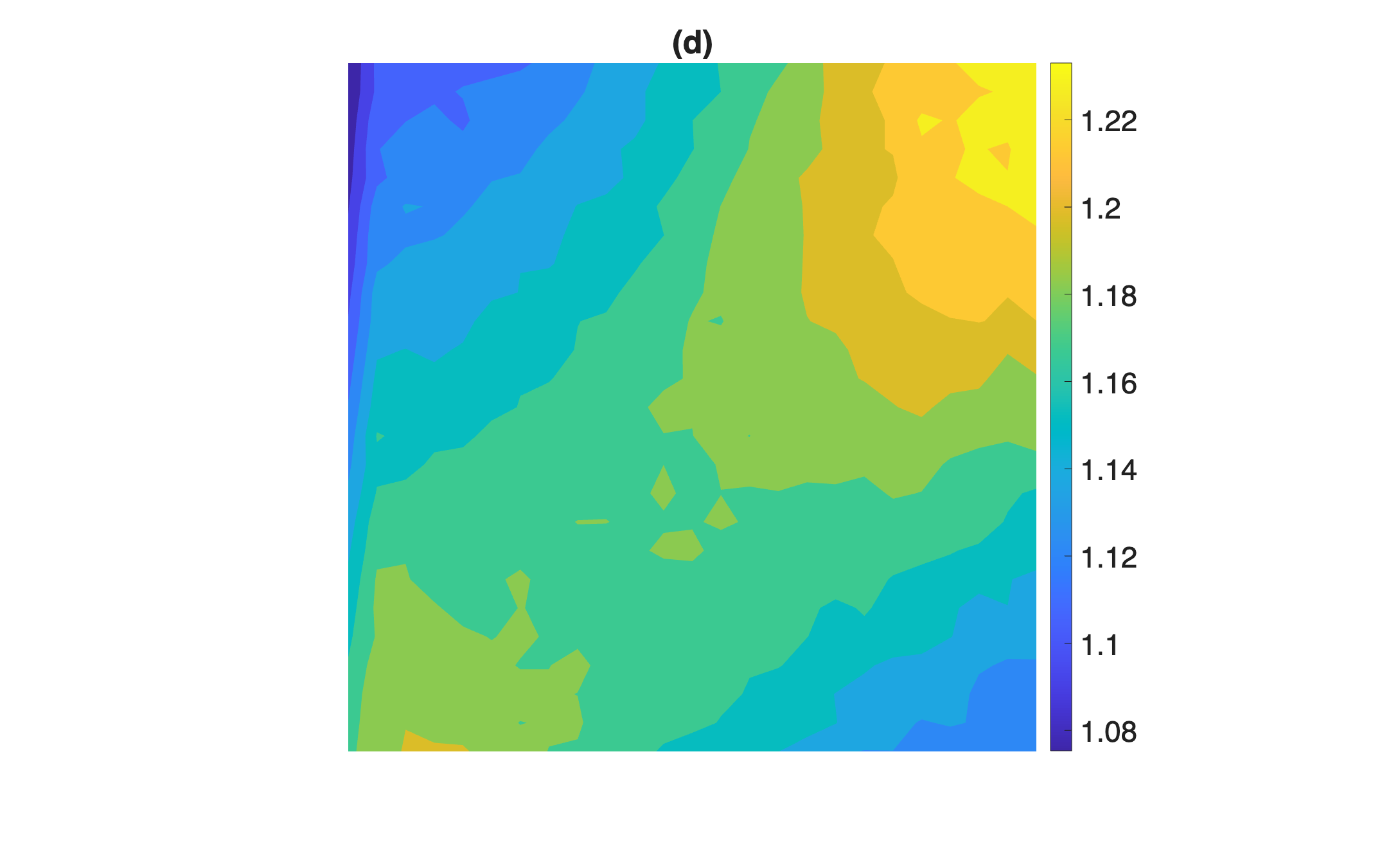} 
\caption{Second test example of the plate subjected to plane stress. (a) Distribution of the actual elastic modulus (Ground Truth), (b) Average prediction of that field, (c) Lower limit of that field, (d) Upper limit of that field.}
\label{NASA05}
\end{figure*}

\begin{figure*}[h!]
\centering
\includegraphics[trim = 50mm 0mm 47mm 0mm,width=0.49\linewidth]{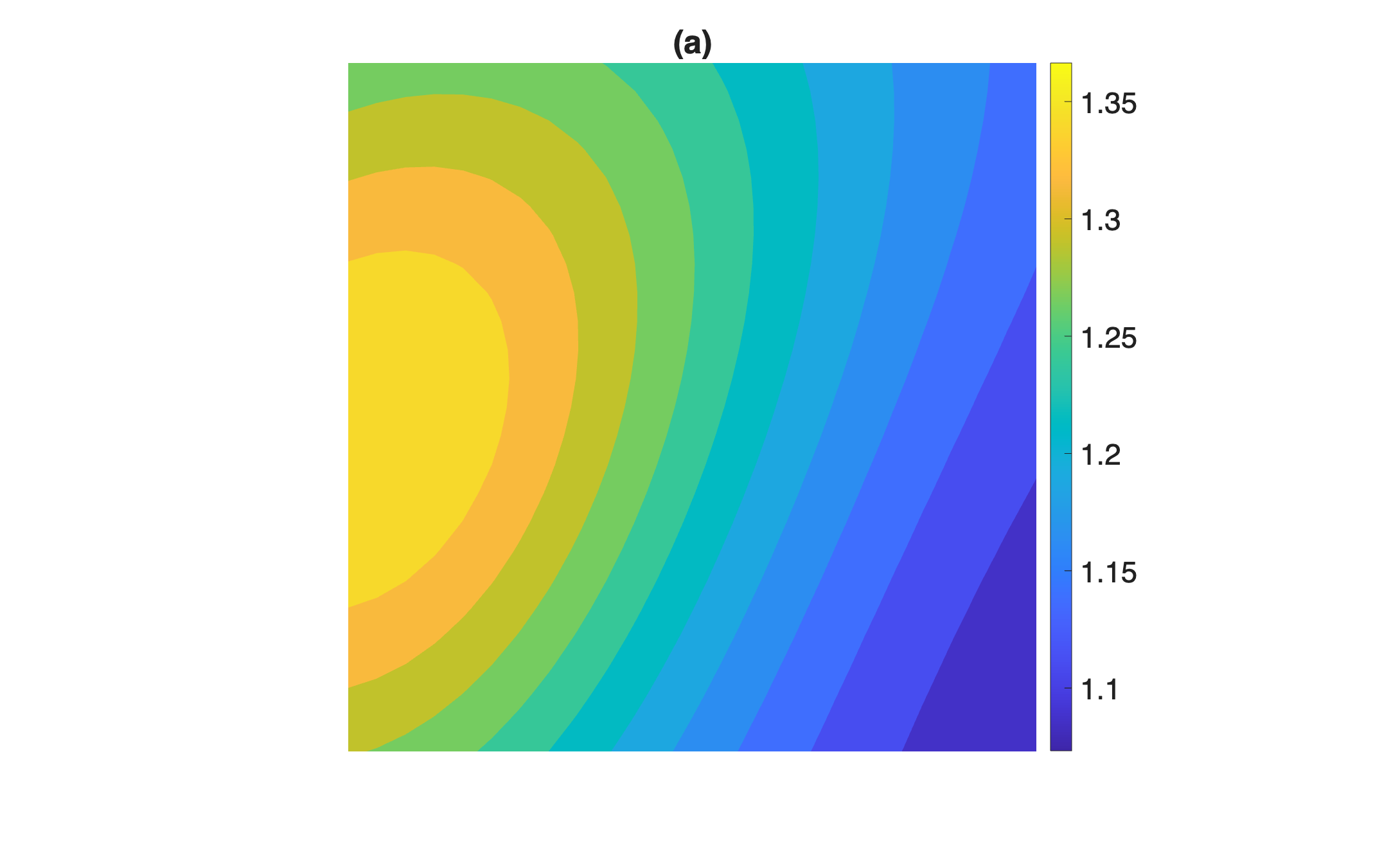} 
\includegraphics[trim = 50mm 0mm 50mm 0mm,width=0.49\linewidth]{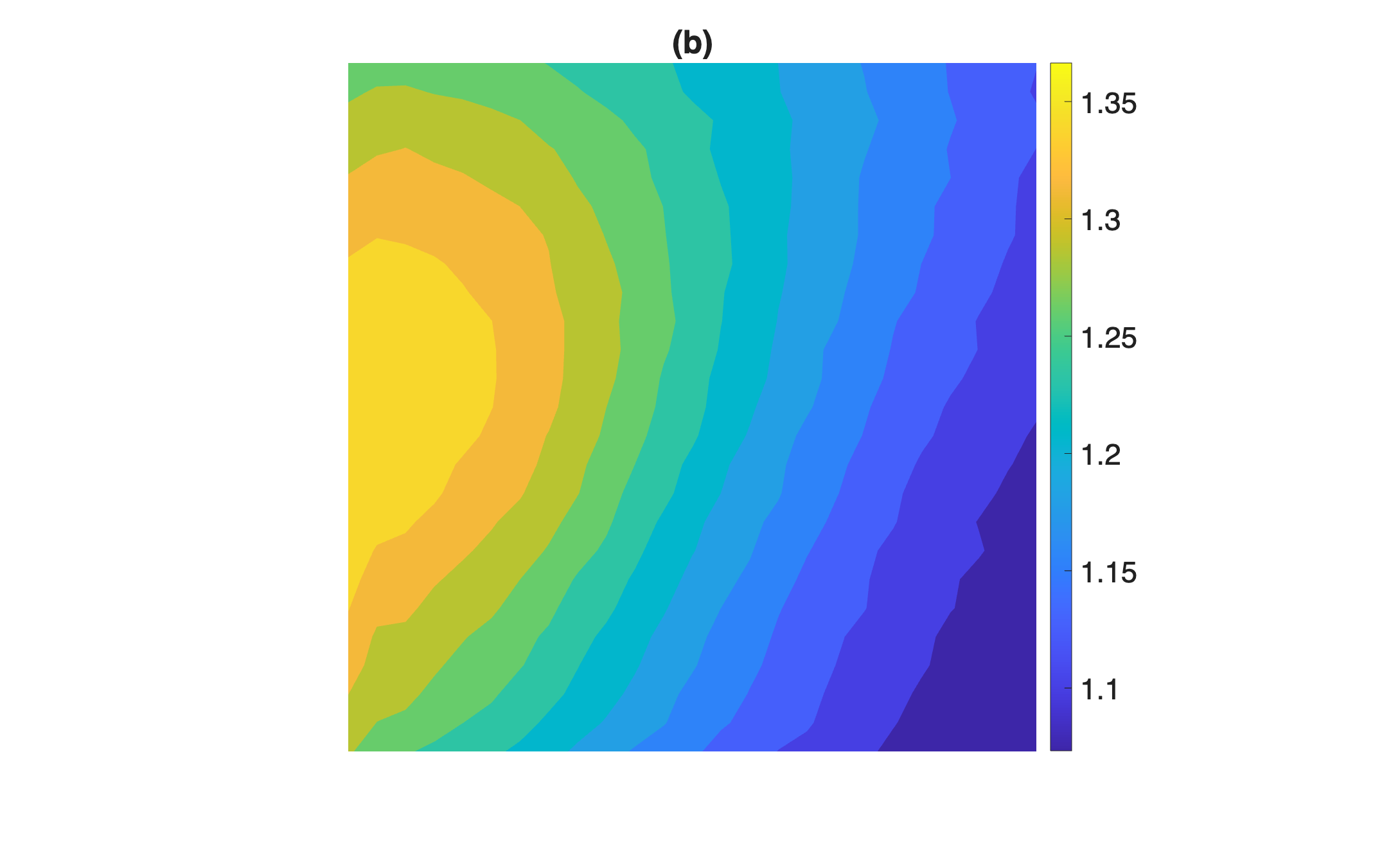} 
\includegraphics[trim = 50mm 0mm 47mm 0mm,width=0.49\linewidth]{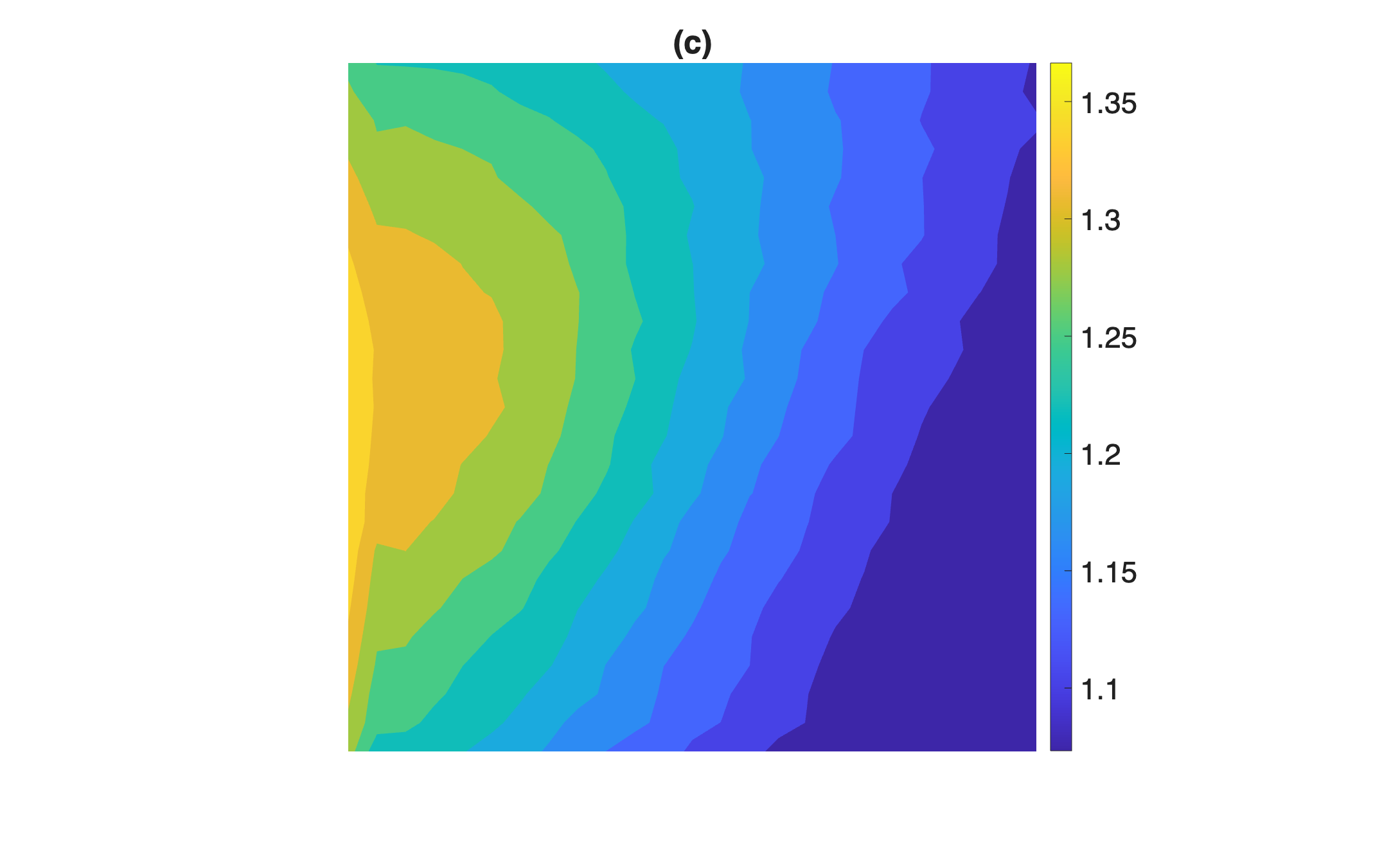} 
\includegraphics[trim = 50mm 0mm 50mm 0mm,width=0.49\linewidth]{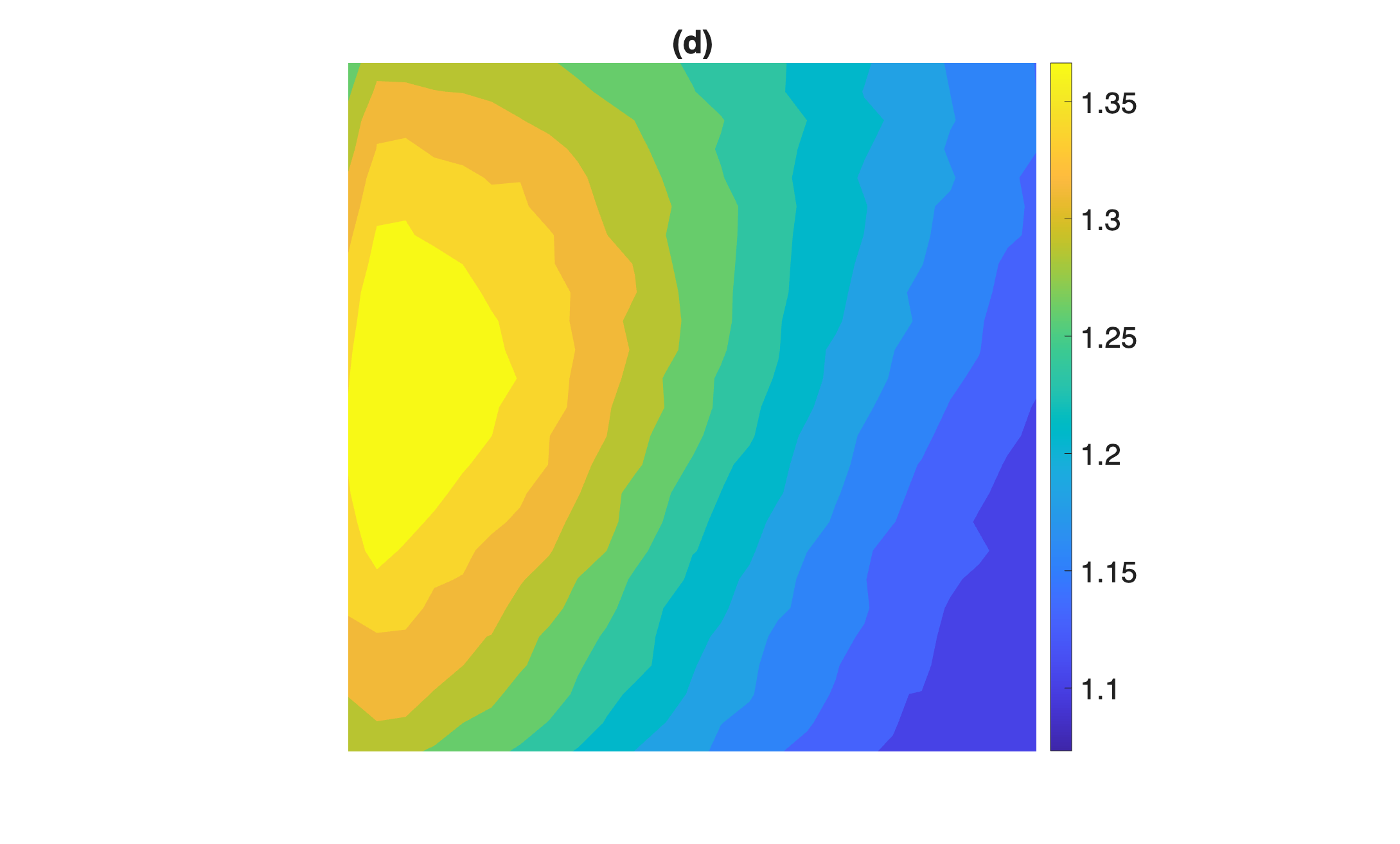} 
\caption{Third example test of the plate subjected to plane stress. (a) Distribution of the actual elastic modulus (Ground Truth), (b) Average prediction of that field, (c) Lower limit of that field, (d) Upper limit of that field.}
\label{NASA06}
\end{figure*}

\begin{figure*}[h!]
    \centering
    \includegraphics[trim = 50mm 0mm 20mm 0mm,width=0.45\textwidth]{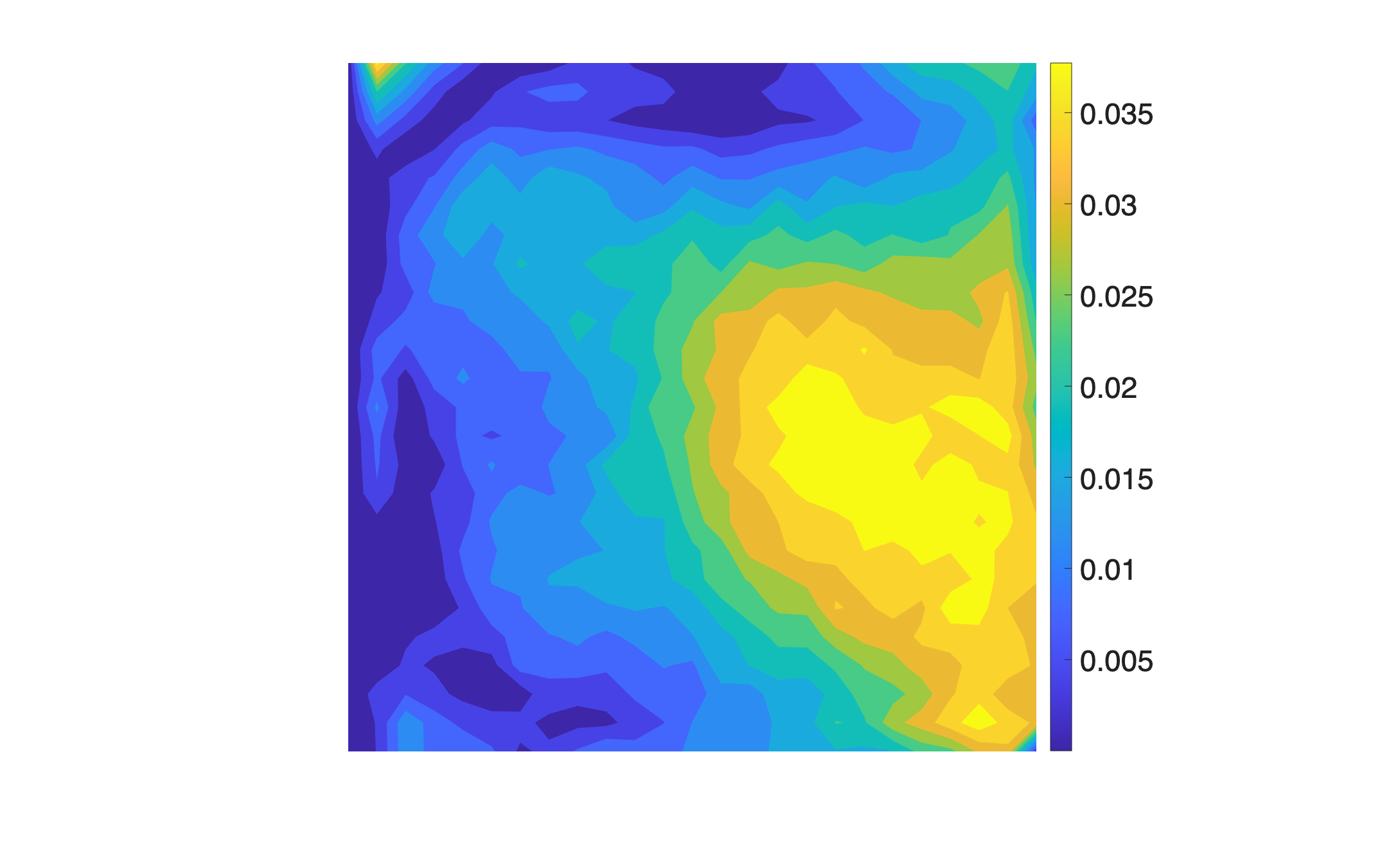}
    \includegraphics[trim = 50mm 0mm 20mm 0mm,width=0.45\textwidth]{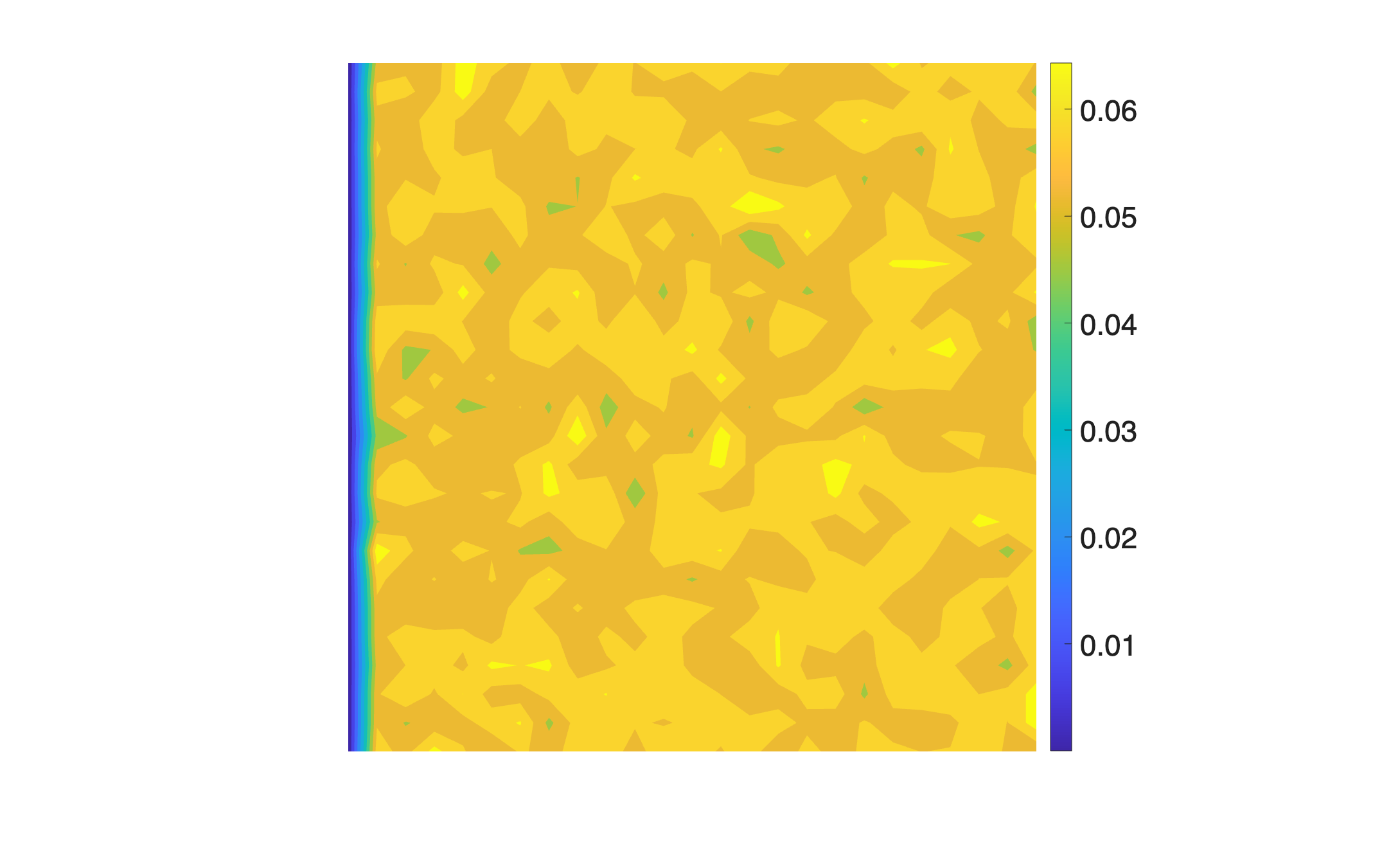}
    
    \includegraphics[trim = 50mm 0mm 20mm 0mm,width=0.45\textwidth]{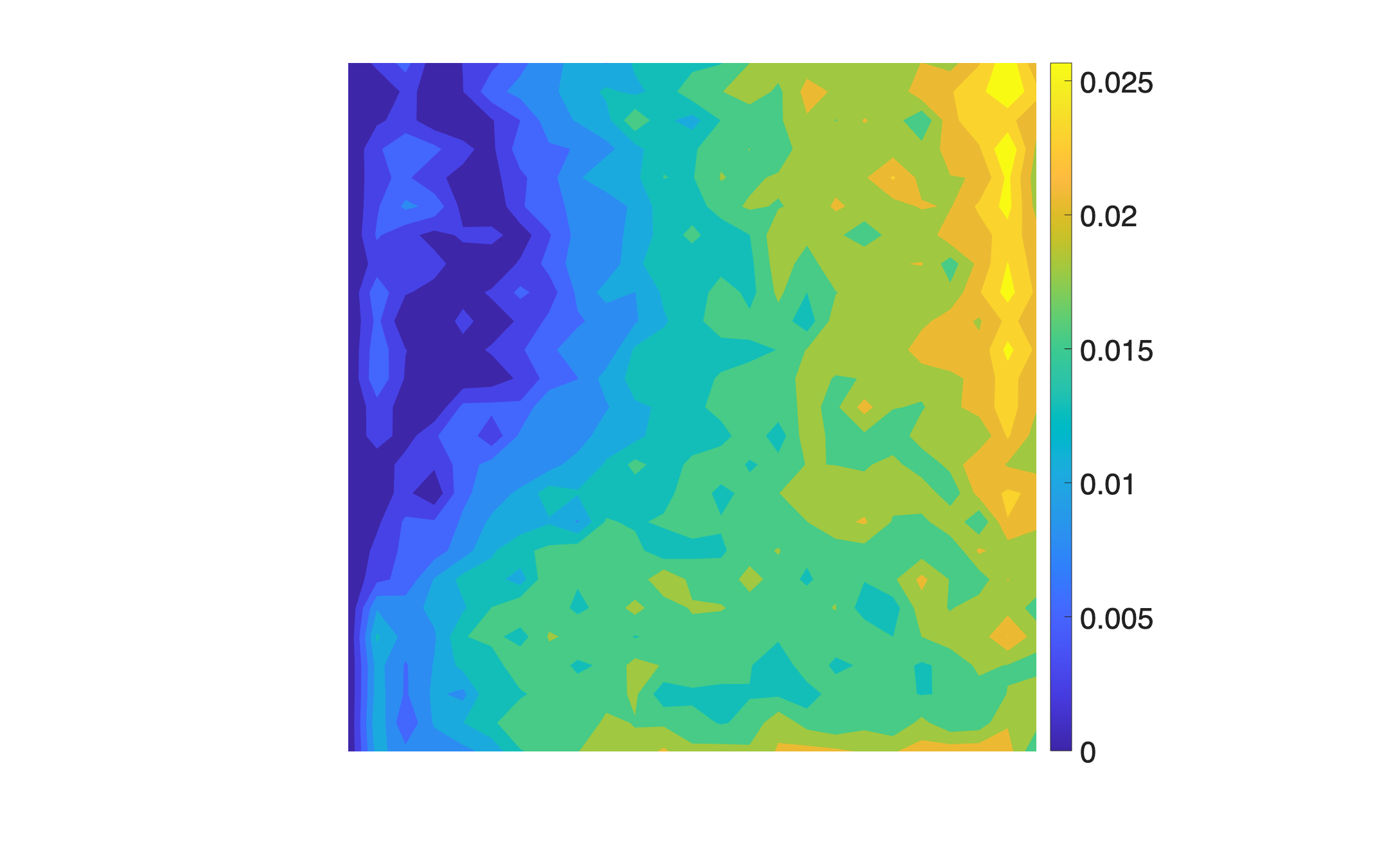}
    \includegraphics[trim = 50mm 0mm 20mm 0mm,width=0.45\textwidth]{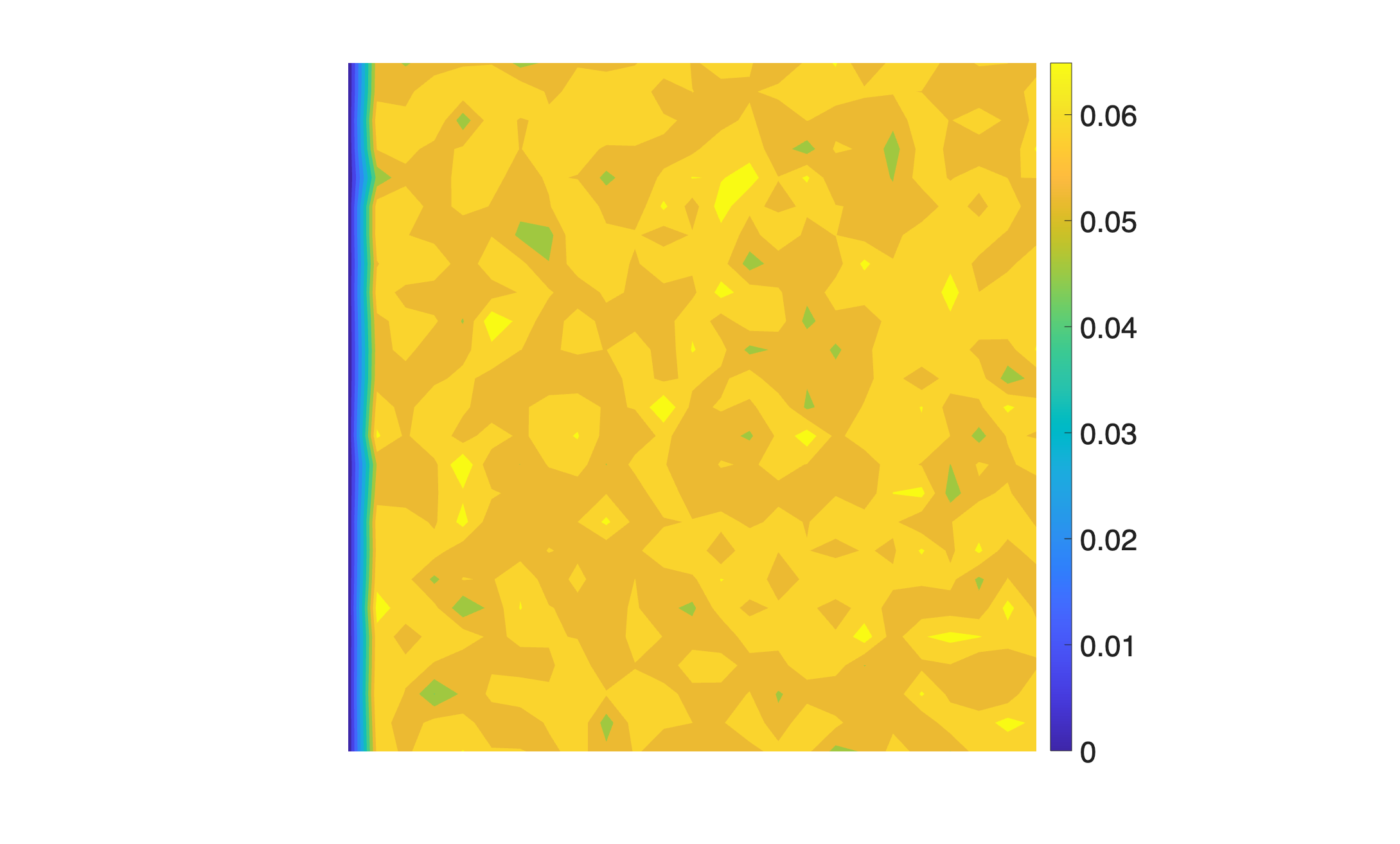}
    
    \includegraphics[trim = 50mm 0mm 20mm 0mm,width=0.45\textwidth]{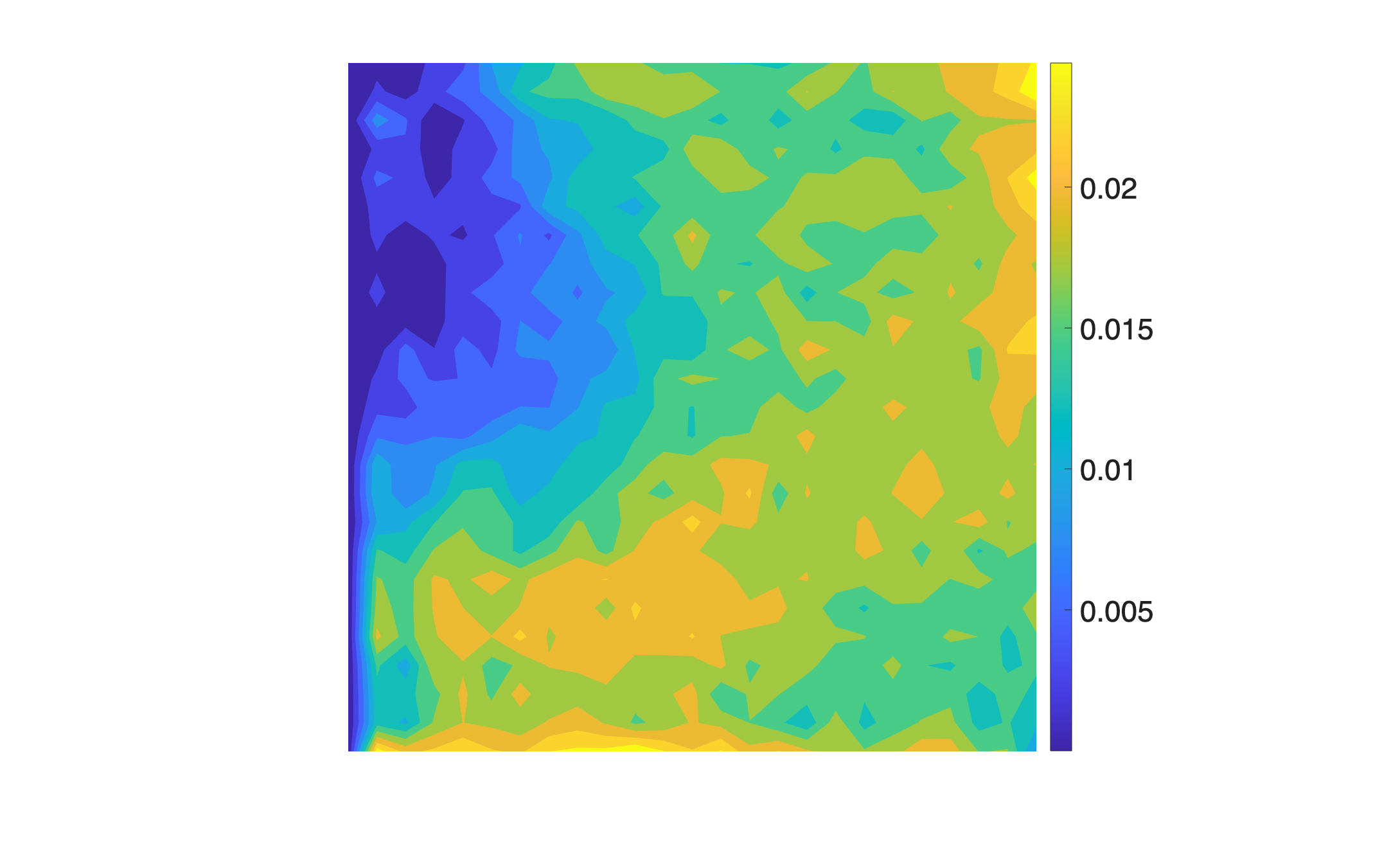}
    \includegraphics[trim = 50mm 0mm 20mm 0mm,width=0.45\textwidth]{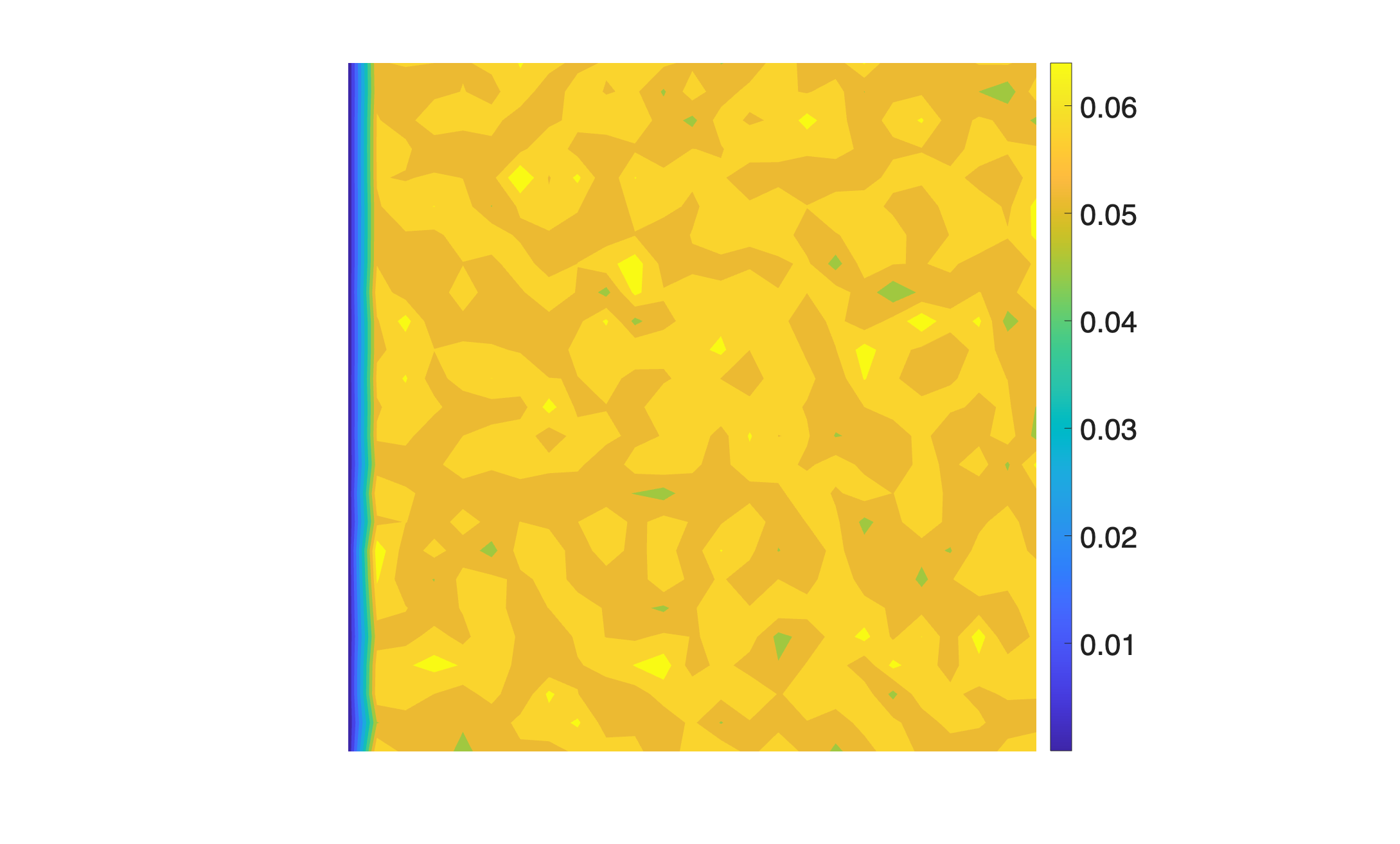}
    \caption{Spatial distribution of the absolute error $|y - \mu|$ (left), alongside the spatial distribution of the uncertainty spreads (right) for first, second and third 2D plate examples, respectively.}
    \label{NASA07}
\end{figure*}

Fig. \ref{NASA03} shows the relative error (RRMSE) obtained for both the training and test datasets. Figures \ref{NASA04}, \ref{NASA05}, and \ref{NASA06} show the inference results for some of the test simulations, comparing the reference result—the actual distribution of the elastic modulus that leads to the displacement field used for the inference input—and the neural network prediction. It is important to note that, due to the architecture used in this work, the network also provides a confidence interval for the elastic modulus at each point. The figures showing the results reflect the lower and upper limits of the elastic modulus for each point. To further facilitate the interpretation of the results and provide a better view of the model's performance, the absolute deviation between the ground truth and the predicted mean field ($|y - \mu|$), alongside spatial distributions of the uncertainty spreads are represented in Fig. \ref{NASA07}.


The results show how the neural network is capable of effectively recovering the nonlinear distribution of Young's modulus, as well as providing the distribution of uncertainty in the result offered, which allows us to see the areas of greater or lesser confidence in the results. {A two-sided confidence interval of $95.45\%$} has been used in the representation of the results. This confidence interval in our Gaussian formulation corresponds to $4\sigma$. 
These metrics allow for a direct assessment of the local precision and demonstrate that the uncertainty quantification is not a global scalar, but a position-dependent field that reflects the physical sensitivity of the inverse problem. The uncertainty spread depends of $\sigma_{\text{noise}}$ it has been a trainable parameter that represents statistical noise inherent for the data.

\begin{figure}
\centering
\begin{subfigure}{0.32\textwidth}
    \includegraphics[trim=370 110 370 50, clip,width=\textwidth]{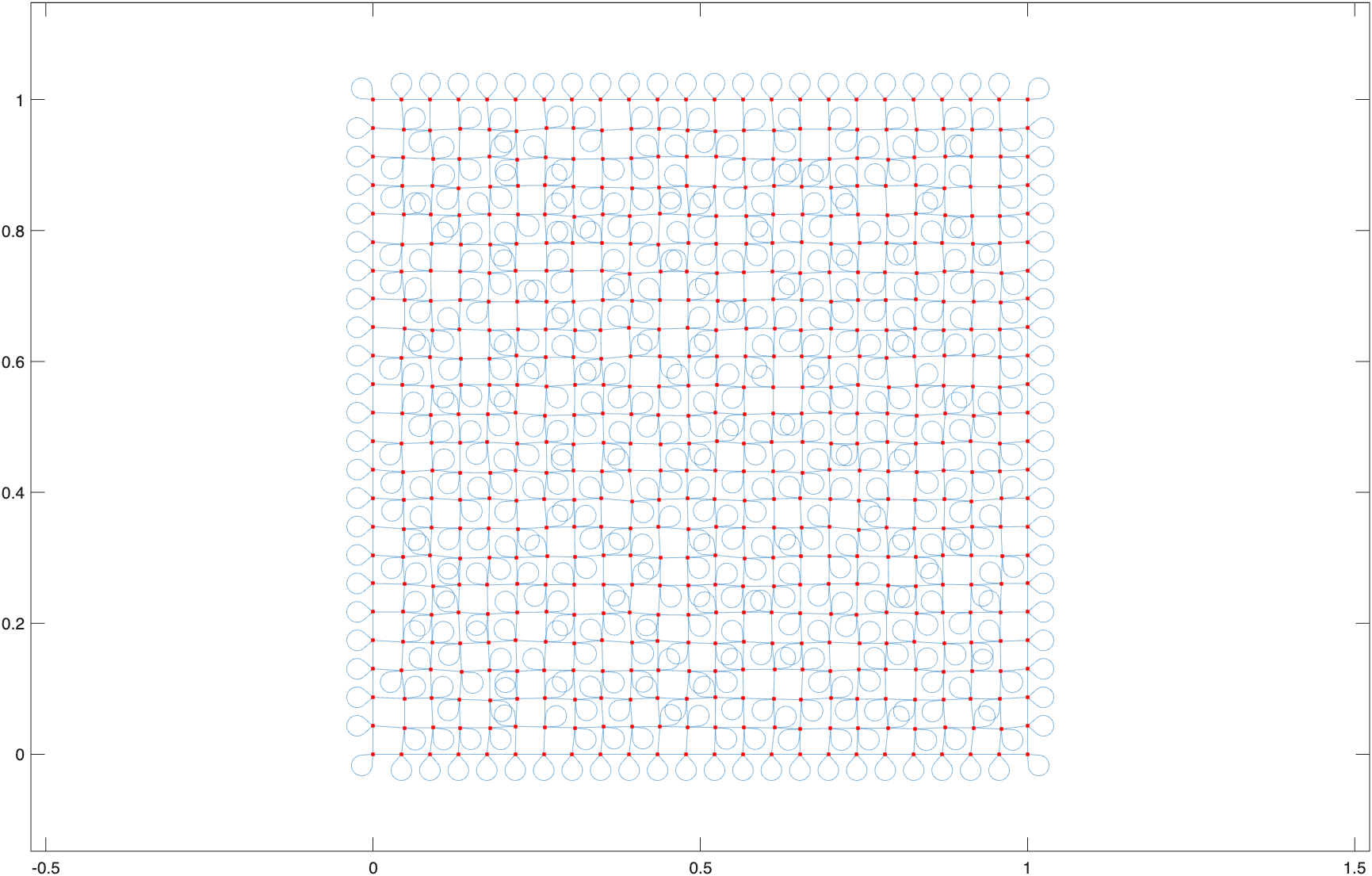}
    \caption{}
    \label{fig:first}
\end{subfigure}
\hfill
\begin{subfigure}{0.32\textwidth}
    \includegraphics[width=\textwidth]{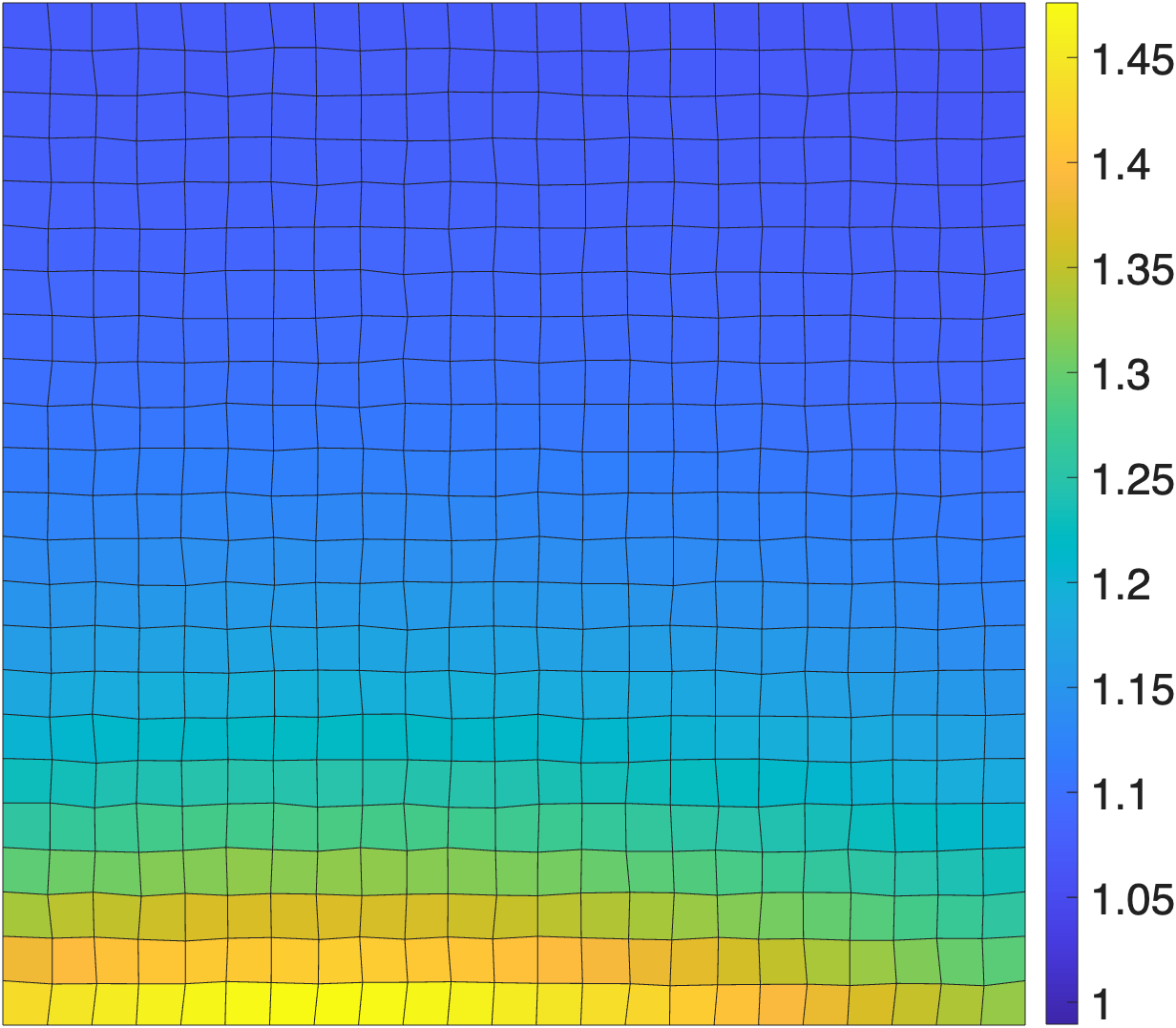}
    \caption{}
    \label{fig:second}
\end{subfigure}
\hfill
\begin{subfigure}{0.32\textwidth}
    \includegraphics[width=\textwidth]{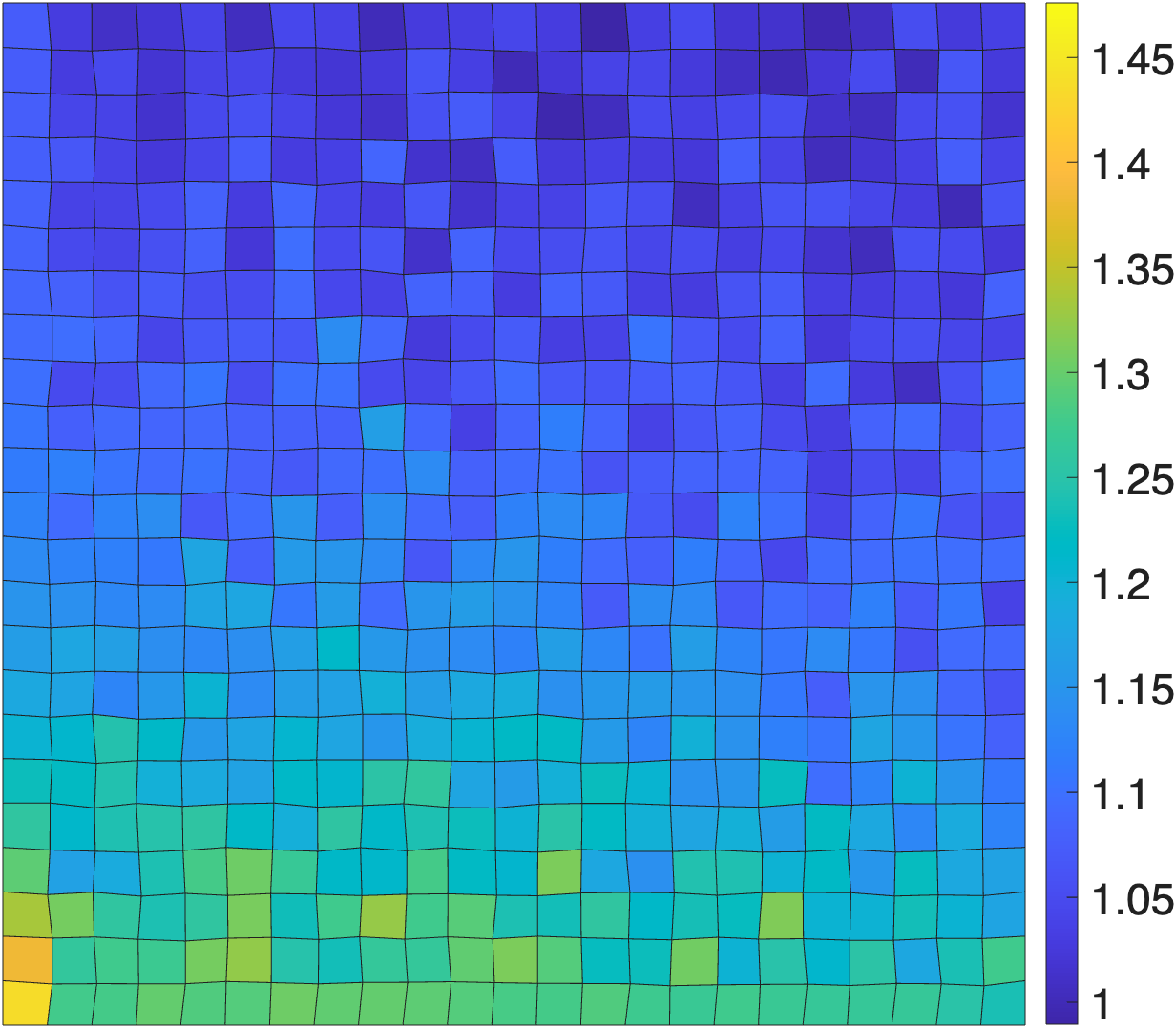}
    \caption{}
    \label{fig:third}
\end{subfigure}
\caption{{ (a) Graph associated to the modified mesh. (b) Ground truth of the Young's modulus distribution. (c) Obtained mean for the Young's modulus.}\label{f1}}
\end{figure}

{

\subsubsection{Robustness to mesh perturbations}

 One of the most interesting features of GNNs is their geometric robustness. It is well known that, when properly trained, they can cope with drastic changes in the domain’s geometry. To test this feature, we have attempted to make inferences using a perturbed, irregular mesh, different from the original one used for training.
 
 To this end, an inference was carried out on a new mesh, consisting of 576 nodes, compared with the 625 used in the previous training and testing phases, which was slightly distorted so that it was not perfectly regular. The mean relative error, calculated on the basis of the mean of the estimated field, is 4 per cent, which demonstrates the robustness of the proposed method. Results are presented in Fig. \ref{f1}.}
 
\subsection{Estimation of the Load Applied to a 3D Hyperelastic Beam}\label{Ej:Beam}
\begin{figure*}[ht]
\centering
\includegraphics[width=5.5cm]{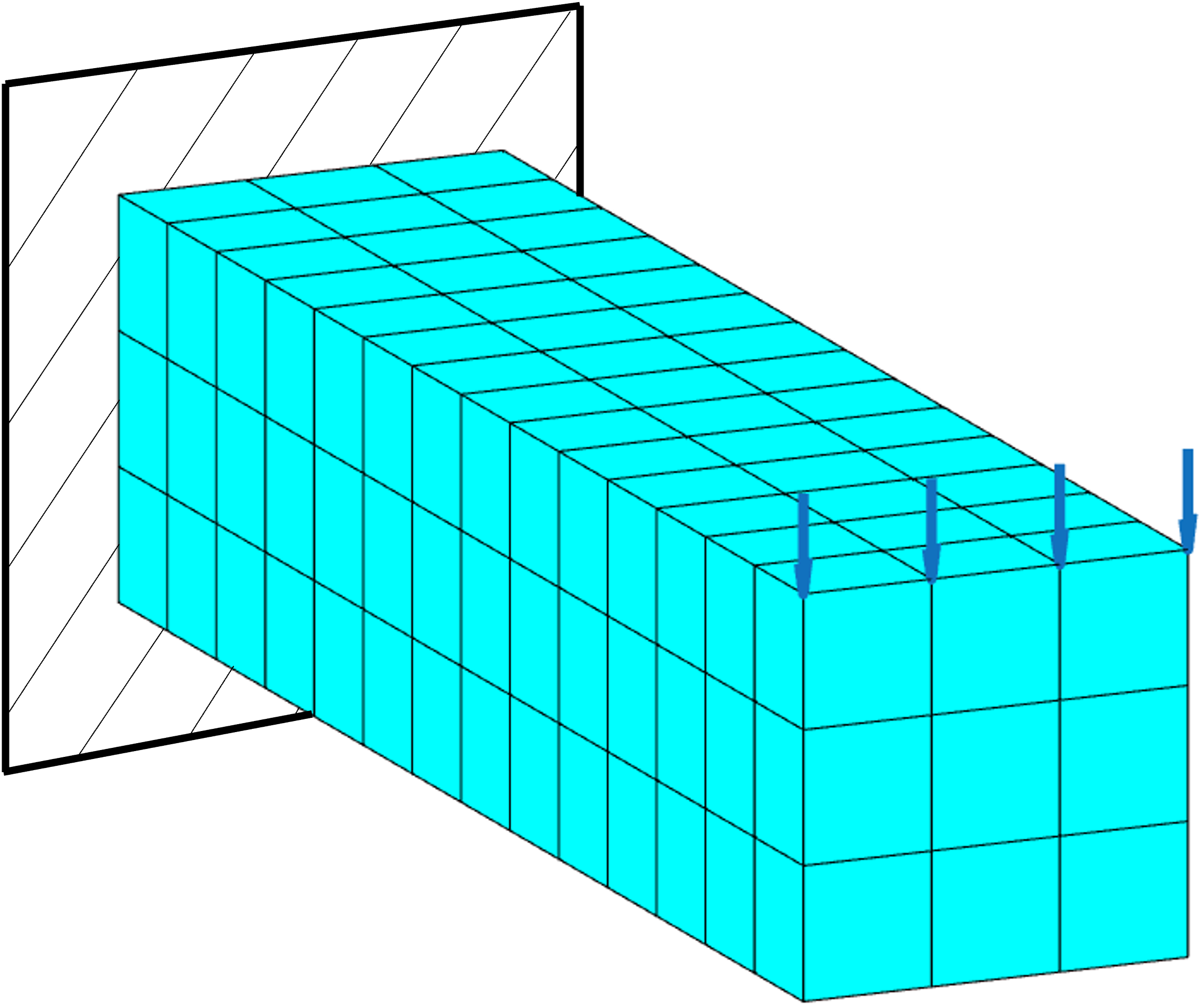}
\caption{Example of a loaded mesh associated with the hyperelastic beam problem.}
\label{BEAM_01}
\end{figure*}

In this second example, we seek to locate the position and value of the load applied to a cantilevered hyperelastic beam, given its displacement field. The beam is assumed to follow a Neo-Hookean constitutive law, with loads ranging from $F \in [5, 100] kN$, every $5kN$ and at different positions on its upper face from the cantilevered end to an area near the fixed end, specifically linear loads at $13$ in the transverse direction and on the upper face. The dataset consists of $260$ high-fidelity simulations performed using the finite element method with nonlinear behavior of the hyperelastic beam with dimensions $40\times10\times 10$. In Fig. \ref{BEAM_01} a loaded, meshed 3D beam is represented.

\begin{figure*}[ht]
\centering
\includegraphics[width=0.95\textwidth]{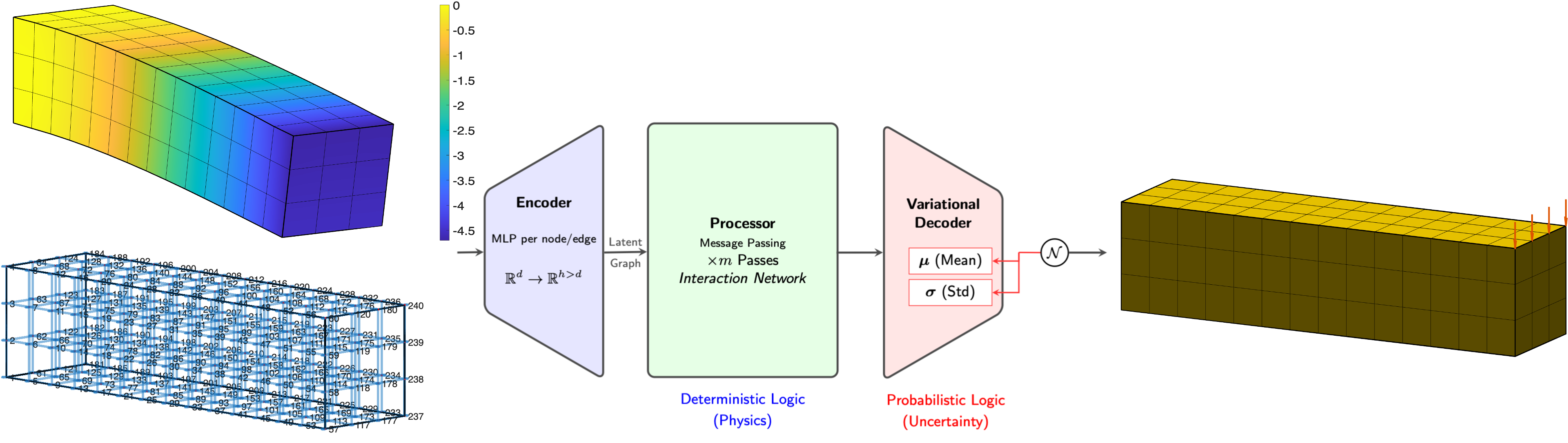}
\caption{Schematic procedure for training the 3D hyperelastic beam problem.}
\label{BEAM_02}
\end{figure*}

For every experiment meshes are identical, so the only difference between each simulation is the position and value of the applied load, which induces a displacement field that will be the data feeding the neural network training. All simulations have the same Dirichlet boundary conditions and material characteristics. For all the above reasons, the associated graph, see Fig. \ref{BEAM_02}, does not vary and consists of $240$ nodes and $824$ edges (it has been considered appropriate to generate self-informed connectivity) in a three-dimensional distribution. In Fig. \ref{BEAM_02} a schematic procedure to train our network architecture is shown. The input dataset is composed exclusively of a displacement field and a graph connectivity, the output is the position and value of the loads that perform those displacements. Once neural networks is trained, in inference only needs new input data (displacement and connectivity) to obtain the inverse problem result: position and value of loads.

\subsubsection{Training parameters}
The neural network used in this example has only $6900$ parameters, and the most important hyperparameters are summarized in Table \ref{tab:ex02}.

\begin{table}[htbp]
    \centering
    \caption{Most important hyperparameters used in Example \ref{Ej:Beam}} \vspace{0.1cm}
    \label{tab:ex02}    
    \begin{tabular}{l|c}
        \textbf{Hyperparameter} & \textbf{Value}\\
        \hline
        {\tt nBatch}  & $2$ \\
        Latent space dimension & $12$ neurons\\
        Message Passing steps ($m$) & $4$ \\
        $\sigma_1$ &  $\frac{1}{\exp(1)}$ \\
        $\sigma_2$ &  $\frac{1}{\exp(2)}$ \\
        Activation function & SwishLayer \\
        Number of Epochs & $5000$ \\ 
        Optimizer & ADAM \\
        Learning Rate & $10^{-3}$ \\
        Learning rate decay & $98\%$ every $1000$ epochs\\
    \end{tabular}
\end{table}

Once again, ${\tt {\tt nBatch}}=2$ has been used, which corresponds to $2$ simulations of the entire mesh, for a single load value. In addition, It was not considered necessary to add noise to the data.

\begin{figure*}[ht]
\centering
\includegraphics[width=0.4\linewidth]{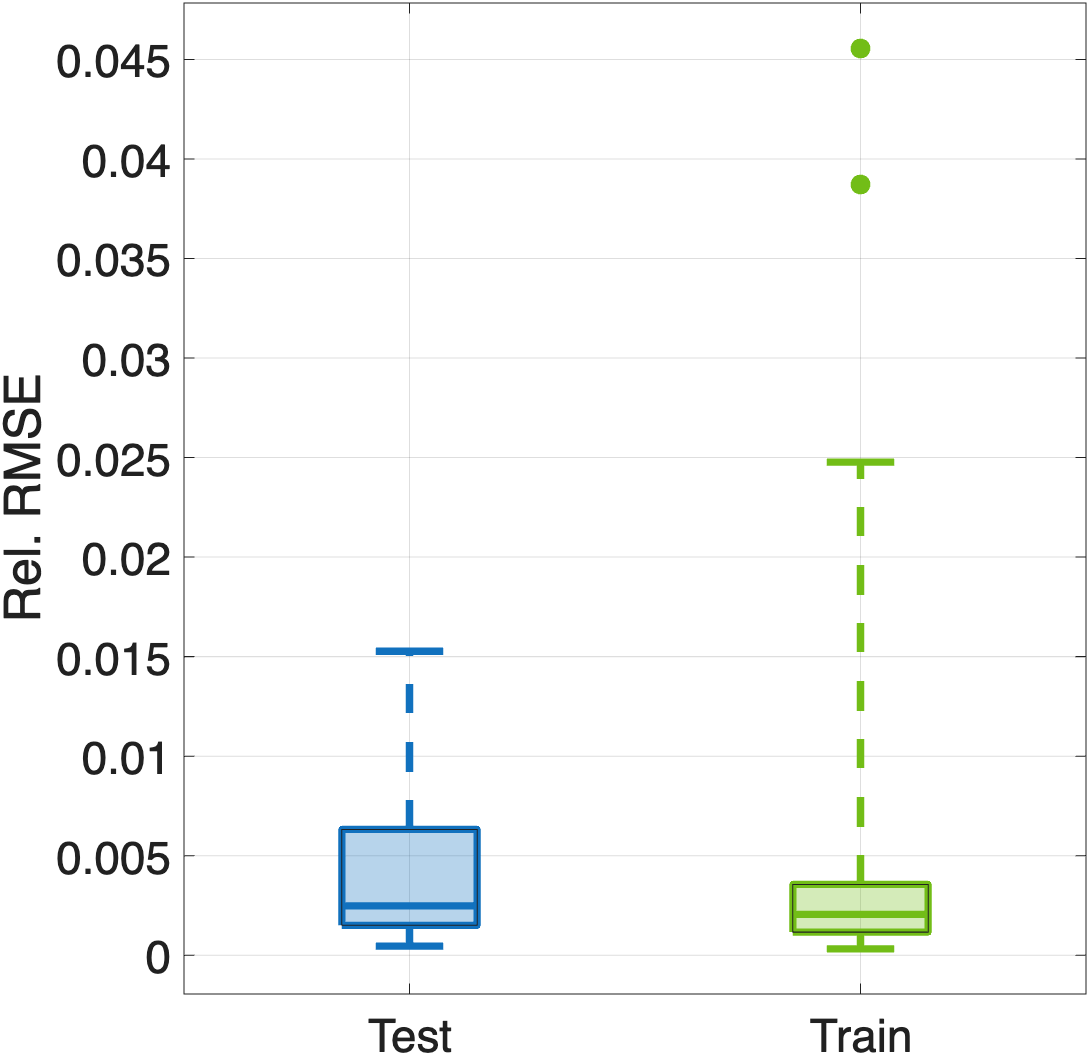} 
\caption{Relative RMSE error for training and test data for the 3D Hyperelastic Beam problem.}
\label{BEAM_03}
\end{figure*}

\begin{figure*}[h!]
\centering
\includegraphics[width=\linewidth]{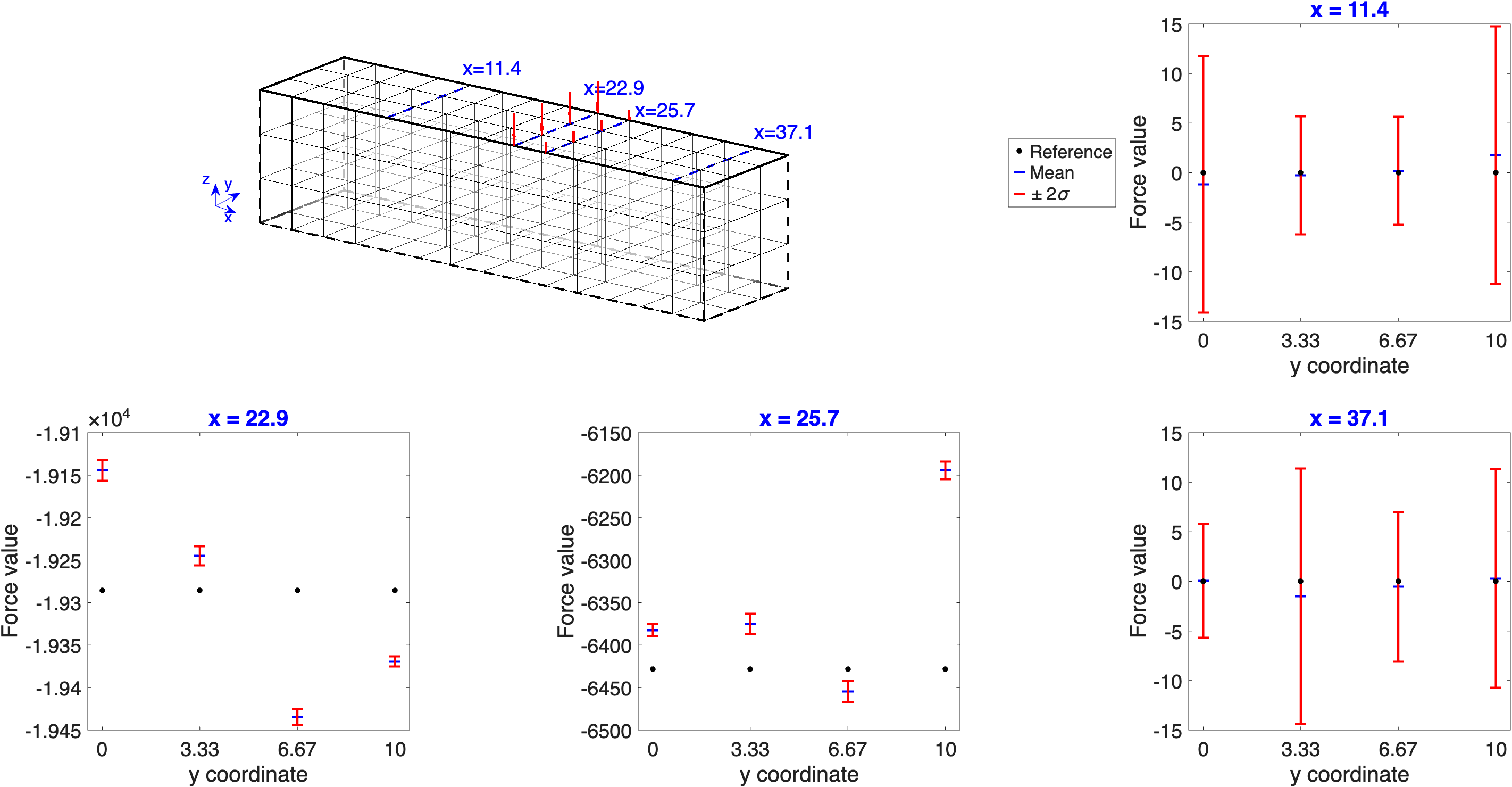} 
\caption{{Analysis of the first test beam. For each node on the top load surface, the reference, mean and limit load values are shown. }}
\label{BEAM_04}
\end{figure*}

\begin{figure*}[h!]
\centering
\includegraphics[width=\linewidth]{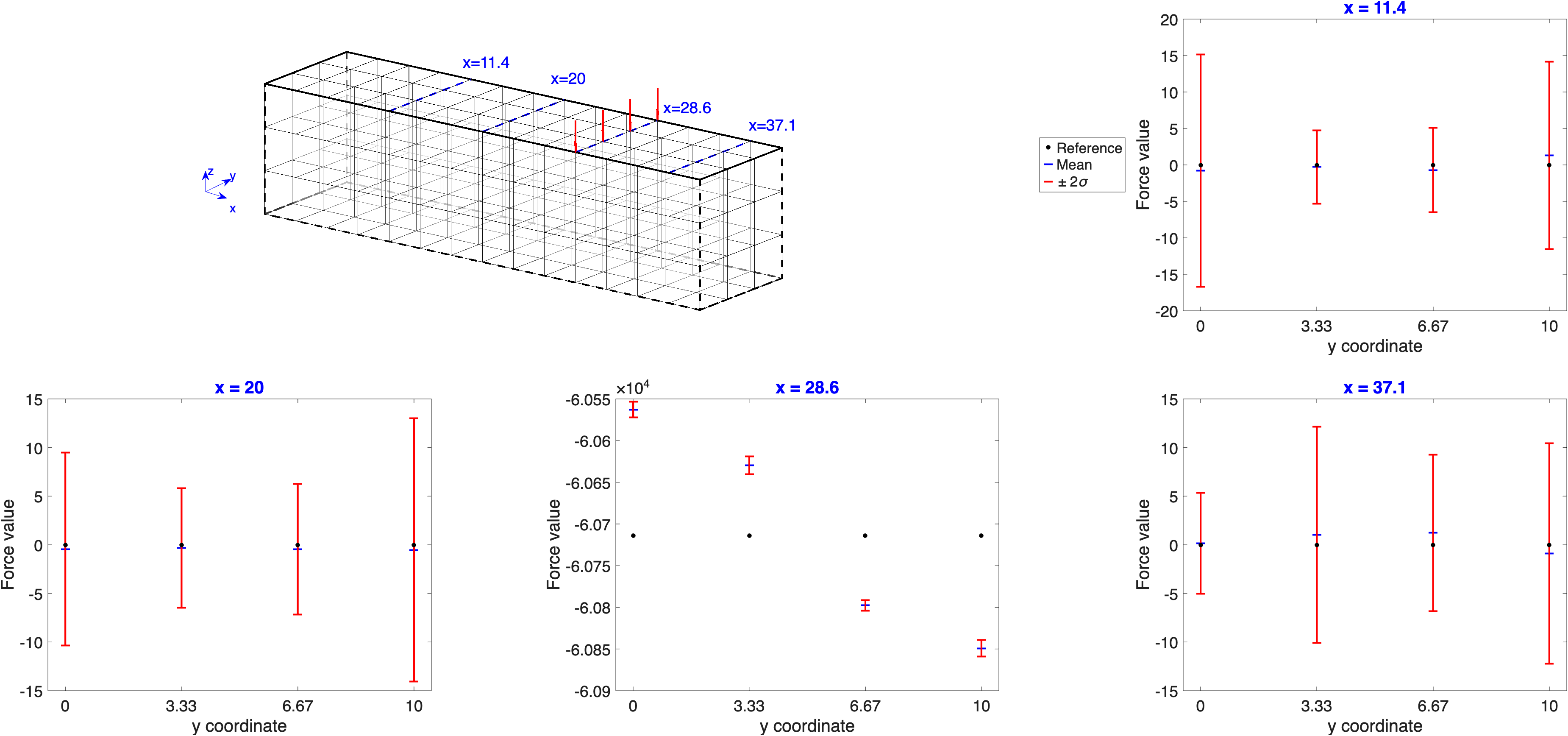} 
\caption{{Analysis of the second test beam. For each node on the top load surface, the reference, mean and limit load values are shown.} }
\label{BEAM_05}
\end{figure*}

\begin{figure*}[h!]
\centering
\includegraphics[width=\linewidth]{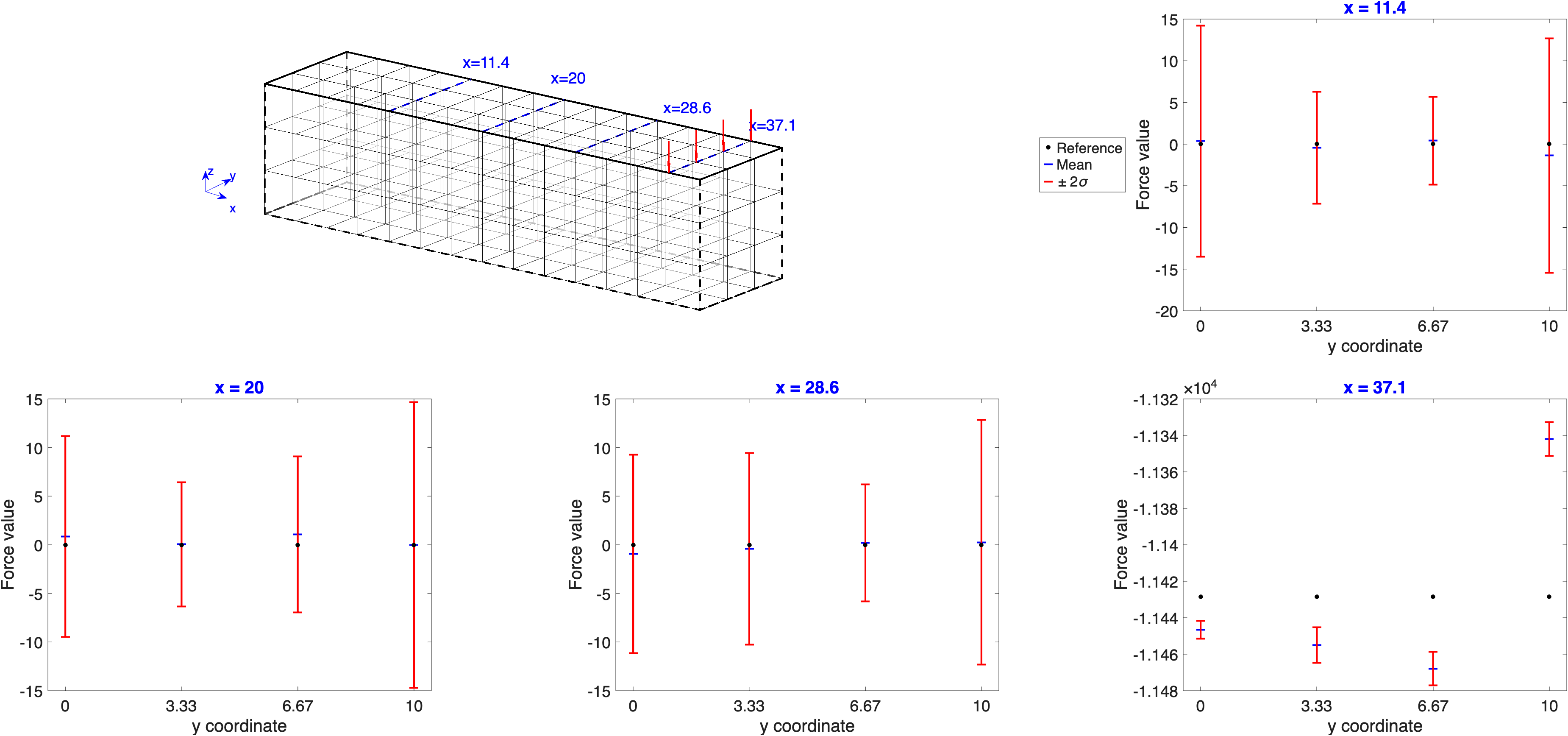} 
\caption{{Analysis of a third test beam. For each node on the upper loaded surface the reference, mean and bounded values of the load are represented.}}
\label{BEAM_06}
\end{figure*}

Fig. \ref{BEAM_03} shows the relative error (RRMSE) obtained for both the training and test datasets.  Figs. \ref{BEAM_04}, \ref{BEAM_05}, and \ref{BEAM_06} show the inference results for three of the test simulations, comparing the reference result—the actual value and position of the load applied to the beam, which leads to the displacement field used for the inference input and represented along the beam—with the predictions generated by the neural network. For each experiment, the average value of the load at each node is represented. Since this value is zero or almost zero at most nodes, it will not be reflected in the result, as the vectors are shown relative to their modulus. Again, the network also provides a confidence interval for the value of this load at every point in the domain.

In Table \ref{tab:ex03} the values in Figs. \ref{BEAM_04}, \ref{BEAM_05}, and \ref{BEAM_06} are quantified. Although in some cases the reference value falls outside the confidence interval, the relative error is minimal, and perhaps most importantly, the location of the loaded nodes is excellent. Those values are computed as a mean of loaded nodes in each experiment, for predicted values, only a mean of higher values, the representative ones, are considered.

\begin{table}[htbp]
    \centering
    \caption{Load values in loaded nodes in example \ref{Ej:Beam}} \vspace{0.1cm}
    \label{tab:ex03}    
    \begin{tabular}{c|c|c|c|c|c|c}
        \textbf{Example} & \textbf{Ref. Load}& \textbf{Lower Bd.}& \textbf{Mean}& \textbf{Upper Bd.}& \textbf{Spread Size}& \textbf{Rel. Error}\\
        \hline
        $1^{st} $& $-12857$ & $-12824$ & $-12834$ & $-12814$ & $20.38$ & $1.5E-3$ \\
        $2^{nd}$ & $-60714$ & $-60710$ & $-60719$ & $-60701$ & $17.74$ & $2.9E-4$ \\
        $3^{rd}$ & $-11428$ & $-11427$ & $-11436$ & $-11418$ & $17.94$ & $1.5E-3$ \\ \hline
    \end{tabular}
\end{table}

\subsection{Estimation of the Load Applied to a previously unseen Beam}
\begin{figure*}[h!]
\centering
\includegraphics[width=0.5\linewidth]{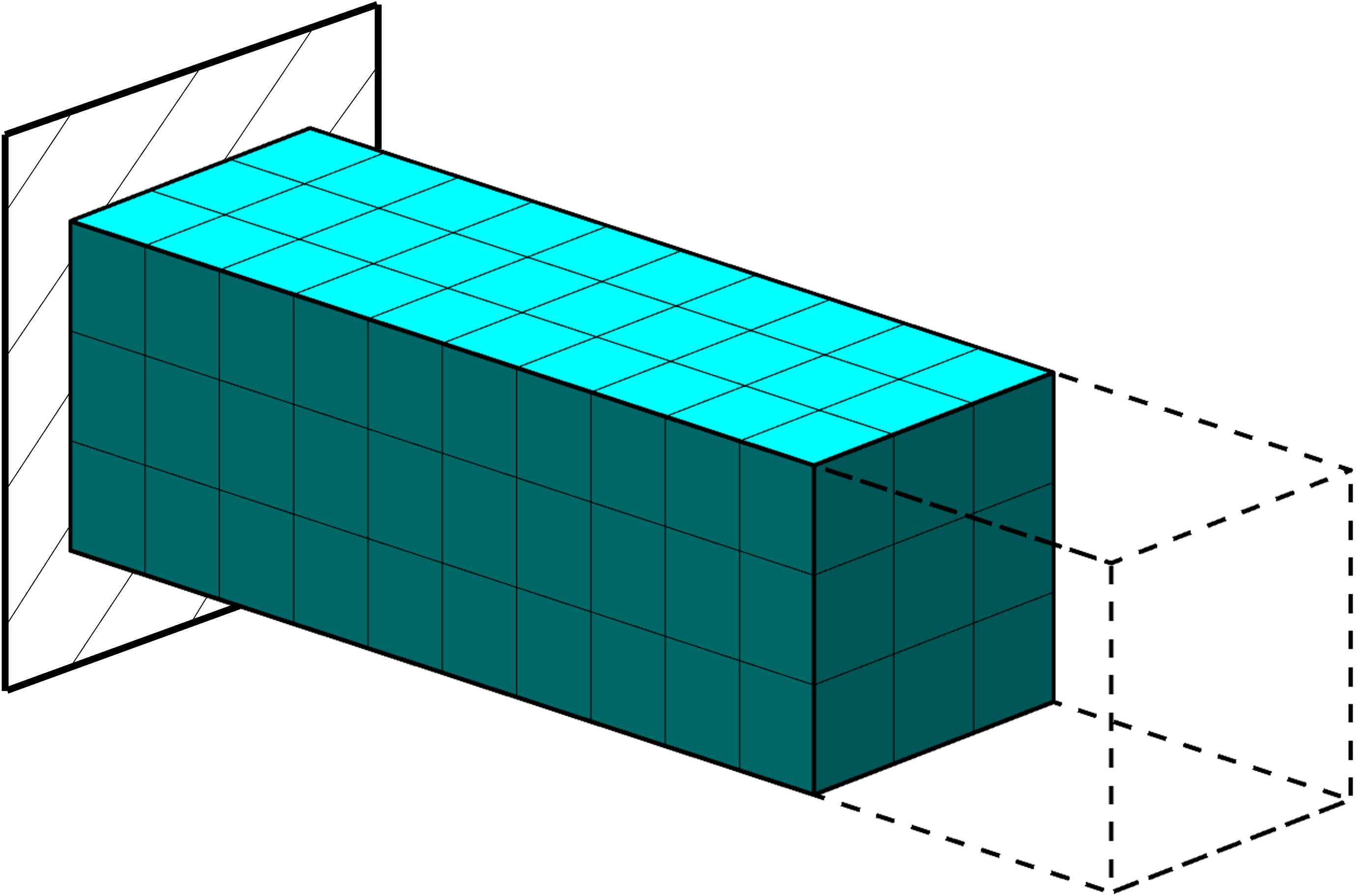} 
\caption{Comparative of geometries of trained beam and the new shorter one.}
\label{BEAM_07}
\end{figure*}

In this last example, the training of the previous hyperelastic beam has been used to estimate the value and position of the load on a new beam, with a different geometry to that trained in the previous Section \ref{Ej:Beam}. Fig. \ref{BEAM_07} compares the geometry of the beam used to train the neural network with the new (shorter) beam for which the load is to be predicted.

\begin{figure*}[h!]
\centering
\includegraphics[width=0.4\linewidth]{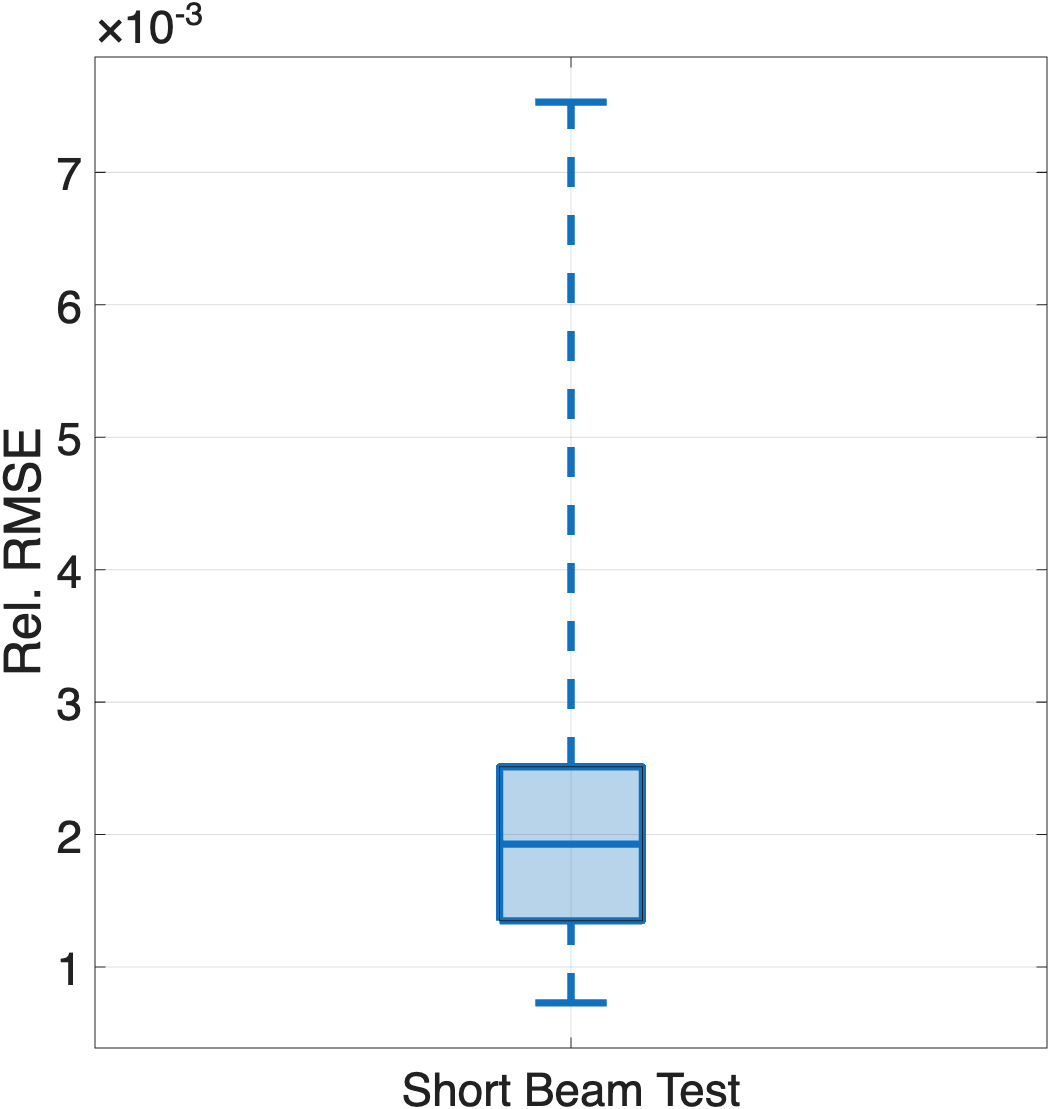} 
\caption{Relative RMSE error for test data for the new (shorter) 3D hyperelastic  Beam problem.}
\label{BEAM_08}
\end{figure*}

Fig. \ref{BEAM_08} shows the relative error in the prediction of $10$ new experiments, in which the value and position of the load are changed.

\begin{figure*}[h!]
\centering
\includegraphics[width=\linewidth]{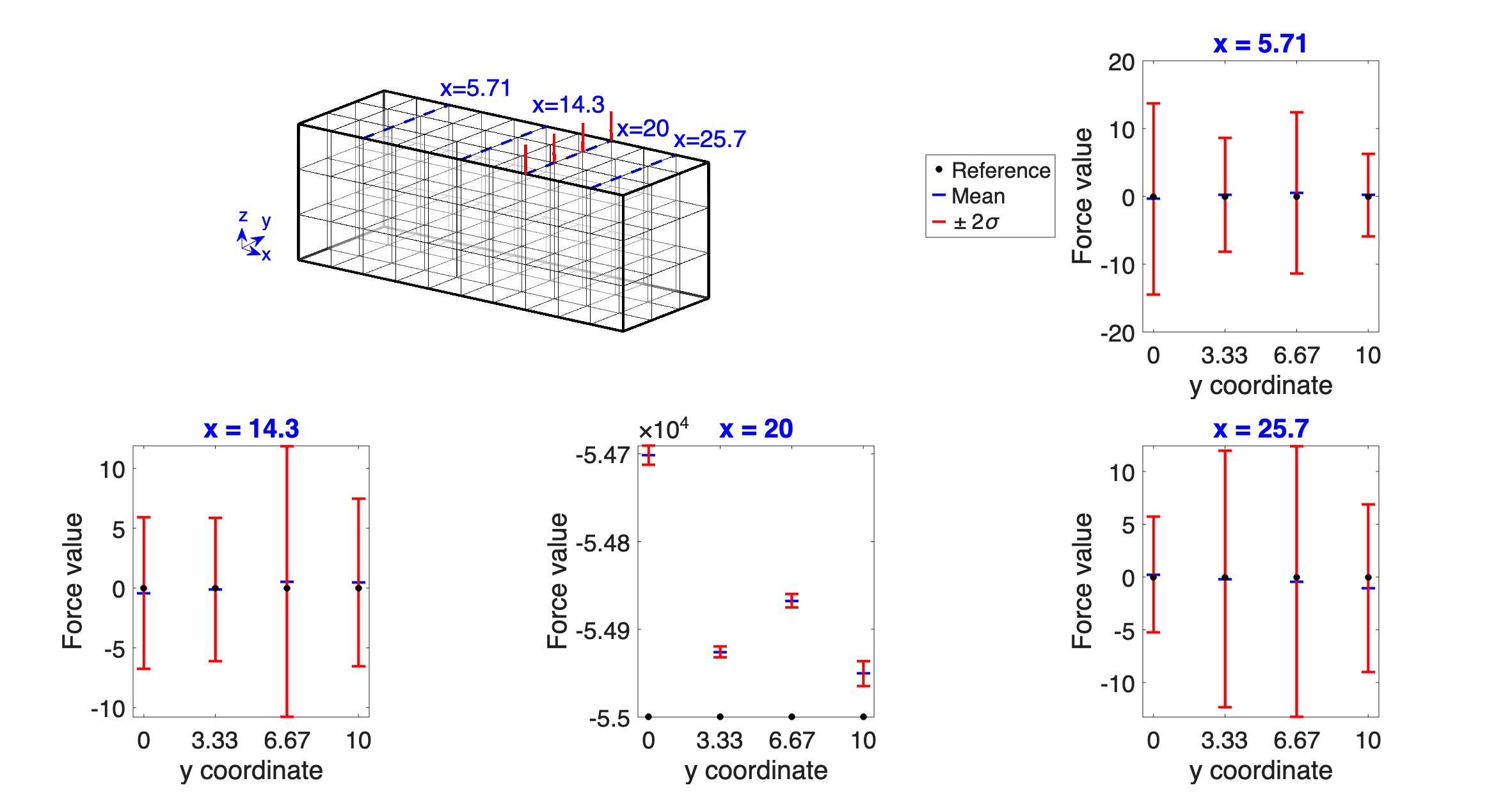} 
\caption{{Analysis of the first test short beam. For each node on the top load surface, the reference, mean and limit load values are shown. }}
\label{BEAM_09}
\end{figure*}

\begin{figure*}[h!]
\centering
\includegraphics[width=\linewidth]{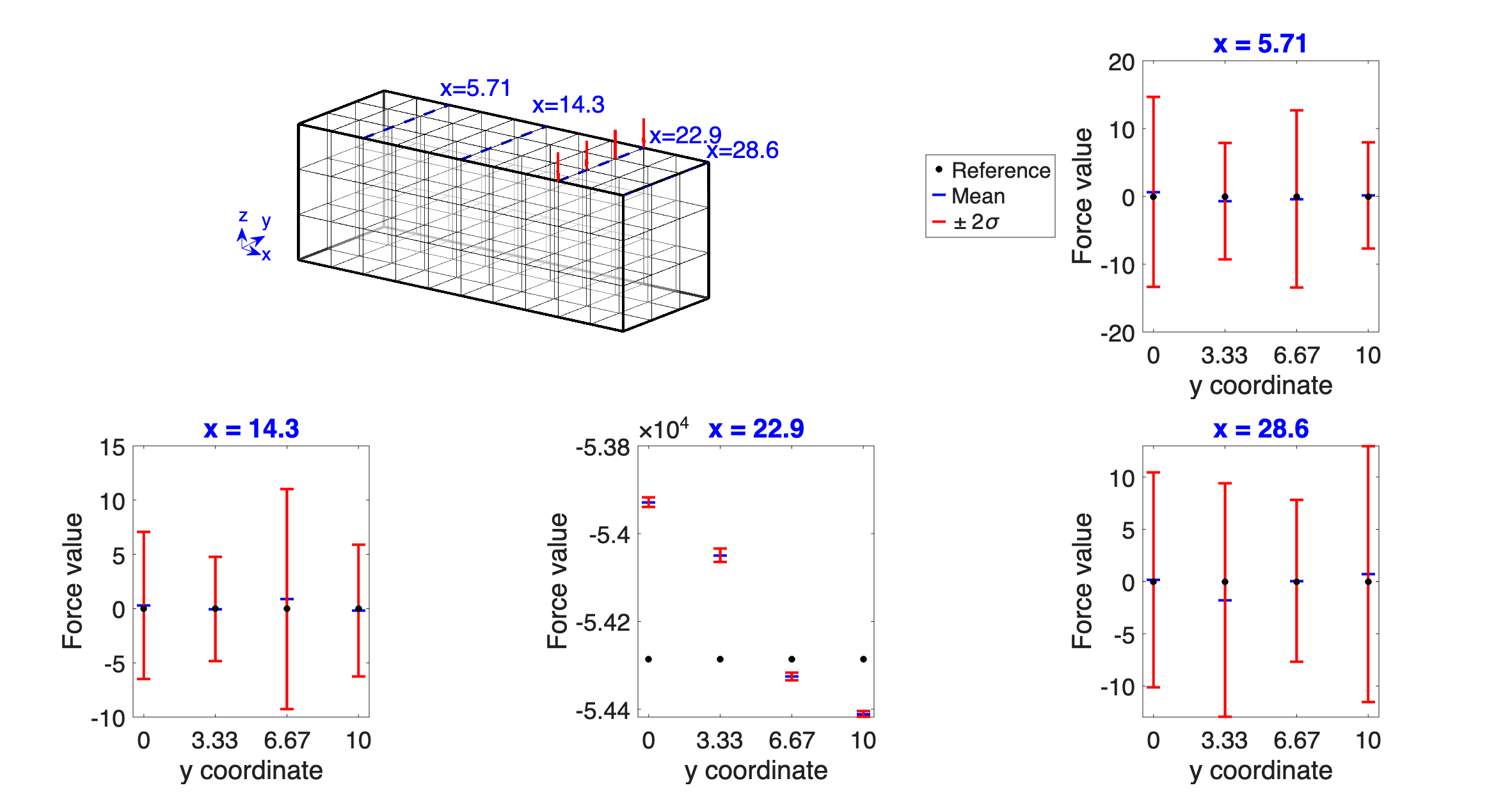} 
\caption{{Analysis of the second test short beam. For each node on the top load surface, the reference, mean and limit load values are shown. }}
\label{BEAM_10}
\end{figure*}

\begin{figure*}[h!]
\centering
\includegraphics[width=\linewidth]{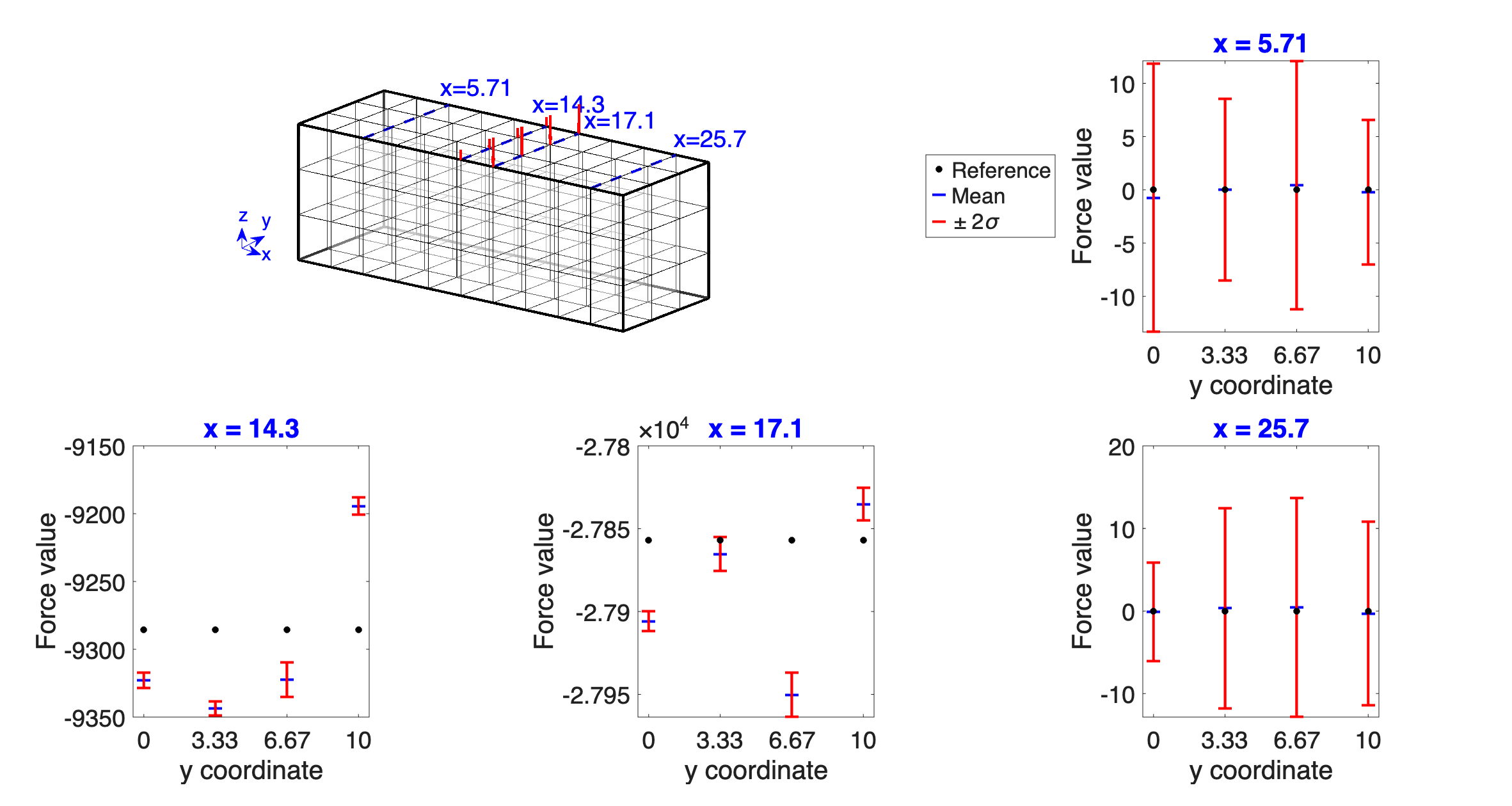} 
\caption{{Analysis of the third test short beam. For each node on the top load surface, the reference, mean and limit load values are shown. }}
\label{BEAM_11}
\end{figure*}

{In the Figs.\ref{BEAM_09}, \ref{BEAM_10} and \ref{BEAM_11} we show the inference results for three of the test simulations for previously unseen geometries, comparing the reference  value and position of the load applied to the new beam, which
leads to the displacement field used for the inference input and represented along the beam, with the predictions generated by the neural network. Again, for each experiment, the average value of the load at each node is represented. As previous examples, the network also provides a confidence interval for the value of this load at every point in the domain.}

\section{Conclusions}\label{s5}

The implementation of variational layers in graph neural network architectures allows for robust estimation of uncertainty thanks to the variational nature of the neural network without sacrificing the predictive and versatile performance offered by graph networks. The training algorithm presented effectively balances model complexity (KL divergence) and accuracy (likelihood), providing a valuable tool for complex problems where risk must be controlled.

The methodology developed in this work maintains the fundamental characteristics of GNNs by respecting the topology of the meshes of the proposed problems. The variational formulation in the decoder allows real-time confidence metrics to be obtained, and its application in quantifying uncertainty in nonlinear inverse problems confirms the robustness of the method.

\section*{Acknowledgements}

This work was supported by the Spanish Ministry of Science and Innovation, AEI/10.13039/501100011033, through
Grant numbers PID2023-147373OB-I00 and PID2023-148952OB-I00, and by the Ministry for Digital Transformation and the Civil Service, through
the ENIA 2022 Chairs for the creation of university-industry chairs in AI, through Grant TSI-100930-2023-1.

Funded by the European Union. Views and opinions expressed are however those
of the author(s) only and do not necessarily reflect those of the European Union or
the European Research Council Executive Agency. Neither the European Union nor
the granting authority can be held responsible for them.
This work is supported by ERC grant PHYSIA 101264273.

The authors also acknowledge the support of ESI Group, Keysight Technologies, through the chair at the University of Zaragoza.

AM is a Fellow of the Serra H\'unter Programme of the Generalitat de Catalunya.

BM acknowledges the support of the CPJ-ANR ITTI.

\end{document}